# UNIVERSIDAD POLITÉCNICA DE VALENCIA

## Escuela Técnica Superior de Ingeniería Informática

\_\_\_\_\_\_\_\_\_\_\_\_\_\_\_\_\_\_

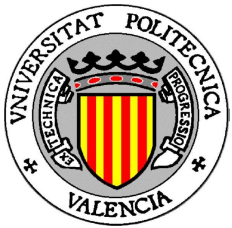 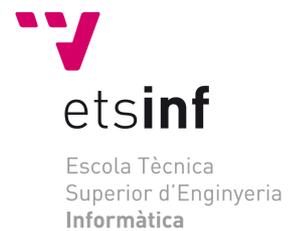

# Una arquitectura para la evaluación de sistemas inteligentes

**PROYECTO FINAL DE CARRERA**

Realizado por:

**Javier Insa Cabrera**

Dirigido por:

**José Hernández Orallo**

Valencia, 01 de octubre de 2010

# Resumen

*Uno de los principales ámbitos de investigación en Inteligencia Artificial es la codificación de agentes (programas) que sean capaces de aprender por sí solos en cualquier situación y que no sirvan únicamente para el fin que fueron creados, como por ejemplo para jugar al ajedrez. De este modo pretendemos acercarnos más a lo que realmente es la llamada Inteligencia Artificial.*

*Uno de los problemas para saber si realmente un agente es inteligente o no es la medición de su inteligencia, ya que de momento no existe forma alguna de medirla de forma fiable.*

*El objetivo de este proyecto es la creación de un intérprete que permita ejecutar diferentes entornos, incluso generados aleatoriamente, para que un agente (una persona o un programa) pueda interactuar en éstos. Una vez que la interacción entre el agente y el entorno ha terminado, el intérprete medirá la inteligencia del agente en función de los diferentes estados por los que ha pasado el entorno y las acciones realizadas en cada estado por el agente durante la prueba.*

*De este modo se conseguirá medir la inteligencia de los agentes en cualquier entorno posible, así como hacer comparaciones entre los distintos agentes evaluando cual de todos ellos es el más inteligente.*

*Para poder realizar las pruebas, el intérprete debe ser capaz de generar aleatoriamente distintos entornos que realmente sean útiles para la medición de la inteligencia de los agentes, ya que no cualquier entorno generado aleatoriamente será útil para tal propósito.*

# Resum

*Un dels principals àmbits d'investigació en Intel·ligència Artificial és la codificació d'agents (programes) que siguen capaços d'aprendre per si mateixos en qualsevol situació i que no servisquen únicament per la finalitat que van ser creats, com per exemple per jugar als escacs. D'aquesta manera pretenem apropar-nos més al que realment és l'anomenada Intel·ligència Artificial.*

*Un dels problemes per saber si realment un agent és intel·ligent o no és el mesurament de la seua intel·ligència, com que de moment no hi ha cap manera de mesurar-la de forma fiable.*

*L'objectiu d'aquest projecte és la creació d'un intèrpret que permeta executar diferents entorns, fins i tot generats aleatòriament, perquè un agent (una persona o un programa) puga interactuar dins d'aquests. Una vegada que la interacció entre l'agent i l'entorn ha acabat, l'intèrpret mesurarà la intel·ligència de l'agent en funció dels diferents estats pels quals ha passat l'entorn i les accions realitzades en cada estat per l'agent durant la prova.*

*D'aquesta manera s'aconseguirà mesurar la intel·ligència dels agents en qualsevol entorn possible, així com fer comparacions entre els diferents agents avaluant quin de tots ells és el més intel·ligent.*

*Per poder realitzar les proves, l'intèrpret ha de ser capaç de generar aleatòriament diferents entorns que realment siguen útils per al mesurament de la intel·ligència dels agents, atés que no qualsevol entorn generat aleatòriament serà útil per al propòsit esmentat.*

# Abstract

*One of the main research areas in Artificial Intelligence is the coding of agents (programs) which are able to learn by themselves in any situation. This means that agents must be useful for purposes other than those they were created for, as, for example, playing chess. In this way we try to get closer to the pristine goal of Artificial Intelligence.*

*One of the problems to decide whether an agent is really intelligent or not is the measurement of its intelligence, since there is currently no way to measure it in a reliable way.*

*The purpose of this project is to create an interpreter that allows for the execution of several environments, including those which are generated randomly, so that an agent (a person or a program) can interact with them. Once the interaction between the agent and the environment is over, the interpreter will measure the intelligence of the agent according to the actions, states and rewards the agent has undergone inside the environment during the test.*

*As a result we will be able to measure agents' intelligence in any possible environment, and to make comparisons between several agents, in order to determine which of them is the most intelligent.*

*In order to perform the tests, the interpreter must be able to randomly generate environments that are really useful to measure agents' intelligence, since not any randomly generated environment will serve that purpose.*

# Índice

# 1. Introducción

## 1.1. Contexto

Este proyecto de fin de carrera surge a raíz de una Acción Complementaria dentro del Programa Explora-Ingenio financiada por el Ministerio de Ciencia e Innovación con título "**Anytime Universal Intelligence (ANYNT)**" (http://users.dsic.upv.es/proy/anynt/), cuyo investigador principal es José Hernández Orallo. El objetivo del programa Explora-Ingenio es explorar líneas de investigación de alto riesgo e impacto, posiblemente heterodoxas, cuya viabilidad se intenta determinar. Según el equipo que evaluó el proyecto, éste "es una propuesta con la suficiente osadía intelectual y riesgo como para ser considerada en este Programa explora".

En el contexto de este proyecto de investigación, en 2009 se ofertaron en la ETSINF diferentes proyectos de fin de carrera relacionados con la implementación de prototipos y estructuras básicas de los tests de inteligencia que se proponía desarrollar. A finales del 2009, empecé a trabajar en el tema con José Hernández Orallo. El propio carácter del proyecto ANYNT implicaba que el proyecto de fin de carrera que me proponía tenía un carácter marcadamente investigador y que, frecuentemente, habría que tomar decisiones de implementación y de evaluación a medida que avanzara la implementación o se obtuviera algún resultado investigador. En definitiva, un proyecto bastante cerrado en su contexto (el del propio proyecto de investigación) pero inicialmente bastante abierto en objetivos, y que naturalmente ha de ser entendido en un contexto investigador, y no tanto desde la perspectiva de un buen acabado de un producto final.

Durante la primera mitad de 2010, y compaginando los estudios, el proyecto fin de carrera fue avanzando, siguiendo fundamentalmente las publicaciones [Hernandez-Orallo 2010a] [Hernandez-Orallo 2010b] y [Hernandez-Orallo & Dowe 2010]. En junio de 2010 me presenté a una convocatoria de contratado dentro del mismo proyecto de investigación y fui seleccionado, principalmente por el premio extraordinario al expediente académico. Desde mi incorporación como contratado, el proyecto de fin de carrera constituye un primer paso a partir del cual seguir trabajando en una segunda fase más ambiciosa.

## 1.2. Motivación

Una de las cuestiones más debatidas en el campo de la inteligencia artificial, la ciencia cognitiva, la psicometría, la biología y la antropología es la definición de inteligencia y su medición.

La medición de la inteligencia humana es de suma relevancia para muchas áreas de aplicación. Consecuentemente, los tests psicométricos son comunes para (i) ayudar a niños y alumnos a aprender de manera eficiente (educación), (ii) evaluar al personal (selección y reclutamiento) y también (iii) en el tratamiento de diferentes enfermedades y problemas de aprendizaje (terapias de aprendizaje).

De una manera bastante similar, la medición de la inteligencia en máquinas es requerida en áreas tales como (i) crear agentes, robots, y otros tipos de "sistemas inteligentes" que adquieran conocimientos rápidamente (adquisición de conocimiento), (ii) para seleccionar/evaluar las habilidades de "sistemas inteligentes" (acreditación/certificación) y también (iii) cuando se trata de corregir o mejorar las capacidades de estos sistemas (diseño o corrección de sistemas inteligentes).

Aunque la terminología es diferente, las similitudes en ambas áreas son claras. Adicionalmente, una tercera área afín es la psicología o cognición comparativa para varios tipos de animales, donde existe un debate sobre cómo evaluar las habilidades cognitivas en los animales superiores (en especial los grandes simios y cetáceos).

El caso de la medición de la inteligencia en máquinas es especialmente relevante debido a varias cuestiones:

- Una ciencia o tecnología no puede avanzar si no disponemos de técnicas de medición adecuadas para evaluar sus progresos. Es difícil desarrollar la aeronáutica incluso a través de la observación de aves en pleno vuelo si no tenemos las mediciones de peso, altura o de aerodinámica. La inteligencia artificial tiene una referencia (el homo sapiens) pero sufre una falta de medidas y técnicas de medición para sus artefactos.
- La ubicuidad de los robots, ayudantes, mayordomos y otros tipos de agentes autónomos que realizan tareas en lugar de los seres humanos, hace que sea difícil tratar apropiadamente con ellos en un nuevo espacio de colaboración donde la interacción virtual sustituye a la interacción física, tales como plataformas de colaboración, redes sociales, arquitecturas orientadas a servicios, e-burocracia, Web 2.0, etc. En muchas aplicaciones necesitamos diferenciar entre estúpidos robots capacitados para realizar una única tarea y robots más generales e inteligentes.
    - Por un lado, algunas aplicaciones requieren que se les permita a los robots participar en proyectos o concederles permisos como interlocutores para la gestión, negociaciones o acuerdos. Por lo tanto, se espera que los robots sean eficaces para llevar a cabo las tareas que se les ha delegado, para las cuales tienen que ser inteligentes (desean evitarse los robots no inteligentes).
    - Por otro lado, en algunos casos, queremos evitar que robots de cualquier tipo puedan realizar ciertas tareas (p. ej. robots maliciosos podrían crear millones de cuentas en un servicio de correo gratuito). Con este fin, son extensamente utilizados los CAPTCHAs [von Ahn et al 2002] [von Ahn et al 2008], pero mecanismos más precisos y fiables serán necesarios en un futuro cercano.

Demos un ejemplo. Imaginemos un entorno colaborativo, donde varios agentes (algunos de ellos humanos y otros máquinas), tienen que construir un sitio web especializado con información sobre un tema específico (p. ej. una entrada de la Wikipedia, un esbozo de un proyecto europeo, un evento reciente, una biblioteca de música, un proyecto geográfico, etc.). El equipo tendrá que decidir las normas de interacción (cómo la información es aceptada, integrada y publicada). Para llegar a formar parte del equipo, cada componente necesitará *certificar* que él/ella tiene algunas habilidades básicas de razonamiento/aprendizaje. Si un agente inepto (o persona) trata de unirse al club requeriremos algún tipo de evaluación para decidir si debemos dejarle entrar.

¿Existen mecanismos como estos disponibles hoy en día? La respuesta es no. A parte de los CAPTCHAs, los cuales solamente miden habilidades muy específicas, no disponemos de ningún tipo de test donde podamos evaluar rápida y fiablemente la inteligencia de un sistema inteligente (un agente).

El propósito del proyecto de investigación es explorar las posibilidades de construir tests que sean:

- **Universales:** Pueden ser aplicados a humanos, robots, animales, comunidades o incluso híbridos. Para hacer frente a esta propiedad el test debe derivar de principios no antropomórficos universales fundados en ciencias de la computación y en la teoría de la información. Este objetivo, por sí mismo, es muy desafiante ya que los tests actuales se basan en suposiciones antropomórficas.
- **Anytime:** Se pueden adaptar a la velocidad del examinado, y pueden adaptarse dinámicamente sus preguntas, interacciones o temas al examinado, a fin de evaluar su inteligencia más rápido. El test debe ser lo suficientemente flexible para evaluar muy rápidamente (con poca fiabilidad) para algunas aplicaciones (p. ej. aplicaciones donde los CAPTCHAs son usados hoy en día) o proporcionar estimaciones de alta fiabilidad si se proporciona más tiempo (p. ej. aplicaciones de las cuales se derivan serias consecuencias debido a la intromisión de un agente estúpido).

Siguiendo las ideas de la primera definición de inteligencia y tests basados en la teoría de información algorítmica [Dowe & Hajek 1997] [Hernandez-Orallo 2000a] [Legg & Hutter 2007], nos enfrentamos al reto de construir el primer test de inteligencia universal, formal, pero al mismo tiempo *práctico*. La cuestión principal es la noción de test "anytime", el cual permitirá una convergencia rápida del test al nivel de inteligencia del sujeto y una evaluación progresivamente mejor cuanto más tiempo le proporcionemos. Si tenemos éxito, la ciencia será capaz de medir la inteligencia de animales superiores (p. ej. simios) humanos y máquinas de una manera universal y práctica.

Dentro del proyecto de investigación anterior existen diferentes tareas de implementación a las cuales este proyecto de fin de carrera se circunscribe, tal y como se especifica en los objetivos que veremos a continuación.

## 1.3. Objetivos

Podemos distinguir un objetivo general, a partir del cual detallamos una serie de objetivos específicos.

Objetivo general

Desarrollar un entorno de aplicación de tests de inteligencia a diferentes agentes (máquinas o personas) usando la teoría de medición desarrollada en [Hernandez-Orallo & Dowe 2010] y la clase de entornos desarrollada en [Hernández-Orallo 2010b]. Básicamente esto significa la construcción de un sistema en donde distintos tipos de agentes sean capaces de interactuar con entornos autogenerados en donde, en base a observaciones facilitadas por el sistema y acciones respondidas por los agentes, pueda ser medida la inteligencia de estos agentes.

Objetivos específicos

- Intérprete de entornos: La construcción de un sistema que permita la interacción entre unos agentes en un entorno y, posteriormente, la medición de la inteligencia demostrada por el agente.
- Codificación manual de los entornos: Permitir la construcción de entornos utilizando un mecanismo manual que permita su definición interna y su comportamiento.
- Generación automática de los entornos siguiendo alguna distribución: El sistema deberá permitir la construcción de entornos automáticamente utilizando una distribución para decidir su representación y su posterior construcción en base a dicha representación.
- Entorno gráfico de evaluación de los entornos y que proporcione los resultados: Se debe facilitar al usuario una interfaz con la que interactuar con el entorno, ofreciéndole lo necesario para la realización de los tests y que le muestre posteriormente los resultados obtenidos.
- Realización de pruebas y experimentos con entornos y agentes muy sencillos, y con entornos y agentes aleatorios: Pruebas para comprobar que el sistema funciona correctamente y experimentos para estudiar los resultados que distintos agentes obtienen al interactuar en algunos entornos codificados manualmente y en otros entornos generados automáticamente.

# 2. Precedentes

Existen dos tipos de tests que son de interés en esta área: tests psicométricos y tests de inteligencia en máquinas.

Los tests psicométricos [Martinez-Arias et al 2006] tienen una larga historia [Spearman 1904], son efectivos, fáciles de administrar, rápidos y muy estables cuando son usados en el mismo individuo a través del tiempo. De hecho, han proporcionado una de las definiciones de inteligencia más prácticas: "la inteligencia es lo que se mide por tests de inteligencia". Sin embargo, los tests psicométricos son antropomórficos: no pueden evaluar la inteligencia de sistemas diferentes del Homo sapiens.

Nuevos enfoques en la psicometría como el *Item Response Theory* (IRT) permiten la selección de items basándose en sus características de demanda cognitivas, proporcionando resultados para entender lo que se está midiendo y adaptando el test al nivel del individuo que se está examinando. Los items generados por la teoría cognitiva y analizados por IRT son una herramienta prometedora, pero estos modelos no son implementados completamente en las pruebas [Embretson & Mc Collam 2000].

A pesar de que estos y otros esfuerzos han intentado establecer "a priori" cómo debe ser un test de inteligencia (p. ej. [Embretson 1998]), y se han encontrado adaptaciones para distintos tipos de sujetos, en general necesitamos diferentes versiones de los tests psicométricos para evaluar a niños de diferentes edades, ya que los tests psicométricos para el Homo sapiens adulto se basan en unos conocimientos y habilidades que los niños no han adquirido todavía.

Lo mismo sucede con otros animales. Psicólogos comparativos y otros científicos en el área de la cognición comparativa normalmente diseñan tests específicos para especies diferentes. Se puede ver un ejemplo de estos tests especializados para niños y simios en [Herrmann et al 2007]. Adicionalmente, se ha demostrado que los tests psicométricos no funcionan para máquinas en su etapa de progreso actual de la inteligencia artificial [Sanghi & Dowe 2003], ya que pueden ser engañados por programas de ordenador muy simples. Pero el principal inconveniente de los tests psicométricos para evaluar otros sujetos diferentes a los humanos es que no existe una definición matemática tras ellos. Por lo tanto es difícil diseñar un test que funcione en cualquier sujeto inteligente (máquina, humano, animal, …), y no solo en el tipo específico de sujeto en el que experimentalmente hemos evaluado que funciona.

Los tests de inteligencia para máquinas han sido propuestos desde que Alan Turing [Turing 1950] introdujo el juego de imitación en 1950, actualmente conocido como el test de Turing. En este test, un sistema es considerado inteligente si es capaz de imitar a un humano (p. ej. ser indistinguible de un humano) durante un periodo de tiempo y sujeto a un diálogo (tele-texto) con uno o más jueces. Aunque todavía se acepta como referencia para comprobar si finalmente la inteligencia artificial se acerca a la inteligencia de los humanos, ha generado debate a lo largo del tiempo. Por supuesto, también se han sugerido varias variantes y alternativas internas. El test de Turing e ideas relacionadas presentan varios problemas como test de inteligencia para máquinas: el test de Turing es antropomórfico (mide la humanidad, no la inteligencia), no es gradual (no proporciona una puntuación), no es práctico (cada vez es más fácil de engañar y requiere mucho tiempo para obtener evaluaciones fiables) y requiere de un juez humano.

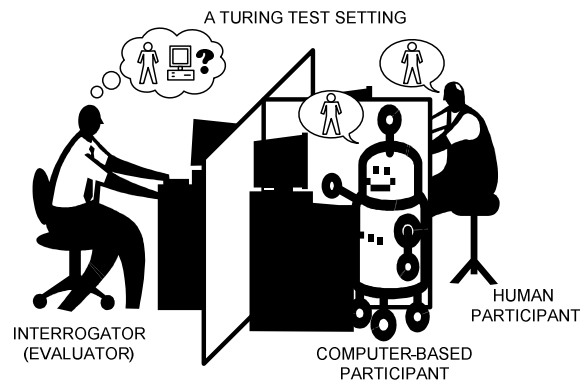

Una reciente e ingenua aproximación a los tests de inteligencia para máquinas son los llamados CAPTCHAs (Completely Automated Public Turing test to tell Computers and Humans Apart) [von Ahn et al 2002] [von Ahn et al 2008]. Los CAPTCHAs son cualquier tipo de preguntas simples que puedan ser fácilmente resueltas por un humano pero no por las tecnologías de inteligencia artificial actuales. Los CAPTCHAs típicos son los problemas de reconocimiento de caracteres donde las letras aparecen distorsionadas. Estas distorsiones hacen que para las máquinas (robots) sea difícil reconocer las letras. A continuación podemos ver un ejemplo de CAPTCHA procedente de Gmail:

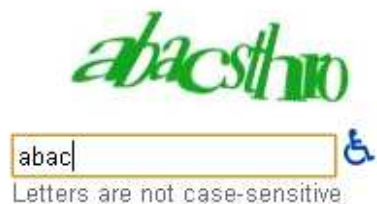

El objetivo inmediato de un CAPTCHA es separar a los humanos y a las máquinas. El objetivo final es prevenir que robots u otro tipo de agentes o programas sean capaces de crear miles de cuentas (u otras tareas que solo los humanos deberían hacer). Nótese que los robots podrían bloquear o dañar muchos servicios de Internet que utilizamos diariamente si esta u otras técnicas de control no existieran.

El problema de los CAPTCHAs es que se están haciendo cada vez más y más difíciles para los humanos, ya que los robots se están especializando y mejorando para poder leerlos. Siempre que una nueva técnica CAPTCHA es desarrollada, aparecen nuevos robots que tienen posibilidades de pasar el test. Esto fuerza a los desarrolladores de CAPTCHAs a que vuelvan a cambiarlos, y así sucesivamente. La razón es que son específicos y se basan en algunas tareas particulares. Aunque los CAPTCHAs funcionan razonablemente bien en la actualidad, dentro de 10 ó 20 años, se tendrán que hacer las cosas tan difíciles y generales, que los humanos necesitarán más tiempo y varios intentos para pasarlos. De hecho, ya está sucediendo, y perdemos cada vez más y más tiempo en los CAPTCHAs cada día.

A parte de los tests, ciertas aproximaciones más teóricas y formales a la inteligencia de las máquinas han sido llevadas a cabo por científicos prominentes en el siglo 20 como A.M. Turing, R.J Solomonoff, E.M. Gold, C.S. Wallace, J.J. Rissanen, M. Blum, G.J. Chaitin y otros. El hito es el desarrollo de la teoría de información algorítmica (también conocida como Complejidad Kolmogorov) (ver [Li & Vitanyi 2008] para una referencia más completa), su relación con el aprendizaje (inferencia inductiva y predicción) [Solomonoff 1964] [Wallace & Boulton 1968] [Solomonoff 1986] [Wallace & Dowe 1999] [Wallace 2005] y finalmente su relación con la inteligencia.

Estas ideas llevaron a algunos miembros del equipo del proyecto de investigación ANYNT a introducir varias definiciones formales a la inteligencia, a saber, las obras [Dowe & Hajek 1997, 1998] [Hernandez-Orallo & Minaya-Collado 1998] [Hernandez-Orallo 2000a] [Hernandez-Orallo 2000b]. Todos estos tests y definiciones son matemáticas, no antropomórficas (p. ej. universales), significativos e intuitivos. En este sentido no toman al Homo sapiens como referencia ni juez (como en el test de Turing), no han evolucionado a través de prueba y error a través de la experimentación en tests sobre sujetos durante un siglo (como generalmente se da en psicometría) sino que son construidos sobre nociones fundamentales y matemáticas en la teoría de computación. Y algunos de ellos lograron que fueran prácticos (un test de inteligencia factible) a coste de ser parcial (en el sentido de que miden rasgos necesarios de la inteligencia, pero no todos ellos). Veamos a continuación estas aproximaciones en mayor detalle:

Por un lado, Dowe & Hajek [Dowe & Hajek 1997, 1998] sugirieron la introducción de problemas de inferencia de inducción en los tests de Turing para, entre otras cosas, hacer frente a la paradoja de Searle "La habitación China", y también porque una habilidad de inferencia inductiva es un requisito necesario (aunque posiblemente "no suficiente") para la inteligencia.

Por otro lado, al mismo tiempo y de manera similar, y también de forma independiente, en [Hernandez-Orallo & Minaya-Collado 1998] [Hernandez-Orallo 2000a] la inteligencia se definió como la habilidad de comprender, dando una definición formal de la noción de comprensión como la identificación de un patrón 'predominante' de una evidencia dada, derivada de la teoría de inducción de Solomonoff, la complejidad de Kolmogorov y la búsqueda óptima de Levin. La definición es el resultado de un test, llamado C-test [Hernandez-Orallo & Minaya-Collado 1998] formado por series de complejidad creciente computacionalmente obtenidas. Las secuencias fueron formateadas y presentadas de una manera bastante similar a las pruebas psicométricas, y como resultado, el test fue administrado a humanos. No obstante, el objetivo principal fue que el test pudiese llegar a ser administrado a otros tipos de sistemas inteligentes.

En [Hernandez-Orallo 2000b] se esboza una factorización (y por lo tanto extensión) de estos tests de inferencia inductiva para explorar qué otras habilidades podrían conformar un completo (y por lo tanto suficiente) test. Con el fin de aplicar el test para sistemas inteligentes inferiores, (todavía) incapaces de entender el lenguaje natural, el propósito de una extensión dinámica del C-test en [Hernandez-Orallo 2000a] se expresó así: "La presentación del test debería cambiar ligeramente. Los ejercicios deberán proporcionarse uno por uno y, después de cada respuesta, se le deberá proporcionar la respuesta correcta al sujeto (deberán usarse recompensas y castigos)".

Algunos trabajos recientes realizados por Legg y Hutter (p. ej. [Legg & Hutter 2007]) han seguido los pasos anteriores y, fuertemente influenciados por la teoría de Hutter de los agentes óptimos AIXI [Hutter 2005], han dado otra definición de inteligencia de máquina, denominada "Inteligencia Universal", también surgida de la complejidad de Kolmogorov y la inducción de Solomonoff.

La idea básica es evaluar la inteligencia de un agente π en varios entornos μ, elegidos utilizando una distribución universal (derivadas de la complejidad de Kolmogorov, p. ej. $p(\mu) = 2^{-K(\mu)}$), utilizando recompensas que se acumulan durante la interacción con el entorno. Por tanto la inteligencia se define como la competencia de un agente en distintos entornos, donde los entornos simples tienen una mayor probabilidad que los entornos complejos.

La comparación con los trabajos de Hernández-Orallo (pero también el test de compresión de Dowe & Hajek) se resumen en [Legg & Hutter 2007] con la siguiente tabla:

|  | **Agente universal** | **Test universal** |
|---|---|---|
| **Entorno pasivo** | Inducción de Solomonoff | C-test |
| **Entorno activo** | AIXI | Inteligencia Universal |

De hecho, la definición basada en el C-test puede ser considerada una instancia del trabajo de Legg y Hutter ya que el agente no tiene permitido hacer una acción hasta que ha visto un número de observaciones (la secuencia de inferencia inductiva). Una de las contribuciones más relevantes en el trabajo de Legs y Hutter es que su definición de *Inteligencia Universal* permite evaluar formalmente el funcionamiento teórico de algunos agentes: un agente aleatorio, un agente especializado, … o un agente super-inteligente, como AIXI [Hutter 2005] que según se afirma, si alguna vez se construye, sacaría la mejor puntuación en el test de Inteligencia Universal. En pocas palabras, el trabajo de Legg y Hutter es otro paso que ayuda a dar forma a las cosas que se deben abordar en un futuro próximo a fin de alcanzar finalmente una teoría de la inteligencia de máquina.

Sin embargo, existen cinco problemas que hemos identificado en su definición que evitan que se ponga en práctica. Primero, tenemos una suma infinita de todos los entornos. Segundo, también tenemos una suma infinita de todas las posibles recompensas (la vida de los agentes en cada entornos es infinita). Tercero, K() no es computable. Cuatro, y más importante, no se tiene en cuenta el tiempo. Y, quinto, hay cierta confusión entre la inducción y la predicción.

# 3. Marco conceptual

## 3.1. Requisitos

En el proyecto de investigación se propone una modificación y extensión de las definiciones y tests previos a fin de construir un primer test y definición general, formal y factible. La definición y el test se basa en desarrollos previos en tests basados en la complejidad de Kolmogorov [Dowe & Hajek 1997, 1998] [Hernandez-Orallo & Minaya-Collado 1998] [Hernandez-Orallo 2000a, 2000b] [Legg & Hutter 2007]. La principal idea de todos ellos es que se usa la **Distribución Universal** para la generación de preguntas y entornos, y por lo tanto cualquier sesgo particular hacia un individuo específico, especies o culturas se evita. Esto hace al test universal para cualquier posible tipo de sujeto. Pero aparte de esta condición "Universal" nos centramos en algunos requisitos prácticos adicionales:

- Debe permitir medir cualquier tipo de sistema inteligente (biológico o computacional) que exista actualmente o pueda ser construido en el futuro (sistemas *anytime*).

- El test debe adaptarse rápidamente al nivel de inteligencia y escala de tiempo del sistema. Debe permitir evaluar tanto sistemas ineptos como brillantes (cualquier nivel de inteligencia) así como a sistemas muy lentos y muy rápidos (escala *any time*).

- La calidad de la evaluación dependerá del tiempo que dejemos al test. Esto significa que el test puede ser interrumpido en cualquier momento, produciendo una aproximación a la puntuación de la inteligencia. Cuanto más tiempo dejemos para realizar el test, mejores evaluaciones (test *anytime*).

Debido a estos requisitos llamaremos a los tests "**anytime universal intelligence tests**". Si tenemos éxito, la ciencia será capaz de medir la inteligencia de animales superiores (p. ej. simios), humanos, máquinas, híbridos o comunidades de humanos y máquinas e incluso seres extraterrestres, de un modo absoluto.

La principal dificultad para hacerlo factible es que cuanto más general tratamos que sea el test, menos cosas del agente debemos asumir. Esto significa que es posible que necesitemos más tiempo para evaluar la inteligencia, ya que no podemos confiar en un conocimiento común ni un lenguaje común para explicar las instrucciones ni dar nada por sentado. El problema es similar para evaluar niños o animales, donde todo debe ser muy simbólico y simple, y los sujetos deben ser dirigidos hacia el objetivo a través de recompensas y castigos. Esta es la razón por la que estos tests son más interactivos que los tests psicométricos tradicionales para los Homo sapiens adultos.

De hecho, existe una tesis común en todas las pruebas de inteligencia (tanto de la psicometría, la cognición comparativa o la inteligencia artificial): *el tiempo que se requiere para evaluar la inteligencia de un sujeto depende (1) del conocimiento o características que se esperan y conocen sobre el sujeto (misma especie, misma cultura, mismo lenguaje,* *etc.) y (2) en la adaptabilidad del examinador*. Mientras que (1) se lleva al extremo en psicometría humana, donde los tests son simplemente formularios con preguntas explicadas en lenguaje natural, (2) se lleva al extremo en el Test de Turing o cualquier tipo de entrevista de evaluación, donde los examinadores son humanos los cuales dinámicamente adaptan sus preguntas dependiendo de las respuestas del sujeto.

Consecuentemente, si queremos evaluar diferentes tipos de sujetos sin depender de ninguna suposición acerca de su naturaleza o conocimiento, y si queremos conseguir una valoración fiable en un reducido (práctico) periodo de tiempo, debemos necesariamente diseñar tests interactivos y adaptables.

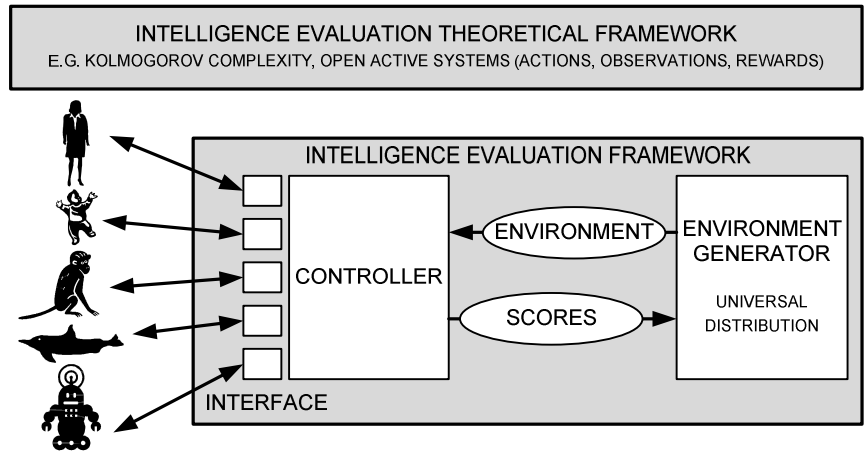

## 3.2. Entornos y agentes

Aunque muy diferentes, existen tres dimensiones donde cualquier planteamiento de medición de inteligencia tiene que tomar decisiones. En todos los ajustes de evaluación de inteligencia, encontramos (1) un sujeto o agente a examinar, (2) un conjunto de problemas, tareas o entornos y (3) un protocolo para la aplicación de medida y la derivación de uno o más resultados de rendimiento. Dependiendo de los supuestos o limitaciones impuestas a cada uno de estos temas, tenemos, como resultado, un marco de evaluación diferente. La literatura de la inteligencia artificial se ha centrado en (1) y (2), especialmente en el área de las arquitecturas cognitivas, donde implícitamente o explícitamente un conjunto de restricciones o requisitos se establece en los agentes y entornos.

Por ejemplo, SOAR [Laird et al 1987][Laird 2008], una de las más exitosas y conocidas arquitecturas cognitivas en la inteligencia artificial, implícitamente asume algunas características sobre los agentes y entornos. Algunos de estos supuestos se convierten en requisitos explícitos en [Laird & Wray 2010], donde la relación entre los requisitos de la inteligencia y los requisitos en la arquitectura cognitiva son dilucidados. Más precisamente, este trabajo describe ocho características de los entornos, tareas y agentes los cuales son considerados importantes para el nivel de inteligencia de los humanos, de los cuales se derivan doce requisitos para las arquitecturas, p. ej. para construir agentes inteligentes. De todas formas, las características de los entornos, tareas y agentes son una buena cuenta de los requisitos en los agentes y entornos, los cuales han ido apareciendo en la literatura en los últimos quince años. Las características pueden ser resumidas del siguiente modo: los entornos deben ser dinámicos, reactivos y complejos, conteniendo diversos objetos interactivos, y también algunos otros agentes que afecten al rendimiento. Las tareas deben ser complejas, diversas y novedosas. Debe haber regularidades en múltiples escalas de tiempo. La percepción de los agentes es limitada, y el entorno puede ser parcialmente observable. Los agentes deben ser capaces de responder rápidamente a la dinámica del entorno pero los recursos computacionales de los agentes deben considerarse limitados. Finalmente, la existencia de un agente es a largo plazo y continua, por lo que debe equilibrar las tareas o recompensas inmediatas con más objetivos a largo plazo. Podemos ver que ninguno de los enfoques de medición vistos en la subsección anterior sigue estos requisitos.

La noción de agente es hoy en día la corriente principal de la inteligencia artificial y, aparte de los requisitos previos (u otros), no merece una mayor aclaración. La distinción entre objetivo, tarea y entorno es una cuestión más compleja, ya que depende de la intención y la voluntad del agente. En la psicometría para niños y la cognición comparativa de animales, los tests de inteligencia no pueden asumir que seamos capaces de programar algunos objetivos o explicar explícitamente una tarea al examinado. Consecuentemente, las pruebas, se realizan generalmente utilizando recompensas, un enfoque acondicionado para que los sujetos se centren indirectamente en la tarea. Curiosamente, este es el mismo criterio adoptado en el *aprendizaje por refuerzo* [Sutton & Barto 1998]. A pesar de que el *aprendizaje por refuerzo* generalmente se ve como una formalización de este ajuste en el contexto de la inteligencia artificial y el aprendizaje de máquinas, también se puede ver con una perspectiva más amplia dado a que el aprendizaje por refuerzo es el estudio de cómo los animales y los sistemas artificiales optimizan su comportamiento condicionado por recompensas y castigos dados.

La idea más general de este ajuste es la interacción entre un agente y un entorno a través de acciones, recompensas y observaciones, el cual es también similar a la configuración que típicamente se usa en control o teoría de sistemas, y puede ser esbozado como se ve en la Figura 1:

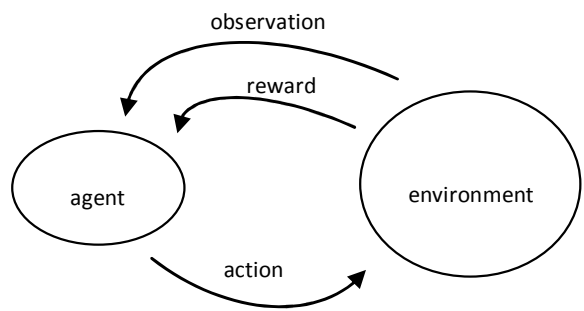

*Figura 1.Interacción entre un agente y un entorno [Legg & Hutter 2007].*

Las acciones son limitadas por un conjunto finito de símbolos $A$, (p. ej. {*izquierda, derecha, arriba, abajo*}), las recompensas se recogen de un subconjunto $R$ de números racionales entre 0 y 1, y las observaciones son también limitadas por un conjunto finito $O$ de posibilidades (p. ej. una rejilla de celdas binarias de $n \times m$ y los objetos situados en ella, o un conjunto de diodos emisores de luz, LEDs). Utilizaremos $a_i$, $r_i$ y $o_i$ para (respectivamente) denotar acciones, recompensas y observaciones en la interacción o ciclo *i* (o estado). Las recompensas y las observaciones son las salidas del entorno. El par <$r_i$, $o_i$> es también conocido como una percepción. El orden de los eventos es siempre: recompensa, observación y acción. Una secuencia de estos eventos es entonces una cadena como $r_1o_1a_1r_2o_2a_2$. Tanto el agente como el entorno se definen como una medida probabilística. Por ejemplo, dado un agente, denotado como π, el termino π($a_k$ | $r_1o_1a_1r_2o_2a_2 \ldots r_ko_k$) denota la probabilidad de que el agente π ejecute la acción $a_k$ después de la secuencia de eventos $r_1o_1a_1r_2o_2a_2 \ldots r_ko_k$. De forma similar, un entorno μ también es una medida probabilística la cual asigna probabilidades para cada posible par de observaciones y recompensas. Como μ($r_ko_k$ | $r_1o_1a_1r_2o_2a_2 \ldots r_{k-1}o_{k-1}a_{k-1}$), por ejemplo, denota la probabilidad en el entorno μ de proporcionar $r_ko_k$ después de la secuencia de eventos $r_1o_1a_1r_2o_2a_2 \ldots r_{k-1}o_{k-1}a_{k-1}$. Nótese que si para todos los ciclos existe una única acción/percepción con probabilidad 1 (y 0 para el resto), entonces tenemos un agente/entorno determinista. Si combinamos la medición probabilística para el agente y el entorno tenemos una medida probabilística para el historial de interacción o secuencia. Los historiales de interacción serán deterministas (respectivamente, computables) si tanto el agente y el entorno son deterministas (respectivamente, computables). Denotaremos con $a_i^{\mu,\pi}$, $r_i^{\mu,\pi}$ y $o_i^{\mu,\pi}$, la acción, la recompensa y la observación en la interacción o ciclo *i* para el entorno μ y el agente π.

> EJEMPLO 1.
>
> Considera el escenario de un test donde un chimpancé (el agente) puede presionar uno de tres posibles botones (A = {$B_1$, $B_2$, $B_3$}), las recompensas son solo la entrega (o no) de una banana (R = {0, 1}) y la observación son tres celdas donde una pelota debe estar exactamente dentro de una de ellas (O = {$C_1$, $C_2$, $C_3$}). Un ejemplo de un posible entorno es:
>
> μ($r_ko_k$ | $r_1o_1a_1r_2o_2a_2 \ldots r_{k-1}o_{k-1}a_{k-1}$) = 1 si (($a_{k-1}$ = $B_1$ y $o_{k-1}$ = $C_1$) o ($a_{k-1}$ = $B_2$ y $o_{k-1}$ = $C_2$) o ($a_{k-1}$ = $B_3$ y $o_{k-1}$ = $C_3$)) y ($r_k$ = +1)
>
> μ($r_ko_k$ | $r_1o_1a_1r_2o_2a_2 \ldots r_{k-1}o_{k-1}a_{k-1}$) = 1 si ¬(($a_{k-1}$ = $B_1$ y $o_{k-1}$ = $C_1$) o ($a_{k-1}$ = $B_2$ y $o_{k-1}$ = $C_2$) o ($a_{k-1}$ = $B_3$ y $o_{k-1}$ = $C_3$)) y ($r_k$ = 0)
>
> μ($r_ko_k$ | $r_1o_1a_1r_2o_2a_2 \ldots r_{k-1}o_{k-1}a_{k-1}$) = 0 en cualquier otro caso.
>
> La observación $o_k$ en ambos casos es generada aleatoriamente con una distribución uniforme entre las tres posibilidades de O. La primera recompensa es 1 (empezamos el juego dándole una banana al chimpancé).
>
> De acuerdo con la definición del entorno anterior, un chimpancé que siempre seleccione el botón correspondiente a la celda donde se encuentre la pelota puntuará una recompensa de +1 (una banana) en cada ciclo. Por ejemplo, si el entorno muestra $C_2$ y el chimpancé presiona $B_2$, entonces se le recompensará con una banana.

Aunque el ejemplo anterior es muy sencillo, la configuración general que se muestra en la Figura 1, utilizando entornos, agentes, acciones, observaciones y recompensas es lo suficientemente poderoso como para representar los requisitos establecidos en [Laird &

Wray 2010], así como cualquier medida de la inteligencia establecida vista en la subsección anterior (excepto, quizás, muchos tipos de tests psicométricos como CAPTCHAs o el Test de Turing, debido a la ausencia de recompensas).

Bajo esta configuración primero tenemos que pensar en un conjunto de entornos tales que sean los suficientemente complejos como para cumplir los requisitos anteriores. O, en otras palabras, no podemos asumir ninguna restricción en los entornos y observaciones si queremos que la configuración sea lo suficientemente general para una medición de inteligencia. En particular, en el aprendizaje por refuerzo, muchas técnicas asumen que el entorno es un *Markov Decision Process* (MDP). En otros casos, sin embargo, se asume que los entornos son completamente observables, p. ej. que existe una función entre las observaciones y los estados. No podemos asumir esto, ya que muchos problemas del mundo real no son completamente observables. Este es especialmente el caso en contextos sociales donde otros individuos pueden tener vistas diferentes (y parciales) de la misma situación.

En [Legg 2008], se desarrolla una taxonomía de entornos, mientras que se distingue entre muchos tipos de entornos. Por ejemplo, entornos pasivos son aquellos en los cuales las acciones de los agentes solamente pueden afectar a las recompensas pero no a las observaciones. Una sub-categoría especial es la secuencia de predicción como se utiliza en los tests psicométricos clásicos, y también los problemas de clasificación típicos en el aprendizaje automático. Algunos otros tipos de entornos son los *Markov Decision Processes* (MDP) de orden n, donde la siguiente observación solo puede depender de las últimas *n* observaciones y las últimas *n* acciones. Se puede ver que los MDPs de orden n pueden reducirse a MDPs de primer orden (o simplemente MDPs). En este caso, es natural hablar de "estados", como muchos juegos de mesa y algunos laberintos, ya que la próxima recompensa y observación solo dependen de la observación y acción anterior. Un tipo especial de MDPs son los MDPs ergódicos, los cuales se caracterizan por poderse alcanzar cualquier posible observación (en uno o más ciclos) a partir de cualquier estado. Esto significa que en cualquier estado, las acciones del agente pueden hacerle recuperarse de una mala decisión anterior. Creemos que esto es una gran limitación y, como ha sido anteriormente mencionado y defendido por muchos (p. ej [Laird & Wray 2010]), necesitamos ser tan generales como sea posible en la clase de entornos que necesitamos considerar.

Trataremos de considerar el concepto más general (o clase) de entorno y el concepto más general de agente, a fin de permitir una prueba universal, en la forma en la que da cuenta de una variedad de contextos y aplicable a cualquier tipo de agente. Por el momento, únicamente asumiremos que los entornos son infinitos (p. ej. ninguna secuencia de acciones los hace parar) y que están basados en un modelo, p. ej., tenemos una descripción, programa o modelo tras de ellos. También asumiremos que este modelo es computable.

Aparte de las características de los entornos y agentes, cuando abordemos la cuestión de la evaluación del agente, necesitamos estudiar sin restricción alguna la distribución de las recompensas que se necesita, y, muy especialmente, cómo se agregan las recompensas. En el aprendizaje por refuerzo, se han definido muchas agregaciones de funciones o ganancias. Por ejemplo, la forma más común de evaluar el rendimiento de un agente π en un entorno μ es calcular el valor esperado de la suma de todas las recompensas, p.ej.:

DEFINICIÓN 1. RECOMPENSA ACUMULADA ESPERADA

$$V_\mu^\pi := E\left(\sum_{i=1}^{\infty} r_i^{\mu,\pi}\right)$$

Esta no es la única opción, ya que se puede plantear a qué se le da más relevancia, si a las recompensas inmediatas o a las de largo plazo, con el fin de recompensar las políticas codiciosas o de exploración. Esto se relaciona con la vida del agente (el número de interacciones permitidas) y también si existe un límite en las recompensas [Hutter 2006]. Todo esto se volverá a examinar a continuación.

## 3.3. Medición

Siguiendo la justificación de [Hernandez-Orallo & Dowe 2010], se imponen una serie de restricciones sobre las recompensas y los entornos, con el objetivo de garantizar una serie de propiedades de convergencia y de balance frente a agentes aleatorios.

La primera idea es utilizar recompensas simétricas, las cuales pueden estar en un rango entre -1 y 1, p. ej.:

DEFINICIÓN 2. RECOMPENSAS SIMÉTRICAS

$$\forall i : -1 \leq r_i \leq 1$$

Nótese que esto no impide que la recompensa acumulada en un cierto punto sea mayor que 1 o menor que -1. Por lo que, si hacemos muchas acciones, podemos tener una recompensa acumulada mayor que 1. Observando implementaciones físicas, las recompensas negativas no tienen por qué estar asociadas con castigos, lo cual se considera poco ético para individuos biológicos. Por ejemplo, si estamos evaluando un simio, las recompensas desde -1 a -1/3 podrían implicar no dar nada, desde -1/3 a 1/3 dar una pieza de fruta, y desde 1/3 a 1 dos piezas. O una recompensa negativa puede implicar eliminar una fruta adjudicada previamente.

Si nos fijamos en las recompensas simétricas, también esperamos que los entornos sean simétricos, o más precisamente, que sean balanceados en como proporcionan recompensas. Esto puede verse del siguiente modo: en un test fiable, queremos que muchos (si no todos) los entornos ofrezcan una recompensa de 0 para agentes aleatorios. La siguiente definición lo formaliza.

DEFINICIÓN 3. ENTORNO BALANCEADO

Un entorno μ está balanceado si y solo si

1) $\forall i : -1 \leq r_i \leq 1$
2) Dado un agente aleatorio π, se mantiene la siguiente igualdad:

$$V_\mu^{\pi_r} = E\left(\sum_{i=1}^{\infty} r_i^{\mu,\pi}\right) = 0$$

Esto excluye tanto a entornos hostiles como a entornos benévolos, p. ej. entornos donde realizando acciones aleatorias se conseguiría más recompensas negativas (respectivamente positivas) que positivas (respectivamente negativas). En muchos casos no es difícil de probar que un entorno particular está balanceado. Para entornos complejos, la restricción previa se puede comprobar experimentalmente. Otra aproximación es proporcionar una máquina de referencia que únicamente genere entornos balanceados.

Nótese que las modificaciones previas en las recompensas ahora nos permiten usar una media en lugar de una recompensa acumulada, es decir:

DEFINICIÓN 4. RECOMPENSA MEDIA

$$v_{\mu_j}^{\pi}(n_i) = \frac{V_{\mu_j}^{\pi}(n_i)}{n_i}$$

Y podemos calcular el valor esperado (aunque el límite no pueda existir) de la media previa, denotada por $E(v_{\mu_j}^{\pi})$, para un valor grande arbitrario de *n*. Veamos esto con un ejemplo:

> EJEMPLO 2.
>
> Considera una modificación de las características del test visto en el ejemplo 1. Un robot (el agente) puede presionar uno de tres posibles botones (*A* = {$B_1$, $B_2$, $B_3$}), las recompensas son la no entrega de ninguna banana, una banana o dos bananas (*R* = {-1, 0, 1}) y la observación son tres celdas donde una pelota blanca y otra negra deben estar dentro de una (pero distinta) celda, es decir (*O* = {0WB, 0BW, W0B, B0W, WB0, 0BW}), donde W denota que la celda tiene una pelota blanca, B denota que la celda tiene una pelota negra y 0 denota que está vacía. Un ejemplo de un posible entorno es:
>
> μ($r_k o_k$ | $r_1 o_1 a_1$ ... $r_{k-1} o_{k-1} a_{k-1}$) = 1 si ($a_{k-1}$ = $B_1$ y $o_{k-1}$ = Wxx)
>
>                          o ($a_{k-1}$ = $B_2$ y $o_{k-1}$ = xWx)
>
>                          o ($a_{k-1}$ = $B_3$ y $o_{k-1}$ = xxW) y ($r_k$ = +1)
>
> μ($r_k o_k$ | $r_1 o_1 a_1$ ... $r_{k-1} o_{k-1} a_{k-1}$) = 1 si ($a_{k-1}$ = $B_1$ y $o_{k-1}$ = Bxx)
>
>                          o ($a_{k-1}$ = $B_2$ y $o_{k-1}$ = xBx)
>
>                          o ($a_{k-1}$ = $B_3$ y $o_{k-1}$ = xxB) y ($r_k$ = −1)
>
> μ($r_k o_k$ | $r_1 o_1 a_1$ ... $r_{k-1} o_{k-1} a_{k-1}$) = 1 si ¬(($a_{k-1}$ = $B_1$ y $o_{k-1}$ = Wxx)

> o ($a_{k-1}$ = $B_2$ y $o_{k-1}$ = xWx)
> o ($a_{k-1}$ = $B_3$ y $o_{k-1}$ = xxW)) y
> ¬(($a_{k-1}$ = $B_1$ y $o_{k-1}$ = Bxx) o ($a_{k-1}$ = $B_2$ y $o_{k-1}$ = xBx)
> o ($a_{k-1}$ = $B_3$ y $o_{k-1}$ = xxB) y ($r_k$ = 0)
>
> μ( $r_k o_k$ | $r_1 o_1 a_1 ... r_{k-1} o_{k-1} a_{k-1}$) = 0 en cualquier otro caso
>
> La observación $o_k$, se genera de entre las cuatro observaciones {0WB, 0BW, W0B, WB0} de modo uniformemente aleatorio. La primera recompensa $r_1$ es 0.
>
> Un primer robot ($π_1$) tiene el comportamiento de siempre presionar el botón $B_1$, p. ej. $π_1(B_1| X)$ para toda secuencia de $X$. Consecuentemente, el rendimiento de $π_1$ en este entorno es:
>
> $$E(v_{\mu_j}^{\pi_1}) = \underset{n_i \to \infty}{E}\left(\frac{\sum_{k=1}^{n_i} r_k^{\mu,\pi_1}}{n_i}\right) = \frac{1}{2}\lim_{n_i \to \infty}\frac{n_i}{n_i} + \frac{1}{2}\lim_{n_i \to \infty}\frac{0}{n_i} = \frac{1}{2}$$
>
> Un segundo robot ($π_2$) tiene el comportamiento de siempre presionar el botón $B_2$, p.ej. $π_2(B_2| X)$ para toda secuencia de $X$. Consecuentemente, el rendimiento de $π_2$ en este entorno es:
>
> $$E(v_{\mu_j}^{\pi_2}) = \underset{n_i \to \infty}{E}\left(\frac{\sum_{k=1}^{n_i} r_k^{\mu,\pi_2}}{n_i}\right) = \frac{1}{4}\lim_{n_i \to \infty}\frac{n_i}{n_i} + \frac{1}{2}\lim_{n_i \to \infty}\frac{-n_i}{n_i} + \lim_{n_i \to \infty}\frac{0}{n_i} = -\frac{1}{4}$$
>
> Un tercer robot ($π_3$) tiene el comportamiento de siempre presionar el botón $B_3$, p. ej. $π_3(B_3| X)$ para toda secuencia de $X$. Consecuentemente, el rendimiento de $π_3$ en este entorno es:
>
> $$E(v_{\mu_j}^{\pi_3}) = \underset{n_i \to \infty}{E}\left(\frac{\sum_{k=1}^{n_i} r_k^{\mu,\pi_3}}{n_i}\right) = \frac{1}{4}\lim_{n_i \to \infty}\frac{n_i}{n_i} + \frac{1}{2}\lim_{n_i \to \infty}\frac{-n_i}{n_i} + \lim_{n_i \to \infty}\frac{0}{n_i} = -\frac{1}{4}$$
>
> Un cuarto robot ($π_4$) tiene un comportamiento aleatorio. Por lo tanto el rendimiento de $π_4$ es:
>
> $$E(v_{\mu_j}^{\pi_4}) = \underset{n_i \to \infty}{E}\left(\frac{\sum_{k=1}^{n_i} r_k^{\mu,\pi_4}}{n_i}\right) = ... = 3 \cdot \frac{1}{3}\left(\frac{1}{2}\lim_{n_i \to \infty}\frac{n_i}{n_i} + \frac{1}{4}\lim_{n_i \to \infty}\frac{-n_i}{n_i} + \frac{1}{4}\lim_{n_i \to \infty}\frac{-n_i}{n_i}\right) = 0$$
>
> Consecuentemente, el agente $π_1$ es mejor que el aleatorio ($π_4$) en este entorno, y $π_2$ y $π_3$ son peores. Y, finalmente, ya que la recompensa global esperada de un agente aleatorio es 0, este entorno está balanceado.

## 3.4. Distribución universal y su aplicación a entornos

En esta sección daremos una breve introducción al área de la Teoría de Información Algorítmica y a las nociones de la complejidad de Kolmogorov, las distribuciones universales, la complejidad Kt de Levin, y su relación con las nociones de dificultad, comprensión, aleatoriedad, el principio MML, la predicción y la inferencia inductiva. Después, estudiaremos las aproximaciones que han aparecido utilizando estas nociones formales para dar definiciones matemáticas a la inteligencia o desarrollar tests de inteligencia a partir de ellos, empezando por una comprensión mejorada de los tests de Turing, el C-test y la definición de Legg y Hutter de la Inteligencia Universal.

La Teoría de Información Algorítmica es un campo en la informática que se relaciona adecuadamente con las nociones de la computación y la información. La idea clave es la noción de la Complejidad de Kolmogorov de un objeto, el cual es definido como la longitud del programa más corto *p* el cual genera una cadena dada *x* sobre una máquina *U*. Formalmente,

DEFINICIÓN 5. COMPLEJIDAD DE KOLMOGOROV

$$K_U(x) := \min_{p \text{ such that } U(p)=x} l(p)$$

Donde *l(p)* se refiere a la longitud en bits de *p* y *U(p)* se refiere al resultado de ejecutar *p* en *U*. Por ejemplo, si *U* es el lenguaje de programación Lisp y *x* = 1010101010101010, entonces $K_{List}(x)$ es la longitud en bits del programa más corto en Lisp que genera la cadena *x*. La relevancia en la elección de *U* depende mayoritariamente en el tamaño de *x*. Dado que cualquier máquina universal puede emular a otra, se mantiene que para cualquier par de máquinas *U* y *V*, existe una constante *c(U,V)*, la cual solo depende de *U* y de *V* y no depende de *x*, y así para todo *x*, $|K_U(x) - K_V(x)| \leq c(U,V)$. El valor de *c(U,V)* es relativamente pequeño para *x* suficientemente largas.

De la definición previa, podemos definir la probabilidad universal para la máquina *U* como sigue:

DEFINICIÓN 6. DISTRIBUCIÓN UNIVERSAL

$$p_U(x) = 2^{-K_U(x)}$$

que le da mayor probabilidad a objetos cuya descripción más corta es pequeña y menor probabilidad a objetos cuya descripción más corta sea larga. Cuando U es universal, la distribución es similar (hasta una diferencia constante) a la distribución universal para cualquier otra máquina universal diferente, dado que una puede emular a la otra. Teniendo en cuenta los programas como hipótesis en el lenguaje hipotético definido por la máquina, esto allana el camino a la teoría matemática de la inferencia inductiva, la cual fue desarrollada por Solomonoff [Solomonoff 1964], formalizando la navaja de Occam de manera adecuada para la predicción, al afirmar que la predicción que maximiza la probabilidad universal finalmente descubrirá cualquier regularidad en los datos, que se relaciona con la noción de la Longitud Mínima del Mensaje (MML) para la inferencia inductiva [Wallace & Boulton 1968] [Wallace & Dowe 1999] [Wallace 2005], y también se relaciona con la noción de la compresión de datos.

Las nociones de *predicción* e *inducción* son muy parecidas pero no idénticas, debido a que la predicción se puede obtener por una combinación (p. ej. Bayesiana) de varios modelos plausibles, mientras que la inducción normalmente se centra en descubrir el modelo más plausible y por lo general implica una explicación de las observaciones. Sin embargo, estas nociones se usan frecuentemente como sinónimos. De hecho, el papel seminal de Solomonoff [Solomonoff 1964] se refiere a "la teoría de la inferencia inductiva" cuando realmente se refiere a "la teoría de la predicción". Además, también hay importantes diferencias entre la compresión en una parte y la compresión en dos partes (inducción MML). En el primer caso, el modelo no distingue entre patrones y excepciones mientras que la segunda explícitamente separa las regularidades (patrón principal) de las excepciones. Véase [Wallace 2005] (sec. 10.1) y [Dowe 2008] (parte de sec. 0.3.1 refiriéndose a Solomonoff) para más detalles de esto.

Uno de los principales problemas de la Teoría de Información Algorítmica es que la Complejidad Kolmogorov es incomputable. Una solución popular al problema de la computabilidad de *K()* para cadenas finitas es utilizar una versión de tiempo limitado o ponderada de la complejidad de Kolmogorov (y por lo tanto de la distribución universal de la que se deriva). Una elección popular es la complejidad Kt de Levin [Levin 1973] [Li & Vitanyi 2008]:

DEFINICIÓN 7. COMPLEJIDAD KT DE LEVIN

$$Kt_U(x) := \min_{p \text{ such that } U(p)=x} \{l(p) + \log time(U, p, x)\}$$

Donde *l(p)* denota la longitud en bits de *p*, *U(p)* denota el resultado de ejecutar *p* en *U* y *time(U,p,x)* denota el tiempo que *U* utiliza ejecutando *p* para producir *x*.

Finalmente, a pesar de la incomputabilidad de K y la complejidad computacional de sus aproximaciones, han habido algunos esfuerzos en utilizar la Teoría de Información Algorítmica para diseñar una búsqueda óptima o estrategias de aprendizaje. La búsqueda de Levin (o universal) [Levin 1973] es un algoritmo de búsqueda iterativo para resolver problemas de inversión basados en Kt, el cual ha inspirado a otras estrategias de agente general como AIXI de Hutter, un agente que es capaz de adaptarse óptimamente en algunos entornos [Hutter 2007], para los que existe una aproximación de trabajo [Veness et al 2009].

En [Hernandez-Orallo & Dowe 2010] se introduce una variante de la complejidad anterior que permite asegurar que las interacciones terminan en un tiempo corto, haciendo por tanto factible la medición usando la distribución que se deriva de ella.

La aproximación considera un tiempo máximo para cada salida. Primero definimos $\Delta ctime(U, p, i)$ como el tiempo necesario para imprimir el par <$r_i,o_i$> tras la acción $a_{i-1}$, es decir el tiempo de ciclo de respuesta. A partir de aquí podemos establecer el límite superior para el tiempo de cómputo máximo que el entorno puede consumir para generar la recompensa y la observación después de la acción del agente.

DEFINICIÓN 8. COMPLEJIDAD KT PONDERANDO LOS PASOS DE LA INTERACCIÓN

$$Kt^{\max}_U(x,n) := \min_{p \text{ such that } U(p)=x} \left\{ l(p) + \log\left( \max_{i \leq n}(\Delta ctime(U, p, i)) \right) \right\}$$

lo que significa la suma de la longitud del entorno más el logaritmo del máximo tiempo de respuesta de este entorno con la máquina U. Nótese que este límite superior puede usarse en la implementación de entornos, especialmente para hacer su generación computable. Para hacer esto, ya que son infinitos, definimos su complejidad para un número límite de ciclos *n*, haciendo su definición computable. Este límite *n* no es solo necesario para la computación; también es práctico en algunos otros casos donde la computabilidad no es un problema pero no existe un máximo. Por ejemplo, considera un entorno cuya salida i-ésima dependa del cálculo de si el número *i* es primo o no. En este caso, el máximo *Δctime* no está delimitado y, por tanto el $Kt^{max}_U$ de esta secuencia sería infinito. Por lo tanto, el entorno sería descartado si no establecemos un límite de *n*.

La función de complejidad previa asegura que el tiempo de respuesta en cualquier interacción con un entorno está delimitado, pero aún conservamos la navaja de Occam en la probabilidad derivada.

## 3.5. Entornos discriminativos: Sensibilidad a las recompensas

Además, muchos entornos (tanto simples como complejos) serán completamente inútiles para evaluar la inteligencia, como entornos que dejan de interactuar, o entornos con recompensas constantes, o entornos que son muy similares a otros entornos usados anteriormente, etc. Incluyendo algunos, o la mayoría, de ellos en la muestra de entornos es una pérdida de recursos de testeo; si pudiéramos hacer una muestra más precisa seríamos capaces de hacer una evaluación más eficiente. La cuestión es determinar un criterio no arbitrario para excluir algunos entornos. Por ejemplo, la definición de Legg y Hutter [Legg & Hutter 2007] fuerza a que los entornos interactúen infinitamente, y puesto que la descripción debe ser finita, debe existir algún patrón, que puede ser eventualmente aprendido (o no) por el examinado. Pero esto incluye obviamente entornos que "siempre producen la misma observación y recompensa". De hecho, no son solo posibles sino altamente probables en muchas máquinas de referencia. Otro caso patológico es un entorno cuyas "observaciones y recompensas producidas sean aleatorias", pero esto tiene una alta complejidad, si suponemos entornos deterministas. En ambos casos el comportamiento de cualquier agente en estos entornos casi sería el mismo. En otras palabras, no tendrían *poder discriminativo*. Así que estos entornos serían inútiles para discriminar entre agentes.

En un entorno interactivo, un requisito claro para que un entorno sea discriminativo es que lo que haga el agente debe tener consecuencias en las recompensas. Sin ninguna restricción, algunos (la mayoría) entornos simples serían completamente insensibles a las acciones de los agentes. Como se ha mencionado antes, en [Legg 2008], se ha desarrollado una taxonomía de entornos, y se presenta el concepto de MDPs *ergódicos* en los que siempre se puede volver a un estado anterior. Los MDP ergódicos son una restricción muy importante, mientras que muchos entornos reales no nos dan una "segunda oportunidad". Si las "segundas oportunidades" estuvieran siempre disponibles, el comportamiento de los agentes tendería a ser más impetuoso y menos reflexivo. Además, parece más fácil aprender y tener éxito en esta clase de entornos que en una clase general.

En lugar de eso, vamos a restringir los entornos para que sean sensibles a las acciones de los agentes. Esto significa que una acción equivocada (p. ej. Ir por una puerta equivocada) podría llevar al agente a una parte del entorno desde donde nunca podrá volver (no-ergódico), pero al menos las acciones hechas por el agente pueden modificar las recompensas en este sub-entorno. Más precisamente, queremos que un agente sea capaz de influenciar en las recompensas en cualquier punto del sub-entorno. Esto no implica ergodicidad pero si al menos sensibilidad a las recompensas. Esto significa que no podemos alcanzar un punto en donde las recompensas se dan independientemente de lo que hagamos (un callejón sin salida). Esto se puede formalizar de esta manera:

DEFINICIÓN 9. ENTORNO SENSIBLE A LAS RECOMPENSAS
Dado un entorno determinista µ, decimos que es n-acciones sensible a las recompensas si para toda secuencia de acciones $a_1a_2...a_3$ de longitud $k$ existe un entorno positivo $m \leq n$ tal que existen dos secuencias de acciones $b_1b_2...b_m$ y $c_1c_2...c_m$ cuya suma de recompensas que se obtiene con la secuencia de acciones $a_1a_2...a_kb_1b_2...b_m$ es diferente a la suma de recompensas de la secuencia $a_1a_2...a_kc_1c_2...c_m$.

Nótese que la definición anterior no significa que cualquier acción tiene un impacto en las recompensas (inmediata o posteriormente), pero sí que en cualquier punto siempre hay al menos dos secuencias de acciones diferentes que puede llevar al agente a obtener recompensas acumuladas diferentes para *n* interacciones. Esto significa que estos entornos pueden tener un agente pegado por un tiempo (en un "agujero") si no se realizan las acciones buenas, pero existe un camino para salir de ahí o al menos para encontrar recompensas diferentes dentro del agujero. En otras palabras, no existen puntos cielo/infierno ni tienen un comportamiento "observador", así que en cualquier punto el agente puede esforzarse para incrementar sus recompensas (o impedir que decrezcan).

Según las definiciones anteriores, muchos juegos de mesa que conocemos no son entornos sensibles a las recompensas. Por ejemplo, existen posiciones en el juego de las damas donde inevitablemente cualquier movimiento conduce a una distinta forma de perder/ganar (asumiendo cierta habilidad por parte del oponente o usando la solución perfecta dada por [Schaeffer 2007]). Pero también nótese que no es muy difícil modificar la puntuación para hacerla completamente sensible a las recompensas asignándole puntos a la puntuación que dependan de los movimientos y la posición antes de perder (p. ej. perder en 45 movimientos es mejor que perder en 25 movimientos).

Ahora vamos a dar a una definición más refinada de la Inteligencia Universal utilizando los entornos sensibles a las recompensas, las recompensas simétricas, el entorno balanceado y la media de las recompensas:

DEFINICIÓN 10. INTELIGENCIA UNIVERSAL (CONJUNTO FINITO DE ENTORNOS BALANCEADOS Y SENSIBLE A LAS RECOMPENSAS, NÚMERO FINITO DE INTERACCIONES, COMPLEJIDAD KTMAX) CON PUNTUACIÓN AJUSTADA.

$$\Upsilon^{IV}(\pi, U, m, n_i) := \frac{1}{m \cdot n_i} \sum_{\mu \in S} V^{\pi}_{\mu_j}(n_i)$$ donde $S$ es un subconjunto finito de $m$ entornos balanceados siendo ambos $n_r$-acciones sensible a las recompensas (con $n_r = n_i$) extraído con $p^t{}_U(\mu) = 2^{-Kt^{\max}{}_U(\mu, n_i)}$

## 3.6. Un entorno no sesgado y balanceado

En esta sección presentamos una clase de entornos apropiada para la evaluación siguiendo las propiedades de la sección anterior.

### 3.6.1. Clases de entornos Turing-completas

Definir cualquier clase de entorno y construir un generador de entornos a partir de ella es fácil. De hecho, hay muchos entornos y generadores de juegos, los cuales construyen diferentes campos de juegos siguiendo algunas metareglas básicas (p. ej. [Pell 1994]). Una cuestión diferente es si queremos seleccionar una clase de entorno que sea universal (p. ej. Turing-completo) e imparcial (no particularmente fácil para algunos tipos de agentes y difíciles para otros). El problema es especialmente incómodo ya que no solo estamos hablando sobre un lenguaje que sea capaz de expresar estos entornos, sino un lenguaje que pueda usarse para automáticamente generar dichos entornos. Piénsese, por ejemplo, que podemos generar aleatoriamente programas en cualquier lenguaje de programación, pero la probabilidad de que alguno de estos programas tenga algún sentido es ridícula.

El tipo y expresividad de entornos es un problema típico en inteligencia artificial. Por ejemplo, [Weyns, Parunak & Michel 2005] es un estudio de entornos usados en sistemas multiagente. En este estudio, podemos ver muchas formas específicas en las cuales las acciones y las observaciones pueden expresarse.

Radicalmente opuesto es ir a lo más general posible y utilizar las propias máquinas de Turing. Esta aproximación es demasiado general en muchos sentidos: haciendo un filtro de tal manera que cada computación entre estados sea finita es un problema difícil (más precisamente, indecidible). Pero si restringimos la clase de entorno a una clase de entorno no universal (o más drásticamente a clases de entornos finitas), no solo podemos tener una gran parcialidad sino también podemos tener algunas otras propiedades indeseables.

Aparte de las máquinas de Turing (o variantes) hay muchos otros modelos de computación Turing-completos. De hecho, en [Hernandez-Orallo & Minaya-Collado 1998] y [Hernandez-Orallo 2000a], para medir la habilidad inductiva secuencial, usamos un tipo de máquina de registro.

Por ejemplo, los algoritmos de Markov [Markov 1960] son formalismos Turing-completos que pueden fácilmente usarse para representar entornos (véase una definición de estos algoritmos y su uso para codificar entornos y agentes en [Hernandez-Orallo 2010b]).

El problema de utilizar un lenguaje universal como un algoritmo de Markov es que muchos conjuntos de producción no construyen un entorno válido. Pueden producir una secuencia inválida de recompensas, observaciones y acciones, o secuencia alguna, o secuencias muy pequeñas (por lo que el entorno deja de interactuar muy pronto) o no terminan. Adicionalmente, considerando solo los entornos sintácticamente válidos y terminantes, hay otras restricciones impuestas por el marco [Hernandez-Orallo & Dowe 2010]: deben ser entornos sensibles a las recompensas y deben estar balanceados con respecto a los agentes aleatorios. Y finalmente, también hay una certeza de que de este modo es muy improbable generar un entorno que sea social, p. ej. que pueda contener otros agentes con los que interactuar. Como se ha tratado frecuentemente (véase p. ej. [Hernandez-Orallo & Dowe 2010]) la inteligencia de los animales (y consecuentemente de los humanos) debe haber evolucionado en entornos reales llenos de objetos y otros animales.

Nuestra aproximación será diferente. En lugar de generar entornos de máquinas universales incontroladas, definiremos un conjunto controlado, y entonces incluiremos dentro el comportamiento universal.

## 3.6.2. Acciones, observaciones y el espacio

Aparte del comportamiento de un entorno, que puede variar desde ser muy simple a muy complejo, debemos primero clarificar la *interfaz*. ¿Cuántas acciones se van a permitir? ¿Cuántas observaciones diferentes? Está claro que el mínimo número de acciones tienen que ser dos, pero no parece decidirse a priori ningún límite superior. Lo mismo ocurre con las observaciones.

Si echamos un vistazo a algunos tests de cognición para humanos y animales no-humanos, las acciones pueden ser muy variadas. Para los humanos, puede ser escribiendo una palabra o número siguiendo una secuencia. Para animales, típicamente es más reducido, con un número de botones a presionar pequeño. Las observaciones pueden tener una variedad mucho mayor. Por ejemplo, una secuencia de cuadrados y símbolos, o una secuencia de números o palabras, o incluso imágenes, son típicas en los tests de IQ. Incluso en los tests para animales, podemos tener observaciones dinámicas (rotaciones, movimientos, sonidos…). Típicamente, incluso para entornos discretos, el espacio de observaciones se hace grande por simple combinatoria.

Antes de meternos en detalles con la interfaz, tenemos que pensar en entornos que puedan contener agentes. Esto no solo sucede en la vida real, sino también es un requisito para la evolución y, por lo tanto, para la inteligencia como la conocemos. La existencia de varios agentes que pueden interactuar requiere de un *espacio*. El espacio no es necesariamente un espacio físico o virtual, sino un conjunto de reglas comunes que gobiernen lo que los agentes pueden percibir y hacer. En el mundo real, este conjunto de reglas comunes es la física.

Lo bueno de pensar en espacios es que un espacio implica la reducción de las percepciones y acciones posibles. Si definimos un espacio común, tenemos muchas elecciones sobre observaciones y acciones ya adoptadas. Una primera (y común) idea para un espacio es una rejilla 2D. Esta es la elección típica en laberintos y videojuegos 2D. A partir de una rejilla 2D, la observación es una imagen de la rejilla con los objetos y los agentes dentro. Las acciones típicas son los movimientos: izquierda, derecha, arriba y abajo. Alternativamente, por supuesto, podemos usar un espacio 3D, ya que nuestro mundo es 3D.

El problema de una rejilla 2D o 3D es que está claramente pensada a favor de los humanos y muchos otros animales que tienen habilidades para orientarse en este tipo de espacios. Otros tipos de animales o personas con discapacidades (p. ej. personas ciegas) podrían tener algunas dificultades en este tipo de espacios.

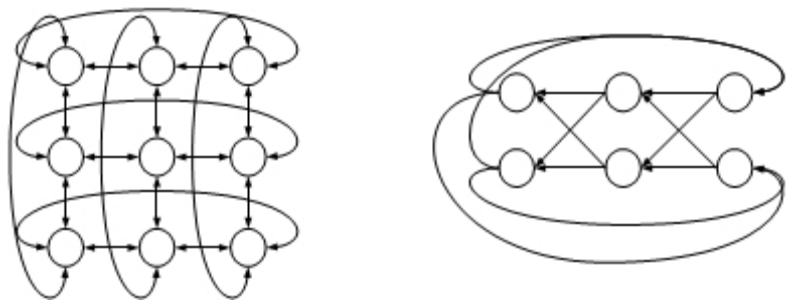

Figura 1: Dos espacios. Una rejilla toroidal y un anillo de dos niveles.

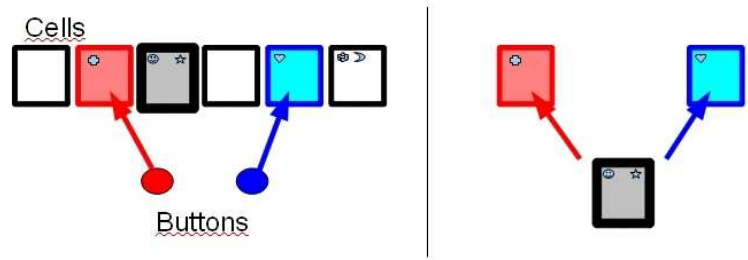

Figura 2: Interfaces posibles para un espacio adimensional. Izquierda: Se muestran todas las celdas. Derecha: Solo se muestran las celdas accesibles.

En vez de esto proponemos un tipo de espacio más general. Una rejilla 2D es un grafo con una topología muy especial, donde hay conceptos como la dirección, adyacencia, etc. Una generalización es un grafo donde las celdas están libremente conectadas a otras celdas sin ningún patrón particularmente predefinido. La Figura 1 muestra dos espacios. Nótese que la proximidad actual de dos celdas (en términos de flechas entre ellas) generalmente no corresponde con su proximidad en una representación 2D (incluso en estos espacios relativamente regulares).

Las observaciones y acciones pueden verse gráficamente, como se puede ver en la parte más a la izquierda de la Figura 2 la cual ilustra dos interfaces posibles. Nótese que las celdas no están distribuidas por su proximidad actual, sino simplemente en una línea horizontal. La celda con el borde más grueso es la celda actual, y las celdas a las que apuntan las flechas son las celdas donde el agente puede moverse. Los agentes y objetos son representados por símbolos diferentes. Dependiendo del agente (p. ej. un humano adulto o un chimpancé), esto puede convertirse en un tipo de interfaz táctil (donde las celdas pueden presionarse) o en uno más robusto (con botones). Para agentes artificiales, realmente da lo mismo.

Nótese que todas las conexiones entre las celdas no son explícitas (no vemos el grafo completo, solo la parte que está conectada a la celda actual). Podemos restringir además la observación a solo la celda actual y las celdas donde llevan las acciones (derecha de la Figura 2). Esto será problemático para algunos agentes (p. ej. niños pequeños y algunos animales) para los cuales lo que no ves no existe. Al final, hay muchas opciones para representar la interfaz y tenemos que ser muy cuidadosos para evitar sesgos.

### 3.6.3. Definición de la clase de entorno

Después de la discusión previa, estamos listos para hacer varias elecciones importantes y dar la definición de la clase de entorno. Primero debemos definir el espacio y los objetos, y entonces las observaciones, acciones y recompensas.

Con $n_a = |A| \geq 2$ representamos el número de acciones, con $n_c \geq 2$ el número de celdas, y con $n_o$ el número de objetos/agentes (sin incluir el agente que será evaluado y dos objetos especiales conocidos como *Good* y *Evil*).

**Espacio**

El espacio se define como un grafo dirigido etiquetado de $n_c$ nodos (o vértices), donde cada nodo representa una celda. Los nodos se numeran, empezando desde 1, por lo que las celdas se referencian como $C_1, C_2,..., C_{n_c}$. Desde cada celda salen $n_a$ flechas (o aristas), cada uno de ellos denota $C_i \rightarrow_\alpha C_j$, significando que la acción $\alpha \in A$ va desde $C_i$ hasta $C_j$. Al menos dos flechas han de dirigirse a diferentes celdas. Al menos una de las flechas debe llevar a la misma celda.

Un camino desde $C_i$ hasta $C_m$ es una secuencia de flechas $C_i \rightarrow C_j$, $C_j \rightarrow C_k,..., C_l \rightarrow C_m$. El grafo debe ser completamente conectado, p. ej., para cada par de celdas $C_i$, $C_j$ existe un camino de $C_i$ a $C_j$ y viceversa.

Dada la definición previa, la topología del espacio puede ser muy variada. Puede incluir una rejilla típica, y también topologías mucho más complejas. En general, el número de acciones $n_a$ es un factor que influencia mucho más a la topología que el número de celdas $n_c$.

**Objetos**

Las celdas pueden contener objetos de un conjunto predefinido de objetos Ω, donde $n_\omega = |\Omega|$. Los objetos pueden realizar acciones siguiendo las reglas del espacio, pero aparte de estas reglas, pueden tener cualquier comportamiento (determinista o no). Los objetos pueden ser reactivos a sus observaciones. Los objetos realizan una y solo una acción en cada interacción del entorno (excepto los agentes especiales *Good* y *Evil*, que pueden realizar varias acciones de golpe).

Aparte del agente evaluable $\pi$, como ya hemos mencionado, existen dos objetos especiales, llamados *Good* y *Evil*, representados por ⊕ y ⊖ respectivamente, los cuales pueden ser vistos por el agente evaluable $\pi$. Sin embargo, son indistinguibles para el resto de objetos (incluyéndose a sí mismos), por lo que para ellos en sus observaciones se representan con el mismo símbolo ⊙.

Good y Evil deben tener el *mismo* comportamiento. Esto no quiere decir que realicen los mismos movimientos, sino que tengan la misma *lógica* o *programa* tras de ellos. Nótese que *Good* y *Evil* se ven mutuamente de la misma forma (e igualmente el resto de objetos excepto $\pi$).

Los objetos pueden compartir una misma celda, excepto *Good* y *Evil*, que no pueden estar en la misma celda. Si su comportamiento les conduce a la misma celda, entonces uno (elegido aleatoriamente con la misma probabilidad) realiza el movimiento y el otro se queda en su celda original. Los objetos ⊕ y ⊖ pueden realizar varias acciones en una única interacción, p. ej. pueden realizar cualquier secuencia de acciones (no-vacía). Una razón para esto es evitar que *Good* sea seguido por el agente de una manera fácil y óptima para conseguir recompensas positivas en la mayoría de los entornos. Los objetos son colocados aleatoriamente en las celdas en la inicialización del entorno.

Aunque *Good* y *Evil* tengan el mismo comportamiento, la celda inicial la cual es asignada (aleatoriamente) a cada uno de ellos debería determinar una situación donde su comportamiento es finalmente muy asimétrico desde el punto de vista del agente $\pi$. Por ejemplo, considera el siguiente ejemplo:

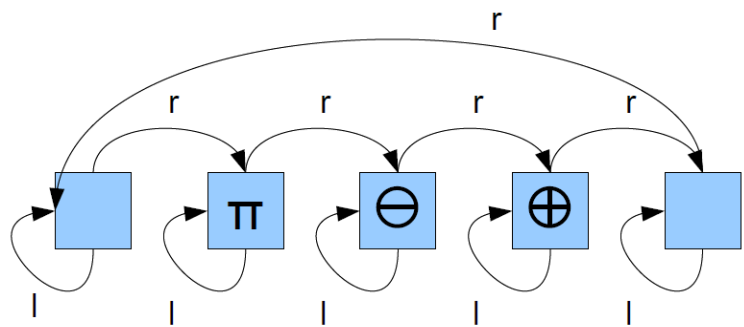

Figura 3: Un espacio anillo donde el estado inicial puede ser crítico.

**Ejemplo** Imagina el espacio visto en la Figura 3 y considera el comportamiento de ⊕ y ⊖ de forma que realizan la acción *r* si y solo si el agente π comparte una celda con cualquiera de ellos. De lo contrario, realizan la acción *l*.

Desde el estado representado en la Figura 3, está claro que la situación relativa de los tres objetos solo puede ser como se ha visto (π, ⊖, ⊕) o cuando el agente comparte una celda con ⊖ seguido por ⊕ a la derecha. Así, en este entorno, es imposible para π compartir celda con ⊕, mientras que si que es posible con ⊖, aunque ⊕ y ⊖ tienen el mismo comportamiento. El estado inicial es crítico.

Siguiendo el ejemplo anterior, podemos definir una "cláusula cíclica" que funciona como sigue. Dado un entorno con $n_a$ acciones, y $n_c$ celdas, calculamos un número aleatorio *n* entre 1 y $n_c^{\wedge}(n_a)$ (uniformemente), y entonces después de *n* interacciones, las posiciones de ⊕ y ⊖ son intercambiadas. Entonces, calculamos nuevamente otro número aleatorio (de la misma forma) y cambiamos de nuevo las posiciones. Y así sucesivamente. La razón de ser al azar es para evitar que el ciclo coincida con ningún ciclo o patrón que esté presente en el comportamiento de los agentes. El objetivo de esta cláusula es evitar la relevancia del estado inicial.

Finalmente, las primeras interacciones con el entorno pueden tener lo que llamamos "basura inicial" [Hernandez-Orallo 2009]. Considera, p. ej., un comportamiento para ⊕ y ⊖ que sea "empezar haciendo $a_1a_2a_0a_1a_1a_1a_0a_2a_2a_0a_1a_1a_0a_0$ y entonces hacer $a_0a_1$ siempre". La primera parte de su comportamiento es completamente aleatoria y completamente no-discriminativa. Solo cuando se llega al patrón (la segunda parte del comportamiento), tiene sentido empezar a evaluar el comportamiento del agente. Consecuentemente, sugerimos dejar a un agente jugar durante *n* interacciones para superar la mayor parte de la basura inicial (si existe) y entonces empezar la evaluación. El valor para *n* puede calcularse igual que el valor usado anteriormente para la "cláusula cíclica".

**Observaciones y acciones**

Una observación es una secuencia de contenido de celdas. Cada elemento en la secuencia muestra la presencia o ausencia de cada objeto, incluyendo al agente evaluable. Adicionalmente, cada celda que es alcanzable por una acción incluye la información de la acción que lleva a la celda.

En particular, el contenido de cada celda es una secuencia de objetos, donde π debe aparecer antes de ⊕ y ⊖, y del resto de objetos siguiendo sus índices. Después siguen las acciones posibles, también ordenadas por su índice, y denotadas por $A_i$ en lugar de $\alpha_i$. Cada secuencia contenida en la celda está separada por el símbolo ':'.

Por ejemplo, si tenemos un entorno con $n_a$ = 2, $n_c$ = 4 y $n_o$ = 2 entonces la siguiente secuencia **πω₂A₁ : ⊖ : ⊕ω₁A₂ :** es una posible observación vista por el agente evaluable π. El significado de esta secuencia es que en la celda 1 tenemos al agente evaluable y al objeto $\omega_2$, en la celda 2 tenemos a *Evil*, en la celda 3 tenemos a *Good* y al objeto $\omega_1$ y la celda 4 está vacía. Adicionalmente, vemos que podemos quedarnos en la celda 1 con la acción $\alpha_1$ y podemos ir a la celda 3 con la acción $\alpha_2$. La misma observación sería vista como **πω₂A₁ : ⊙ : ⊙ω₁A₂ :** por el resto de objetos (incluyendo a *Good* y *Evil*).

**Recompensas**

Trabajaremos con la noción de "rastro" y la noción de "recompensa de celda" que denotamos como $r(C_i)$. Inicialmente, $r(C_i) = 0$ para todo $i$. Las recompensas de las celdas se actualizan con los movimientos de $\oplus$ y $\ominus$. En cada interacción, ponemos $r_i^{\oplus}$ a la recompensa de la celda donde esté $\oplus$ y $-r_i^{\ominus}$ a la recompensa de la celda donde esté $\ominus$. En cada interacción, todas las demás recompensas de las celdas se dividen entre 2. Por lo que, intuitivamente podemos ver que $\oplus$ va dejando un rastro positivo y $\ominus$ va dejando un rastro negativo. El agente $\pi$ *se come* la recompensa que encuentra en la celda que ocupa, actualizando la recompensa acumulada $p = p + r(C_i)$. Con *se come* se refiere a que justo después de obtener la recompensa, la recompensa de la celda se establece a 0.

Los objetos $\oplus$ y $\ominus$ dan una recompensa de 0 si comparten una celda con $\pi$. Los valores de $r_i^{\oplus}$ y $-r_i^{\ominus}$ que dejan $\oplus$ y $\ominus$ en el resto de las ocasiones son también parte del comportamiento de $\oplus$ y $\ominus$ (lo cual es lo mismo, pero esto no significa que $r_i^{\oplus} = -r_i^{\ominus}$, para todo $i$). Solo una restricción es impuesta en cómo estos valores pueden ser generados, $\forall i : 0 < r_i^{\oplus} \leq \frac{1}{2}$ y $0 < r_i^{\ominus} \leq \frac{1}{2}$. Finalmente, nótese que las recompensas y los rastros no son parte de las observaciones, por lo que no se pueden (directamente) observar por un objeto (incluyendo a $\pi$).

A fin de mantener las recompensas acumuladas entre -1 y 1 y prevenir que $\pi$ se duerma en los laureles, al final de la evaluación, dividimos la recompensa acumulada por el número final de interacciones, con un ajuste, como se ve en [Hernandez-Orallo 2010a].

Usaremos el término $\Lambda$ para esta clase de entorno.

**Ejemplos**

Dada la configuración de la clase de entorno dada arriba, veamos algunos ejemplos dentro de $\Lambda$. Empecemos con el entorno más simple dentro de la clase.

**Ejemplo** Considera $n_a = 2 = |\{\alpha_1, \alpha_2\}|$, $n_c = 2$ y $n_o = 0$. El espacio está necesariamente compuesto de dos celdas y dos flechas, $\alpha_1$ que va a la misma celda y $\alpha_2$ que se dirige a la otra celda. El comportamiento más simple para $\oplus$ y $\ominus$ es ejecutar siempre $\alpha_1$ o $\alpha_2$, y teniendo $r_i^{\oplus} = r_i^{\ominus} = \frac{1}{2}$ para siempre. Inicialmente, Good y Evil son situados aleatoriamente en las dos celdas (sin poder situarse en la misma celda). Ya que el entorno es simétrico, considera que $\oplus$ está en la celda 1 y $\ominus$ está en la celda 2. En el caso de que el comportamiento sea $\alpha_1$ (recuerda que es el mismo para ambos $\oplus$ y $\ominus$), entonces ambos *Good* y *Evil* se quedarán para siempre en su celda. La recompensa será entonces ½ si el agente empieza en la posición donde está $\ominus$ y se mueve al lugar donde está $\oplus$, entonces recibirá recompensas de 0 ya que se mantiene donde está $\oplus$. Si el agente empieza donde está $\oplus$ entonces el movimiento a la celda donde está $\ominus$ recibirá -½ y pronto intentará volver donde está $\oplus$, recibiendo ½. Aunque la media convergerá a 0, es conveniente mantenerse donde esté $\oplus$. En el caso de que el comportamiento de $\oplus$ y $\ominus$ sea $\alpha2$, entonces ambos Good y Evil intercambiarán siempre las celdas. Consecuentemente, la mejor estrategia para el agente será moverse a la celda donde esté $\oplus$, obteniendo ½ de media.

Otros entornos más complejos pueden verse en [Hernandez-Orallo 2010b]. De hecho, con un poco de paciencia, cualquier tipo de juego o tarea que conocemos puede ser simulado utilizando esta clase de entorno, como vemos con el tres en raya en [Hernandez-Orallo 2010b].

### 3.6.4. Propiedades

En esta sección, analizamos si la clase de entorno anterior es sensible a las recompensas (el agente puede realizar acciones de forma que pueden afectar a las recompensas), y también es balanceado (un agente aleatorio tendrá una recompensa acumulada esperada igual a 0). Para la definición formal de estas propiedades, véase [Hernandez-Orallo & Dowe 2010]. Para las pruebas de las siguientes proposiciones véase [Hernandez-Orallo 2010b].

**Proposición 0.1** *Λ es sensible a las recompensas.*

**Proposición 0.2** *Λ está balanceado.*

Las proposiciones previas muestran que los entornos siguen los requisitos de un test anytime [Hernandez-Orallo & Dowe 2010].

### 3.6.5. Codificación y generación del entorno

Finalmente, para construir un test, necesitamos generar los entornos automáticamente y calcular su complejidad. Para generar un entorno, necesitamos generar el espacio (constantes y topología) y el comportamiento de todos los objetos (excepto $\pi$). La primera idea puede ser usar una gramática generativa, como de costumbre. Sin embargo, podemos elegir una gramática generativa que no sea universal pero que solo genere espacios válidos. Esto es un error. Con una gramática generativa no-universal podría ocurrir que la complejidad de un espacio con 100 celdas idénticas pueda ser 100 veces la complejidad de un espacio con una celda. Pero esto va completamente en contra de la complejidad de Kolmogorov y el ordenamiento de los entornos que estamos buscando. En lugar de esto, usaremos gramáticas generativas universales, en concreto algoritmos de Markov.

**Codificando y generando espacios**

Primero codificamos el número de acciones $n_a$ utilizando cualquier codificación estándar para los número naturales (p. ej. la función *log\** en [Rissanen 1983] [Wallace 2005]). El grafo del espacio está definido por un algoritmo de Markov sin restricciones en su definición, pero con la siguiente post-condición. El espacio generado tiene que definirse por una cadena como sigue (utilizamos una notación en lenguaje regular)

**[{+|-}$a_1^+$][{+|-}$a_2^+$]...[{+|-}$a_{na}^+$]** para cada celda. Esto significa que enumeramos todas las celdas, y la información en cada celda se compone de las flechas salientes (más precisamente a qué celdas llega por el número de veces que la acción aparece). Utilizamos un índice toroidal donde, p. ej. $1 - 2 = n_c - 1$, por lo que podemos usar referencias positivas o negativas (este es el significado de +|-). Cuando nos referimos a la misma celda se omite la acción.

El espacio de la Figura 3 se codifica con *+r+r+r+r+r* y un algoritmo de Markov que lo genera es:

1. $S \rightarrow +r$
2. $\rightarrow SSSSS\Omega$
3. $\Omega \rightarrow \cdot$

Con la descripción y definiciones anteriores no es difícil ver como *codificar* (inequívocamente) cualquier entorno válido. Sin embargo, si queremos *generar* entornos usando algoritmos de Markov, la cosa es mucho más difícil, ya que una generación aleatoria de las reglas de Markov puede generar cadenas que no representen ningún entorno. Aunque las optimizaciones pueden existir, ya que no queremos perder generalidad y manejar la medición de la complejidad, aquí proponemos dejar al algoritmo ejecutarse un número limitado de iteraciones y entonces pasarle un postprocesamiento (eliminar + repetidos, símbolos inválidos, etc.) y entonces comprobar si la cadena resultante es un entorno válido (sintáctica y semánticamente).

Para el cálculo de la complejidad, cualquier aproximación de la longitud del algoritmo de Markov (p. ej. el número de símbolos del algoritmo entero) sería válido como una aproximación a su complejidad.

**Codificando y generando objetos**

Primero codificamos el número de objetos $n_\omega$, utilizando cualquier codificación para los números naturales como se menciona arriba. Después tenemos que codificar su celda inicial, codificando también $n_o + 3$ números naturales delimitados por $n_c$ (no necesariamente diferentes, consecuentemente $\log n_c$ cada uno). En el caso de que $\oplus$ y $\ominus$ estén en la misma celda, una nueva celda es generada aleatoriamente y asignada a $\ominus$.

El comportamiento (que debe ser universal) es generado por otro algoritmo de Markov como sigue. La cadena de entrada del algoritmo es la observación, codificada como hemos visto en secciones anteriores. Solo con la observación actual los objetos no tendrían acceso a la memoria y estarían muy limitados, por lo que dos símbolos especiales $\Delta$ y $\nabla$ significan poner la cadena actual en memoria o recuperar la última cadena insertada a la memoria respectivamente. Una vez el algoritmo de Markov es ejecutado, todos los símbolos que no están en el conjunto de acciones $A$ son eliminados de la cadena generada. Entonces, para todos los objetos (excepto para $\oplus$ y $\ominus$), la acción más a la derecha en la cadena es la acción a realizar. Para $\oplus$ y $\ominus$ se utiliza toda la cadena.

Veamos un ejemplo del comportamiento para ⊕ y ⊖ (recordando que son iguales) para el espacio visto en la Figura 3 y sin considerar otros objetos:

1. $\pi[\odot]L \rightarrow \cdot rrr$
2. $\rightarrow \cdot l$

La notación [] representa una parte opcional. Consecuentemente, el algoritmo previo significa realizar siempre la acción *l* a menos que $\pi$ se encuentre inmediatamente a la izquierda. En este caso, realizamos la acción *r* tres veces. Por ejemplo, si la observación en la Figura 3 para ⊖ se representa por **: πL : ⊙ : R⊙ :**, entonces la aplicación al algoritmo previo en esta observación proporciona como resultado **: rrr⊙ : R⊙ :** y después del postprocesamiento (todos los símbolos que no están en el conjunto de acciones *A* son eliminados) tendremos ***rrr***.

Para el cálculo de la complejidad, alguna aproximación de la longitud del algoritmo (p. ej. el número de símbolos del algoritmo entero) sería válido como su complejidad.

Es importante remarcar que la complejidad de un conjunto de objetos *no* es la suma de las complejidades de los objetos. Una opción es generar un único algoritmo de Markov para todos los objetos, pero en este caso llegaríamos a la solución general (pero impracticable) de la que hemos hablado al principio. Alternativamente, sugerimos utilizar un bit extra en cada regla y entonces tenemos una forma de referirnos a otras reglas en otros objetos. Consecuentemente, si 100 objetos comparten muchas reglas, la complejidad de todo el conjunto será mucho más pequeña que la suma de las partes.

# 4. Implementación aproximada del marco conceptual

Ya que el marco conceptual es una idea general de cómo debe construirse el sistema, para comenzar a construirlo hemos decidido implementar una aproximación de este marco conceptual anteriormente descrito. Hemos implementado las características y propiedades más generales definidas en el apartado anterior y se ha optado por realizar una simplificación sobre otros aspectos más concretos de implementación sin por ello descuidar las propiedades que debe tener el sistema. La intención es poder empezar a evaluar el marco conceptual con este prototipo, para tener una primera estimación de su viabilidad.

Por una parte hemos mantenido la estructura principal: manteniendo a los agentes, entornos y espacios y las interacciones entre ellos. Por otro lado hemos simplificado otros aspectos como la construcción de las observaciones, la codificación y generación de los espacios, el comportamiento de los agentes y algunos aspectos de las interacciones entre los agentes.

En esta sección se describe cómo se ha estructurado el programa. En concreto veremos un diagrama de clases, donde queda representado a grandes rasgos la estructura interna del sistema, las principales diferencias entre el marco conceptual antes descrito y la implementación realizada, una descripción de las clases más importantes del sistema, como se han codificado y descrito los espacios y la interfaz de usuario construida para el entorno $\Lambda$.

## 4.1. Diagrama de clases

Para facilitar el entendimiento del diagrama de clases se han suprimido las funciones de las clases, ya que, de lo contrario, resultaba complicado entender su estructura.

En la siguiente imagen podemos ver el diagrama de clases simplificado del sistema.

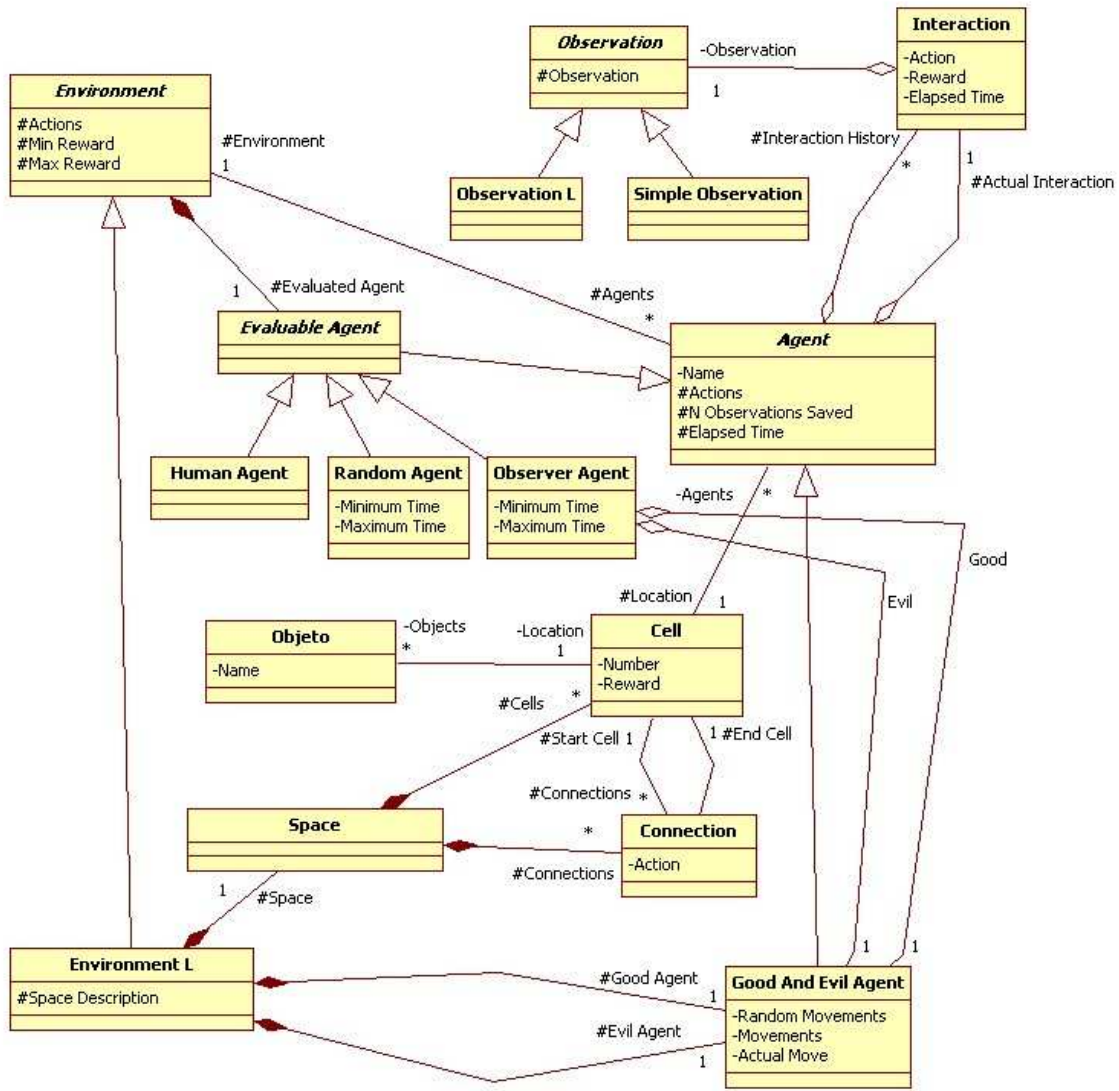

## 4.2. Principales diferencias entre el marco conceptual y la implementación realizada

### 4.2.1. Observaciones

Según el marco conceptual visto anteriormente, a los agentes se les proporcionan las observaciones como una tira de caracteres tal y como podemos ver en el apartado 3.6.3. Sin embargo, para esta primera implementación, hemos optado por realizar una copia de la estructura del espacio que lo representa dentro del programa y facilitarles a todos los agentes esta misma copia. De este modo se les facilita a todos los agentes la observación del entorno dándoles a todos una misma visión del entorno.

### 4.2.2. Objetos y agentes

Hemos realizado una distinción mayor entre los objetos y los agentes. En esta implementación hemos optado por dividir estos dos tipos de modo que los objetos sean objetos inanimados mientras que los agentes se podrán mover a través del espacio.

### 4.2.3. Comportamiento de los agentes

De momento no hemos codificado ningún lenguaje de especificación para la generación del comportamiento de los agentes. A los agentes se les indicará de forma manual cual será su comportamiento, siendo éste una decisión aleatoria, un patrón de movimientos a seguir, etc.

### 4.2.4. Interacciones entre agentes

Por un lado los agentes generadores de recompensas (*Good* y *Evil*) siempre dejan caer la misma recompensa pasen por la celda que pasen y compartan o no celda con cualquier otro agente. Sin embargo hemos optado por dividir las recompensas que los agentes recogen de las celdas entre todos los agentes que se encuentren en ese momento en la celda, de modo que si en una celda que contiene una recompensa de +1 se encuentran 4 agentes dentro de la celda, cada agente recibirá +0.25 de recompensa.

## 4.3. Descripción de las clases del sistema

En las siguientes subsecciones veremos la descripción de las clases implementadas en el sistema.

### 4.3.1. Clases para representar los agentes y objetos

#### 4.3.1.1. Agente

Para representar a las entidades que interactúan dentro del sistema se ha desarrollado la clase **Agent**. Estas entidades representan tanto a las personas que interaccionan con el sistema, a programas o máquinas que pueden interactuar automáticamente con éste, como a entidades que forman parte del sistema y, por lo tanto, son controladas directamente por éste.

Al iniciar la sesión cada agente se sitúa aleatoriamente en una celda y, en cada interacción con el sistema, éste se puede mover a las celdas adyacentes o puede decidir realizar una acción que le mantiene en la misma celda. De este modo los agentes tienen la libertad de moverse libremente a través del espacio donde realizan la sesión.

Debido a que existen varios tipos de agentes que pueden interactuar dentro del sistema, la clase **Agent** se ha especializado en varias clases hijas encargadas de proporcionar una funcionalidad especializada para cada tipo de agente.

A continuación podemos ver una tabla con los atributos y funciones de la clase.

| Atributos | Descripción |
|---|---|
| **Environment** | Entorno en el que está subscrito el agente |
| **Name** | Nombre que identifica al agente |
| **Actions** | Conjunto de acciones que puede realizar |
| **Location** | Celda en la que se encuentra |
| **NObservationsSaved** | Número de observaciones que es capaz de recordar |
| **InteractionHistory** | Historial de interacciones realizadas con el entorno |
| **Funciones** | **Descripción** |
| **SetLocation(NewCell)** | Coloca al agente en la nueva celda |
| **Act()** | Almacena y devuelve la acción realizada y almacena el tiempo tardado en realizarla |
| **Reward(Reward)** | Almacena la recompensa y termina la interacción actual |
| **Observation(Observation)** | Comienza una nueva interacción y almacena la observación |
| **Action()** | Devuelve la acción que realiza durante la interacción con el entorno |

## 4.3.1.2. Agente Evaluable (Subclase de Agente)

Para representar a los agentes que se desea que el sistema evalúe se ha creado la clase **EvaluableAgent**. Los agentes que se instancien bajo esta clase son los que el sistema evaluará.

A continuación podemos ver una tabla con las funciones de la clase.

| Funciones | Descripción |
|---|---|
| **GetTotalReward()** | Calcula y devuelve la recompensa global del agente |

## 4.3.1.3. Agentes Good & Evil (Subclase de Agente)

En todo entorno existen dos agentes llamados *Good* y *Evil*. Estos agentes son los encargados de proporcionar las recompensas en el espacio. Al comenzar cada interacción, estos agentes dejan caer recompensas en la celda en la que se encuentren en ese momento, de forma que van dejando un rastro de recompensas en el espacio. Estas recompensas son valores numéricos entre -1 y 1 y serán el objetivo de los agentes evaluables, los cuales tratarán de conseguir la mayor recompensa posible durante la sesión. Así pues, si la instancia de la clase representa al agente *Good*, éste deja caer recompensas positivas, mientras que el agente *Evil* deja caer recompensas negativas. Al finalizar cada interacción las recompensas que se encuentren por todo el entorno se dividen entre 2, disminuyendo así progresivamente su valor.

Por norma general estos agentes tratan de moverse a una celda adyacente. Como caso excepcional, ambos agentes no pueden encontrarse simultáneamente en la misma celda, por lo que si se diera la situación de que ambos agentes tratasen de moverse a la misma celda al mismo tiempo, uno de ellos no podrá realizar su última acción, quedándose así en la celda en la que se encontraba en la interacción anterior. De este modo se consigue que ambos agentes no se encuentren nunca en la misma celda.

A continuación podemos ver una tabla con los atributos y funciones de la clase.

| Atributos | Descripción |
|---|---|
| **RandomMovements** | Indica si el agente se moverá de forma aleatoria o, por el contrario, seguirá un patrón de movimientos |
| **Movements** | Patrón que contiene la secuencia de movimientos que realizará indefinidamente el agente |
| **Funciones** | **Descripción** |
| **Action()** | Devuelve la acción a realizar tanto si el agente actúa de forma aleatoria o siguiendo un patrón |

## 4.3.1.4. Agente Humano (Subclase de Agente Evaluable)

Dentro de los agentes evaluables se encuentra **HumanAgent.** Esta clase permite al usuario interactuar con el sistema. En cada interacción, el agente puede elegir de entre las acciones posibles, qué acción realizar, teniendo también la opción de permanecer en la misma celda y, por lo tanto, no moverse en esa interacción.

**Página 35**

A continuación podemos ver una tabla con las funciones de la clase.

| Funciones | Descripción |
|---|---|
| **Action()** | Espera a recibir la acción del usuario y la devuelve |

### 4.3.1.5. Agente Aleatorio (Subclase de Agente Evaluable)

Dentro de los agentes evaluables también se encuentra **RandomAgent**. Este agente es controlado por el sistema y representa a un agente que realiza acciones dentro del espacio de forma completamente aleatoria, eligiendo en cada interacción una acción cualquiera de entre todas las posibles dentro del rango de acciones permitidas.

Para controlar el rango de tiempo que tardará un agente aleatorio en realizar una acción, se le especifica el tiempo mínimo y máximo del que dispondrá para realizar dicha acción. Este tiempo será "virtual", de modo que el agente realmente no tardará este tiempo en realizar cada acción.

A continuación podemos ver una tabla con los atributos y funciones de la clase.

| Atributos | Descripción |
|---|---|
| **MinimumTime** | Indica el tiempo mínimo que tardará en realizar un movimiento |
| **MaximumTime** | Indica el tiempo máximo que tardará en realizar un movimiento |
| **Funciones** | **Descripción** |
| **Action()** | Devuelve la acción a realizar elegida aleatoriamente |

### 4.3.1.6. Agente Observador (Subclase de Agente Evaluable)

Al igual que el agente aleatorio, el **ObserverAgent** está controlado por el sistema. A diferencia del agente aleatorio, este agente puede mirar las celdas que tiene a su alrededor para decidir a cuál de ellas moverse. Para ello comprueba si están los agentes *Good* o *Evil* en alguna de estas celdas, en el caso de encontrar al agente *Good* se moverá a la celda donde éste se encuentre, de lo contrario, se moverá a cualquier otra celda adyacente siempre y cuando no se encuentre el agente *Evil* en esa celda.

A continuación podemos ver una tabla con los atributos y funciones de la clase.

| Atributos | Descripción |
|---|---|
| MinimumTime | Indica el tiempo mínimo que tardará en realizar un movimiento |
| MaximumTime | Indica el tiempo máximo que tardará en realizar un movimiento |
| GoodAgent | Referencia al agente *Good* dentro del entorno actual |
| EvilAgent | Referencia al agente *Evil* dentro del entorno actual |
| **Funciones** | **Descripción** |
| Action() | Devuelve la acción a realizar mirando las celdas adyacentes |

### 4.3.1.7. Objeto

Los objetos (**Objeto**) representan entidades inanimadas, las cuales, al igual que los agentes, se encuentran en todo momento en una celda, con la salvedad de que estos objetos no pueden moverse de la celda en la que se encuentren, permaneciendo así en la misma celda durante el transcurso de la sesión.

Estos objetos sirven para enriquecer la definición del entorno, de modo que, por ejemplo, los agentes pueden recordar haber pasado por una celda al ver que en ésta se encuentra un objeto que ya vio anteriormente. Así el agente tiene más información de la zona del espacio en la que se encuentra y podrá decidir mejor que acción realizar.

A continuación podemos ver una tabla con los atributos y funciones de la clase.

| Atributos | Descripción |
|---|---|
| Name | Nombre que identifica al objeto |
| Location | Celda en la que se encuentra |
| **Funciones** | **Descripción** |
| SetLocation(NewCell) | Coloca al objeto en la nueva celda |

## 4.3.2. Clases para representar el espacio

### 4.3.2.1. Celda

Una celda (**Cell**) representa una ubicación donde se pueden situar elementos durante el transcurso de las sesiones. Dentro de cada celda se pueden situar tanto objetos como agentes. Estos agentes pueden ir cambiando de celda durante cada interacción, mientras que los objetos permanecen inmóviles en su celda durante el transcurro de la sesión. Además, cada celda contiene la recompensa que deja caer los agentes *Good* y *Evil* si alguno de estos pasa en algún momento por ésta, e irá disminuyendo su valor dividiéndose entre 2 en cada iteración.

A continuación podemos ver una tabla con los atributos y funciones de la clase.

| Atributos | Descripción |
|---|---|
| **Number** | Número que identifica la celda del conjunto de celdas |
| **Objects** | Lista de objetos que se encuentran situados en la celda |
| **Agents** | Lista de agentes que se encuentran situados en la celda |
| **Reward** | Recompensa que queda en la celda |
| **Funciones** | **Descripción** |
| **Clone()** | Crea una copia de la celda actual |
| **GetActions()** | Devuelve el conjunto de acciones que se pueden realizar desde la celda |
| **Move(Action)** | Devuelve la celda a la que se accede si se realizase la acción **Action** |

### 4.3.2.2. Conexión

Al igual que en la teoría de grafos, donde para pasar de un vértice a otro contiguo se debe atravesar una arista, para moverse de una celda a otra contigua se debe pasar a través de una conexión (**Connection**). Cada conexión representa la acción que se debe gastar desde la celda que la contiene para poder llegar a la celda contigua. De modo que al encontrarse en una celda donde exista una conexión representando a una acción, al utilizar esta acción se llegará a la celda con la que conecte esta conexión.

A continuación podemos ver una tabla con los atributos y funciones de la clase.

| Atributos | Descripción |
|---|---|
| Action | Acción que el agente debe realizar para utilizar la conexión |

| Funciones | Descripción |
|---|---|
| GetStartCell() | Devuelve la celda que contiene a la conexión |
| GetEndCell() | Devuelve la celda a la que se llega tras atravesar la conexión |

### 4.3.2.3. Espacio

Toda sesión se realiza dentro de un espacio (**Space**), donde los agentes podrán moverse libremente. Este espacio se define como un grafo dirigido y, por lo tanto, hereda las características de estos grafos, conteniendo así una lista de vértices que representan a las celdas contenidas en el espacio, y una lista de aristas las cuales representan las conexiones entre las celdas.

A continuación podemos ver una tabla con las funciones de la clase.

| Funciones | Descripción |
|---|---|
| GetNumberOfCells() | Devuelve el número de celdas que existen en el espacio |
| LocateObject(Object, Number) | Coloca un objeto en la celda con número **Number** |
| LocateAgent(Agent, Number) | Mueve un agente a la celda con número **Number** |

## 4.3.3. Clases para representar los entornos
### 4.3.3.1. Entorno

En los entornos (**Environment**) se engloban todos los elementos necesarios para poder realizar las sesiones. Esta clase tiene asociados unos agentes y sus acciones y recompensas, determina el agente que se evaluará, qué acciones podrán realizar los agentes, etc. y se encarga de realizar la secuencia de acciones e interacciones

A continuación podemos ver una tabla con los atributos y funciones de la clase.

| Atributos | Descripción |
|---|---|
| **Actions** | Acciones permitidas |
| **MinReward** | Mínima recompensa que se podrá devolver a un agente |
| **MaxReward** | Máxima recompensa que se podrá devolver a un agente |
| **Agents** | Lista de agentes que interactúan |
| **Objects**[1] | Lista de objetos definidos en el espacio |
| **Funciones** | **Descripción** |
| **SetEvaluableAgent(Agent)** | Indica qué agente va a ser evaluado |
| **AddAgent(Agent)** | Añade un agente a la lista de agentes |
| **Interact (Number of Interactions)** | Realiza una sesión durante el número de interacciones definido |
| **InteractWithTime(Time)** | Realiza una sesión durante el tiempo definido |

## 4.3.3.2. Entorno_L (Subclase de Entorno)

Una de las principales clases del sistema es la clase **Environment_L** que describe un tipo de entorno adecuado para realizar tests de inteligencia como se describe en el capitulo 3 (sección 3.5). En esta clase se puede definir, además de lo que permite su clase padre, cómo será el espacio donde interactuarán los agentes. Además, integra automáticamente a los agentes *Good* y *Evil* dentro del entorno.

Para conseguir que la evaluación sea no sesgada para un agente aleatorio (su esperanza sea 0) existe la posibilidad de relocalizar la posición de los agentes *Good* y *Evil* cada vez que transcurra cierto número de interacciones.

---

[1] Esta funcionalidad no se desarrolla en este PFC.

A continuación podemos ver una tabla con los atributos y funciones de la clase.

| Atributos | Descripción |
|---|---|
| **Space** | Espacio donde se realiza la sesión |
| **SpaceDescription** | Descripción textual del espacio |
| **GoodAgent** | Agente *Good* presente en la sesión |
| **EvilAgent** | Agente *Evil* presente en la sesión |
| **Funciones** | **Descripción** |
| **Interact (Number of Interactions)** | Realiza la sesión durante el número de interacciones definido |
| **Interact (Number of Interactions, Number of Interactions to relocate)** | Realiza la sesión durante el número de interacciones definido y relocaliza a los agentes *Good* y *Evil* cada número de interacciones para relocalizar |
| **InteractWithTime(Time)** | Realiza la sesión durante el tiempo definido |
| **InteractWithTime(Time, Number of Interactions to relocate)** | Realiza la sesión durante el tiempo definido y relocaliza a los agentes *Good* y *Evil* cada número de interacciones para relocalizar |
| **GenerateSpace** | Genera aleatoriamente el espacio donde se realiza la sesión |
| **InterpretSpace (Description)** | Interpreta y construye el espacio a partir de su descripción |

## 4.3.3.3. Realización de una sesión de evaluación

## 4.3.3.3.1. Preparación del entorno

Al iniciar una sesión, se debe especificar en el entorno cuál será el agente que se desea evaluar y cómo será el espacio en el que realizará la sesión.

Al construir el entorno se crean automáticamente los agentes *Good* y *Evil* y se añaden a la lista de agentes que el entorno va a manejar.

Antes de comenzar la interacción con los agentes, éstos deben suscribirse al entorno para que puedan interactuar con éste. Sin embargo, solamente uno podrá ser el agente a evaluar, debiendo indicarlo antes de comenzar la sesión.

### 4.3.3.3.2. Descripción del espacio

Para describir cómo será el espacio donde se realizará la sesión, existen dos métodos para definir la distribución de las celdas y las conexiones entre estas:

- Dejar que el entorno genere aleatoriamente el espacio obteniendo así una descripción del entorno, para posteriormente interpretarlo y construirlo.

  o Para ello se ha diseñado un mecanismo en donde, para cada celda, se van generando las acciones que se pueden realizar y con qué celda se conecta a través de dicha acción. De este modo conseguimos un espacio en donde todas las celdas se comunican con las demás celdas a través de las acciones.

- Proporcionar manualmente la descripción de cómo debe ser el espacio y construirlo en función de ésta.

  o Otro mecanismo para definir el espacio es la introducción manual de la representación del espacio. La representación usada para construir el espacio utilizando este mecanismo es la misma que los espacios creados a través del mecanismo anterior.

No todos los espacios generados son válidos para las sesiones que se desean realizar. Por ejemplo, un espacio en el que no se pueda acceder de ninguna forma a una celda no será válido para la evaluación de la sesión. Para que un espacio sea válido, todas las celdas deberán estar conectadas entre sí, sin dejar ninguna celda aislada del resto. Es decir, como se vio en la sección 3.5 el espacio debe ser un grafo completamente conectado.

En esta implementación se le facilita al usuario unas funciones en donde podrá decidir si desea generar automáticamente un espacio conectado, fuertemente conectado o si, por el contrario, desea crearlo siguiendo la descripción que él proporciona quedando bajo su responsabilidad que el espacio esté o no balanceado. En el caso de que decida generar automáticamente el espacio, el sistema se encargará de generar un espacio y comprobará que sea conectado (o fuertemente conectado, dependiendo de cómo quiere el usuario que sea el espacio), en caso de que no lo sea volverá a generar automáticamente otro nuevo espacio hasta generar uno que cumpla con la conectividad requerida por el usuario.

### 4.3.3.3.3. Comienzo de la sesión

Una vez que se inicia la sesión, el sistema prepara el entorno para poder comenzar.

- En primer lugar se sitúa cada agente en una celda elegida aleatoriamente.
- Hay que tener en cuenta que los agentes Good y Evil no pueden encontrarse en ningún momento en la misma celda, por lo que, en el caso de que al colocarlos inicialmente se encuentren en la misma celda, se deben recolocar de nuevo hasta que acaben en celdas distintas.

- Finalmente se les asocia una recompensa a todas las celdas de 0, indicando así que todas las celdas tienen inicialmente una recompensa nula.

### 4.3.3.3.4. Bucle principal

Tras tener preparado el entorno, el sistema comienza las interacciones con los agentes, siguiendo estos pasos:

- Si en un principio se quería que los agentes Good y Evil se resituasen transcurridos un cierto número de interacciones y ha llegado el momento, se procede a su relocalización.
- Inicialmente los agentes Good y Evil dejan caer sus respectivas recompensas en la celda en la que se encuentran, dejando caer el agente Good una recompensa positiva y el agente Evil una recompensa negativa.
- Posteriormente se crea una copia del espacio tal y como se encuentra en ese preciso instante, el cual será el que se les mande a los agentes como la observación del espacio.
- A continuación y para cada agente:
    - Se entrega la recompensa de la última interacción que realizaron. En caso de ser la primera interacción se le da al agente una recompensa inicial de 0.
    - Se manda la observación actual del entorno.
    - Cada agente proporciona la acción que desea realizar en la iteración actual.
- Una vez que todos los agentes han interactuado en la iteración actual, se actualiza el estado del entorno realizando simultáneamente las acciones de todos los agentes.
- Ya que los agentes Good y Evil no deben poder encontrarse en la misma celda, se comprueba si tras realizar el último movimiento ambos se encuentran simultáneamente en la misma celda. Si ambos se encontrasen en la misma se procede de la siguiente manera para resolver el conflicto:
    - Si el agente Good no se movió en esta interacción, se devuelve al agente Evil a la celda en la que se encontraba.
    - Análogamente a la situación anterior, si el agente Evil no se movió se devuelve al agente Good a la celda en la que se encontraba.
    - Si ambos agentes se movieron durante la interacción, se decide aleatoriamente cual de los dos agentes no se mueve en esta interacción y es devuelto a la celda en la que se encontraba.
- Una vez que todos los agentes han sido colocados, se les almacena, para su posterior entrega en la siguiente iteración, la recompensa de la celda a la que se movieron. Dada la situación de que varios agentes acaban en la misma celda, la recompensa que esté en ésta se divide entre el número de agentes que se encuentren en dicha celda.
- Como último paso del bucle se actualizan las recompensas de todas las celdas, dividiendo todas entre 2 y colocando una recompensa de 0 en las celdas donde se encuentre algún agente.

En la sección A1.1 podemos ver el código fuente del bucle principal escrito en JAVA.

### 4.3.3.3.5. Finalización de la sesión

- Finalmente, cuando se ha completado el número de interacciones o el tiempo disponible para realizar la sesión, el sistema la da por terminada y muestra el resultado que ha obtenido el agente que se está evaluando, obteniendo así la medida de rendimiento o inteligencia que ha mostrado el agente durante el transcurso de la sesión.

## 4.3.4. Otras clases auxiliares

### 4.3.4.1. Interacción

Todos los agentes guardan un historial de las interacciones (**Interaction**) realizadas con el entorno durante la sesión en donde pueden consultar todo lo que realizaron durante ésta.

Para poder guardar la observación de cada interacción se ha tenido que duplicar la información contenida en el espacio, ya que, de lo contrario, al guardar directamente la información del espacio y realizar modificaciones, la información guardada sería modificada.

Un inconveniente de tener que duplicar la información contenida en el espacio, es que esta información es la que más espacio ocupa (computacionalmente hablando), por lo que se ha tenido que dar la posibilidad de dejar un máximo de las últimas interacciones a almacenar, de modo que el agente podrá "recordar" hasta un máximo de observaciones pasadas (Atributo **NObservationsSaved** en **Agent**).

A continuación podemos ver una tabla con los atributos de la clase.

| Atributos | Descripción |
| --- | --- |
| **Observation** | Observación proporcionada por el entorno |
| **Action** | Acción realizada por el agente |
| **Reward** | Recompensa devuelta por el entorno |
| **ElapsedTime** | Tiempo tardado en realizar la acción |

### 4.3.4.2. Observación

Para permitir distintos tipos de representaciones de las observaciones se ha creado una estructura de clases en donde se almacenará la observación. Existen dos opciones para representar las observaciones: las "Observaciones Simples" (**Simple Observation**) que representan a las observaciones como una tira de caracteres, y las "Observaciones L" (**Observation L**) en donde se almacena la estructura del espacio con los objetos y agentes presentes en él para representar las observaciones.

## 4.4. Codificación y descripción de espacios

Existen dos formas de construir los espacios: proporcionando manualmente una descripción del espacio o dejando que el entorno lo genere aleatoriamente.

En ambos casos es necesaria una descripción de cómo será el espacio para que posteriormente el entorno lo interprete y lo genere utilizando un grafo dirigido para representarlo.

A continuación veremos cómo se codifican estos espacios y los distintos procedimientos para su construcción.

### 4.4.1. Codificación del espacio

Para codificar los espacios se ha optado por crear una descripción textual de éstos donde se indicará: cuántas celdas contiene el espacio y las conexiones existentes entre estas celdas.

En primer lugar se divide el espacio por partes en función del número de celdas que tenga, de modo que si tiene 5 celdas, la descripción del espacio se dividirá en 5 partes separadas por barras verticales "**|**".

Una vez ya dividido el espacio por celdas, se procede a codificar qué acciones se pueden hacer desde cada una de las celdas y a qué celda se llegará a través de dicha acción. Para ello inicialmente se indican, de forma numérica, todas las acciones que se pueden realizar durante la sesión y, tras cada acción, a qué celda se moverá una vez realizada la acción. Para poder indicar a qué celda se moverá se ha optado por representar el movimiento como un desplazamiento de un número de celdas a partir de la celda actual, indicándolo como una sucesión de "**+**" o "**-**", que representarán un movimiento hacia delante o hacia atrás respectivamente. En caso de que una acción no tenga ningún desplazamiento significará que desde esa celda el uso de esa acción no hará cambiar al agente de celda. Si al intentar realizar un desplazamiento hacia delante se llegase a una celda fuera del número de celdas se seguirá a partir de la primera celda y, análogamente, al intentar realizarlo hacia atrás se seguirá a partir de la última celda. Es decir se sigue un razonamiento circular.

Para entender mejor la codificación de los espacios vamos a ver un ejemplo:

Como descripción del espacio se tiene la siguiente tira de caracteres:

**1+2++3|1+23-|1+23|1+2--3-**

Como podemos ver en la descripción, el espacio se divide en 4 celdas y se pueden realizar 3 acciones. Desde la primera celda se llega hasta la segunda celda tras realizar la acción **1**, la acción **2** nos lleva a la tercera celda y la acción **3** deja al agente en la misma celda. Si vemos la descripción de la cuarta celda podemos ver como la acción **1** nos mueve a la primera celda, con la acción **2** llegamos a la segunda celda y la acción **3** nos devuelve a la tercera celda.

En la siguiente imagen podemos ver una representación gráfica del espacio descrito.

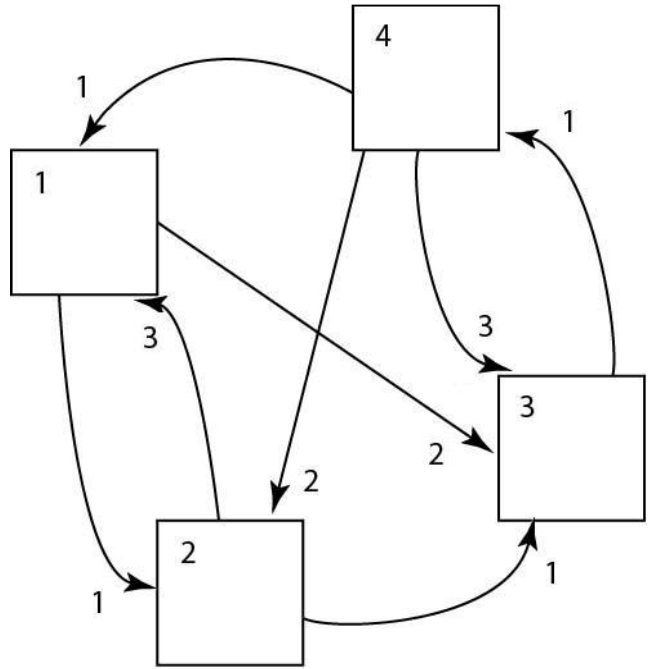

## 4.4.2. Descripción manual del espacio

Una forma de construir el espacio es proporcionándole directamente al entorno, utilizando la codificación antes descrita, un tira de caracteres describiendo cómo debe ser el espacio que se va a utilizar. Una vez obtenida la descripción del espacio, el entorno lo generará automáticamente construyendo el grafo dirigido que lo representa, creando las celdas y conexiones necesarias para representar adecuadamente el espacio descrito.

## 4.4.3. Descripción aleatoria del espacio

La segunda forma de construir el espacio es dejando que el entorno se encargue automáticamente de generar de forma aleatoria la descripción del espacio que se utilizará para la sesión para que, posteriormente, sea interpretado y generado de igual modo que en la forma manual.

Para generar aleatoriamente el espacio se ha decidido utilizar el siguiente método:

- Inicialmente se decide cuál es el número de celdas de las que se compone el espacio a generar. Para ello se ha utilizado una distribución universal como la descrita en la sección 3.3, pero para enteros. En esta distribución se considera que el número con mayor probabilidad de salir será el número más pequeño desde el que iniciamos la distribución, con un 50% de probabilidad, en caso de no elegirse este número se pasará a considerar el siguiente número con la misma probabilidad (50%) y así sucesivamente hasta que se decida el número de casillas de las que se dispondrá. El número mínimo de celdas que tendrá el espacio es 2.
- Al igual que para elegir el número de celdas, se utiliza una distribución universal para decidir el número de acciones disponibles, utilizando como cota máxima el número de celdas que tendrá el espacio, ya que, si hubieran más acciones que celdas, dos acciones de la misma celda siempre acabarán haciendo lo mismo.
- Una vez que disponemos del número de celdas y de acciones que tendrá el espacio, generamos aleatoriamente (con una distribución uniforme), para cada celda y cada acción, con qué celda se conectará (con cada una de las acciones) cada una de las celdas, pudiéndose incluso decidir que una acción no conectará la celda con ninguna otra y asumiéndose en este caso que la acción lo mantiene en la celda. Para ello se elige: el símbolo indicando si la conexión se hará con una celda anterior o posterior y un número aleatorio entre 0 y el número de celdas del espacio para saber cuántas celdas se desplazará el agente.

En la sección A1.2 podemos ver el código fuente de la generación aleatoria de espacios escrito en JAVA.

## 4.5. Interfaz

A continuación vamos a explicar en detalle cómo funciona la interfaz de usuario del programa.

Nada más empezar podemos elegir en el menú inicial cuál será el agente evaluable que utilizaremos para realizar la sesión (humano o aleatorio) y cómo queremos que se construya el espacio (de forma manual o aleatoria).

Una vez decidida la configuración que utilizaremos, podemos empezar la sesión.

Al comenzar la sesión, el entorno se prepara para empezar, preparando el espacio y colocando aleatoriamente a los agentes que interactuarán durante el transcurso de la sesión.

En la siguiente imagen podemos ver cómo queda el entorno tras indicarle que el agente evaluable será un agente **humano**, y que deseamos utilizar un espacio **definido** por nosotros. Para explicar la interfaz hemos habilitado la visibilidad de las recompensas en cada celda, sin embargo para realizar una sesión real esta información permanecerá oculta, por lo que los agentes no podrán utilizarla.

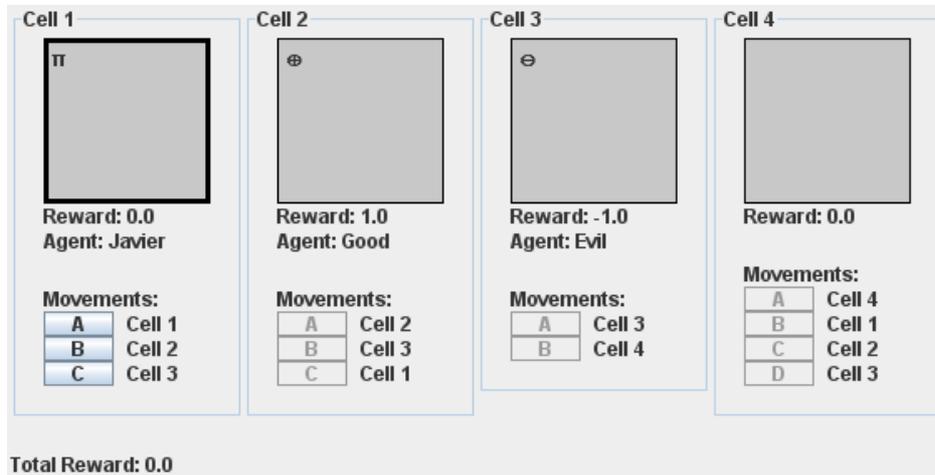

Como podemos ver, la celda con el borde más grueso indica dónde se encuentra situado el agente que el sistema está evaluando. A su vez, dentro de cada celda podemos ver una representación gráfica de los agentes y objetos que están situados dentro de éstas. Entre los agentes vemos representados: al agente que se está evaluando como π, al agente *Good* como ⊕ y al agente *Evil* como ⊖.

A continuación y en cada celda podemos ver de forma textual: la recompensa que tiene la celda, los agentes que están situados en ésta y la lista de acciones que se puede realizar y con qué celda conecta la acción.

Finalmente podemos ver en tiempo real la recompensa global que está teniendo el agente durante el transcurso de la sesión.

Puesto que el agente a evaluar es **humano**, el entorno nos permite seleccionar qué movimientos queremos realizar. Para ello podemos elegir utilizando los botones situados en la sección *Movements* la celda a la que queremos movernos.

En la imagen siguiente podemos ver cómo reacciona el entorno tras realizar el movimiento B (correspondiente a la acción 1).

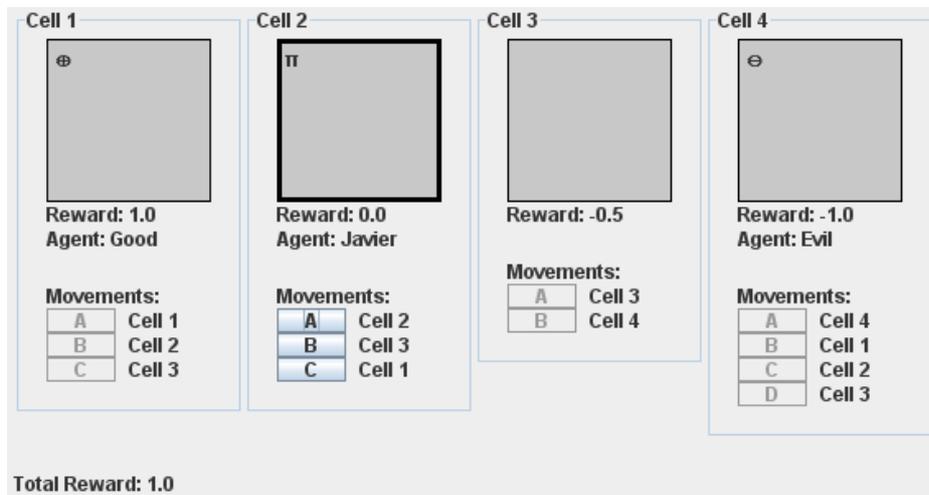

Como podemos ver, el agente evaluable ha recogido la recompensa **1** que estaba situada en la celda **2**, mientras que los agentes *Good* y *Evil* se han movido a las celdas **1** y **4** respectivamente, dejando caer en sus respectivas celdas sus recompensas. Si nos fijamos en la celda **3** podemos ver como el agente *Evil* ha dejado tras de sí su rastro de recompensas, que como se ha explicado anteriormente va dividiéndose entre 2 en cada iteración.

Si hubiéramos elegido al agente **Aleatorio** como agente a evaluar la interfaz de usuario hubiera cambiado ligeramente. Puesto que un agente **Aleatorio** no puede ser controlado por el usuario, el interfaz no nos permite la interacción con el entorno deshabilitándonos las acciones que puede realizar el agente. En la siguiente imagen podemos ver el mismo entorno habiendo seleccionado como agente a evaluar al agente **Aleatorio**.

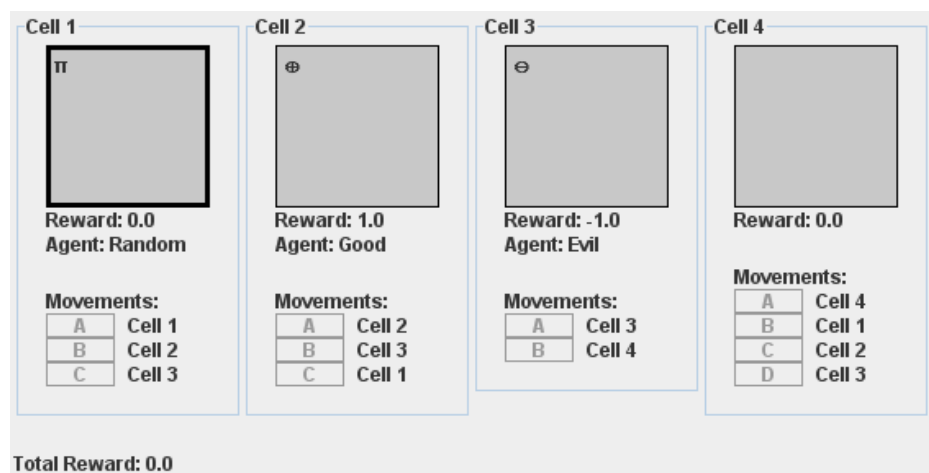

# 5. Experimentos

En esta sección veremos los resultados de algunos experimentos realizados utilizando la arquitectura diseñada. En ellas veremos cómo se cumplen las características de los entornos balanceados, dando para agentes aleatorios resultados cercanos a 0. También veremos los resultados de los mismos tipos de entornos para un agente llamado Observador, el cual dispone de información adicional que le ayuda a conseguir mejores resultados.

A continuación veremos los resultados de los experimentos realizados utilizando espacios previamente diseñados y generados aleatoriamente, los cuales contienen diversidad de celdas y acciones. Los resultados de los experimentos se muestran como la media de varias sesiones siguiendo un amplio rango de iteraciones. También veremos los resultados de experimentos en donde no se garantizan entornos balanceados y en donde se evalúan a varios agentes en una misma sesión.

Todos los experimentos se han realizado tanto relocalizando a los agentes generadores de recompensas (*Good* y *Evil*) como sin relocalizarlos. Para ver la distinción podemos ver, en las tablas donde se resumen los experimentos, la marca "(Sin cambio)" en aquellos experimentos en donde no se relocalizan a los agentes generadores y ninguna marca para aquellos en donde sí se les relocaliza.

## 5.1. Entornos definidos manualmente

En este apartado veremos los resultados de experimentos realizados en entornos completamente balanceados, con los agentes *Good* y *Evil* siguiendo el mismo comportamiento y con espacios desde 2 hasta 10 celdas para los cuales los espacios estarán siempre completamente conectados, es decir, se podrá llegar siempre desde cualquier celda a cualquier otra celda realizando las acciones oportunas. Para cada espacio veremos su codificación y una visión gráfica donde podremos ver como se conecta el espacio.

En todos los espacios mostrados a continuación, todas las celdas tienen siempre una acción por defecto que conecta a una celda consigo misma a través de la acción 0. Para no repetir continuamente esta información en las imágenes y en la codificación de los espacios esta información se obvia, siendo el sistema el encargado de construir esta acción para todas las celdas.

# Experimentos con un espacio de 2 celdas

Espacio definido manualmente

Codificación del espacio: **1+|1-**

Número de celdas: 2

Número de acciones: 1

Good y Evil: Comportamiento aleatorio

Visión gráfica

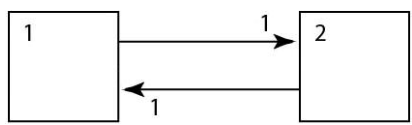

Media de resultados tras 10 sesiones realizadas.

| Agente<br>Iteraciones | Aleatorio | Observador | Aleatorio<br>(Sin cambio) | Observador<br>(Sin cambio) |
|---|---|---|---|---|
| **5** | 0.04 | 0.5 | 0 | 0.5 |
| **10** | -0.02 | 0.5 | 0.03 | 0.5 |
| **20** | -0.015 | 0.5 | -0.055 | 0.5 |
| **50** | 0.0260 | 0.5 | 0.01 | 0.5 |
| **100** | -0.028 | 0.5 | -0.002 | 0.5 |
| **200** | -0.008 | 0.5 | -0.012 | 0.5 |
| **500** | 0.0114 | 0.5 | -0.009 | 0.5 |
| **1000** | 0.0137 | 0.5 | -0.0017 | 0.5 |
| **2000** | -0.0002 | 0.5 | -0.00525 | 0.5 |
| **1000000** | 0.0000453 | 0.5 | 0.0000849 | 0.5 |

Como podemos observar, los resultados obtenidos por los agentes aleatorios van acercándose a resultados cercanos a 0, pudiendo ser estos resultados tanto positivos como negativos, a medida que las sesiones tienen más iteraciones los resultados obtenidos para un agente aleatorio se acercan cada vez más a 0. Esto se debe a que en sesiones con pocas iteraciones no es posible medir fiablemente la inteligencia de un agente, ya que en cualquier momento podría disponer de golpes de suerte (o de mala suerte). Al aumentar el número de iteraciones necesarias para la sesión, el agente aleatorio obtiene cifras mucho más cercanas a 0, lo cual reafirma que la inteligencia demostrada por un agente aleatorio es la esperada para un entorno balanceado.

En lo referente al agente observador podemos ver como en este espacio siempre obtiene una recompensa de 0.5. Esto es debido a que ambas celdas siempre están ocupadas por algún agente distinto al que se está evaluando, los cuales son o *Good* o *Evil*, y por lo tanto siempre deberá compartir la recompensa obtenida en cualquier celda. Debido a que las sesiones se han realizado en un espacio tan pequeño, el agente observador siempre puede ver al agente *Good*, y por lo tanto siempre irá a por su recompensa. Sin embargo, como ya se ha dicho antes, está recompensa se dividirá entre los dos agentes que ocupan la celda obteniendo siempre recompensas de 0.5. Por lo tanto, en este espacio tan pequeño, el agente observador siempre obtiene la mayor recompensa posible dando como resultado una recompensa media de 0.5.

# Experimentos con un espacio de 4 celdas

Espacio definido manualmente

Codificación del espacio: **1+2++3|1+23-|1+23|1+2--3-**

Número de celdas: 4

Número de acciones: 3

Good y Evil: Comportamiento aleatorio

Visión gráfica

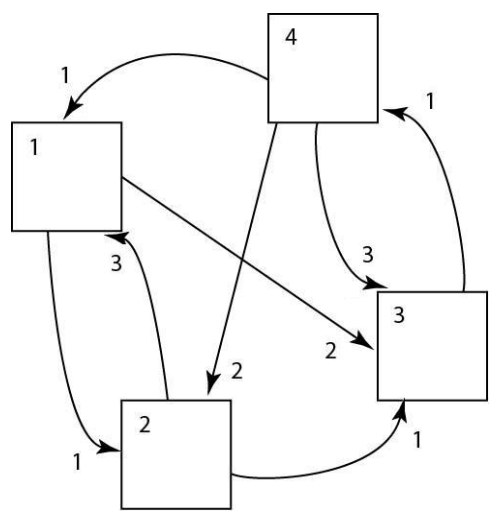

Media de resultados tras 10 sesiones realizadas.

| Agente<br>Iteraciones | Aleatorio | Observador | Aleatorio<br>(Sin cambio) | Observador<br>(Sin cambio) |
|---|---|---|---|---|
| **5** | 0.01375 | 0.61 | 0.1 | 0.5 |
| **10** | 0.03484375 | 0.695 | -0.011875 | 0.685 |
| **20** | -0.06 | 0.66625 | -0.05875 | 0.6525 |
| **50** | -0.045953125 | 0.717 | 0.020953125 | 0.686 |
| **100** | -0.0232026367 | 0.689125 | 0.0274492188 | 0.6895 |
| **200** | -0.0216215820 | 0.6755625 | 0.0034672851 | 0.67675 |
| **500** | -0.0226217712 | 0.6805734375 | -0.0022934555 | 0.69005 |
| **1000** | -0.012627417 | 0.6735982422 | -0.0053155426 | 0.6942 |
| **2000** | -0.0032242306 | 0.6795856934 | -0.0125768757 | 0.68945 |
| **1000000** | -0.0001506778 | 0.6871037385 | 0.0014431725 | 0.7003017 |

Los resultados obtenidos por el agente aleatorio siguen el mismo patrón que en los experimentos anteriores.

Una vez que disponemos de un espacio con un mayor número de celdas, las recompensas que se encuentran en éstas no siempre deberán compartirse entre varios agentes, por lo tanto la esperanza media de recompensas será superior a 0.5. Si realizamos los mismos experimentos sin recolocar a los agentes *Good* y *Evil* las recompensas obtenidas por el agente observador siempre resultan ligeramente superiores a cuando se realizaba el cambio de posiciones de *Good* y *Evil*. Este es debido a que el agente observador no debe volver a buscar al agente *Good* tras cada relocalización y, por lo tanto, no perderá la recompensa que va dejando al tratar de encontrarle.

# Experimentos con un espacio de 6 celdas

Espacio definido manualmente

Codificación del espacio: **1--2+|1++2--|1++2-|12---|1-2+|12---**

Número de celdas: 6

Número de acciones: 2

Good y Evil: Comportamiento aleatorio

Visión gráfica

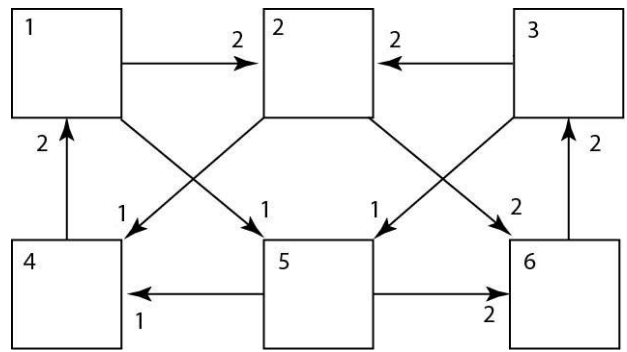

Media de resultados tras 10 sesiones realizadas.

| Agente<br>Iteraciones | Aleatorio | Observador | Aleatorio<br>(Sin cambio) | Observador<br>(Sin cambio) |
|---|---|---|---|---|
| **5** | -0.03 | 0.5 | 0.155625 | 0.405 |
| **10** | 0.1425390625 | 0.54125 | 0.01984375 | 0.60125 |
| **20** | 0.0127319336 | 0.765 | -0.026171875 | 0.698125 |
| **50** | -0.0056818848 | 0.7214375 | -0.0224951172 | 0.7095 |
| **100** | 0.0046063232 | 0.6898037109 | 0.0245116272 | 0.71296875 |
| **200** | -0.0140901871 | 0.6739960926 | 0.0214790373 | 0.7415 |
| **500** | -0.0093691259 | 0.6934712868 | -0.0014536697 | 0.734 |
| **1000** | 0.0033478214 | 0.6812413848 | -0.0005849602 | 0.737475 |
| **2000** | -0.0000222531 | 0.6858013640 | -0.0103504164 | 0.7333 |
| **1000000** | -0.0021152493 | 0.6837011002 | -0.0020848308 | 0.7384596 |

Como podemos ver, el agente aleatorio sigue obteniendo resultados cercanos a 0. Aunque en algunos experimentos ha tenido más suerte que en otros y por lo tanto su recompensa media difiere bastante del resto de experimentos. Podemos ver esto en los experimentos para el agente aleatorio con 10 iteraciones y con 5 iteraciones sin cambio para *Good* y *Evil*. Estos resultados excepcionales normalmente solo se dan en experimentos realizados en sesiones con pocas iteraciones, en donde la media de las recompensas obtenidas aun no es del todo fiable.

Al igual que en los experimentos realizados en el espacio de 4 celdas, el agente observador sigue manteniendo resultados superiores a 0.5. Sin embargo podemos ver que existe una mayor diferencia entre los resultados obtenidos en los entornos donde no se cambia a *Good* ni a *Evil* y aquellos en los que si se les cambia. Esta diferencia va aumentando debido a que al agente observador le cuesta más encontrar nuevamente a *Good* en espacios cada vez más grandes, por lo que pierde cada vez más recompensas positivas tratando de encontrarle. Por otro lado podemos ver como las recompensas del observador son algo superiores que en el espacio anterior, ya que en espacios cada vez mayores los agentes *Good* y *Evil* tendrán una menor probabilidad de encontrarse y el agente *Good* no deberá renunciar a su movimiento, por lo que habrán menos ocasiones en donde su recompensa la comparta con el observador. Por lo tanto, el observador compartirá menos recompensas conforme el espacio contenga más celdas.

# Experimentos con un espacio de 8 celdas

Espacio definido manualmente

Codificación del espacio: **1+2+++3|1+2-3---|1++2-3|1---2++3+|1+++2--3-|1--2+3|1-2+3+++|1-2---3**

Número de celdas: 8

Número de acciones: 3

Good y Evil: Comportamiento aleatorio

Visión gráfica

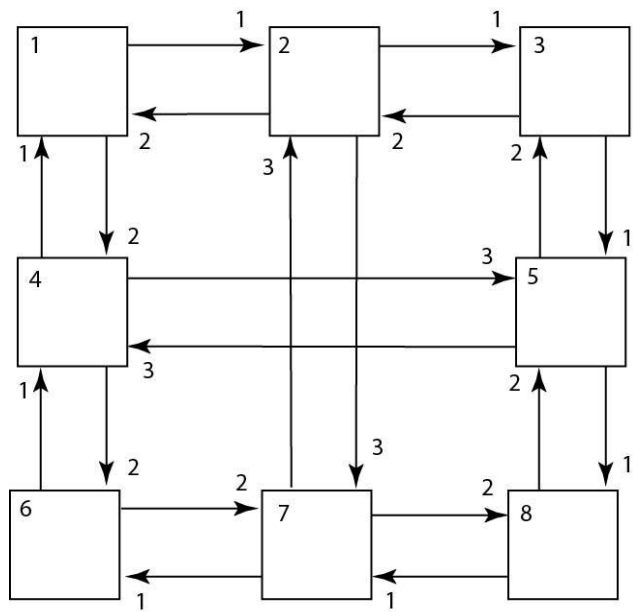

Media de resultados tras 10 sesiones realizadas.

| Agente<br>Iteraciones | Aleatorio | Observador | Aleatorio<br>(Sin cambio) | Observador<br>(Sin cambio) |
|---|---|---|---|---|
| **5** | -0.0675 | 0.6525 | 0.0325 | 0.67 |
| **10** | 0.026328125 | 0.795 | 0.0076171875 | 0.725 |
| **20** | 0.0646356201 | 0.73625 | -0.0195324707 | 0.67125 |
| **50** | 0.0005088654 | 0.731375 | -0.0046048894 | 0.77675 |
| **100** | 0.0064107894 | 0.7168063924 | 0.0373921991 | 0.768 |
| **200** | -0.0098070030 | 0.7353776855 | 0.0008353519 | 0.7845 |
| **500** | -0.0026782425 | 0.7257369799 | 0.0045596895 | 0.7877 |
| **1000** | -0.0076095967 | 0.7322875084 | -0.0029735875 | 0.79015 |
| **2000** | 0.0052535156 | 0.7306702732 | 0.0043954454 | 0.790825 |
| **1000000** | 0.0012538494 | 0.7362988065 | 0.0005897395 | 0.79489045 |

Siguiendo la dinámica de los experimentos anteriores, el agente aleatorio sigue manteniendo resultados en torno a 0.

En este caso, los resultados del agente observador son aún superiores que en los experimentos del espacio con 6 celdas. Como se ha visto antes esto se debe a que existen menos probabilidades de que los agentes *Good* y *Evil* se encuentren y por lo tanto la recompensa de *Good* se comparte con menor frecuencia. Sin embargo existe un motivo añadido por lo que los resultados del observador han aumentado tanto con respecto a la diferencia entre los experimentos del espacio con 4 celdas y el de 6, en donde los resultados difieren en 0.4 mientras que los resultados entre estos experimentos y los experimentos del espacio anterior difieren en un intervalo entre 0.5 y 0.6. Esto es porque en estos experimentos *Good* y *Evil* disponen de más acciones para realizar desde prácticamente todas las celdas y, por lo tanto, decidirán con menor frecuencia permanecer en su celda, aumentando la media de recompensas que recogerá el observador.

# Experimentos con un espacio de 10 celdas

Espacio definido manualmente

Codificación del espacio: **123+++++|1-23+++++|1-2+3+++++|12+3+++++|123++++|1+2++3|1+23|1+++2-----3----|12-3|1--2-3**

Número de celdas: 10

Número de acciones: 3

Good y Evil: Comportamiento aleatorio

Visión gráfica

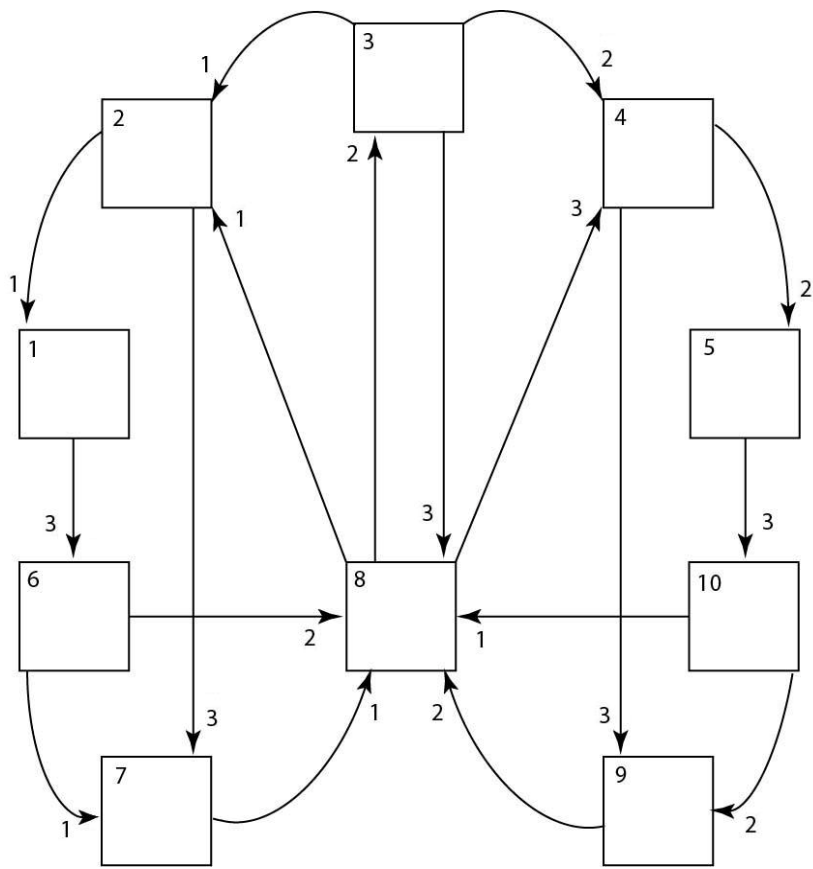

Media de resultados tras 10 sesiones realizadas.

| Agente<br>Iteraciones | Aleatorio | Observador | Aleatorio<br>(Sin cambio) | Observador<br>(Sin cambio) |
|---|---|---|---|---|
| **5** | 0.049375 | 0.3575 | 0.0175 | 0.31 |
| **10** | 0.056484375 | 0.47609375 | 0.0416992188 | 0.535 |
| **20** | -0.0252062988 | 0.6625 | 0.0191992188 | 0.7175 |
| **50** | -0.0045199013 | 0.72475 | -0.0269153652 | 0.673484375 |
| **100** | 0.0023949995 | 0.6618558960 | -0.0203063105 | 0.73071875 |
| **200** | -0.011540667 | 0.6591534714 | -0.0131583569 | 0.75575 |
| **500** | -0.0026265456 | 0.6586628568 | -0.0004145666 | 0.76438125 |
| **1000** | 0.0022877179 | 0.6650044655 | -0.0145762173 | 0.7564539063 |
| **2000** | -0.0041933919 | 0.6580340589 | -0.0041297061 | 0.7626625 |
| **1000000** | -0.0028186475 | 0.6693211141 | -0.0017103821 | 0.7645574133 |

Como cabría esperar al agente aleatorio sigue manteniendo los mismos resultados.

Al agente observador le cuesta más encontrar al *Good* en este espacio con esta topología, por lo que la diferencia de los resultados entre relocalizar o no a *Good* y a *Evil* aumenta en estos experimentos. Cabría esperar que los resultados medios del observador sin relocalizar a *Good* y a *Evil* fuera superior que en los experimentos con el espacio de 8 celdas, ya que contiene más celdas y el mismo número de acciones. Si nos fijamos más en la topología del espacio generalmente solo se pueden realizar una o dos acciones desde cada celda, sin contar quedarse en la misma celda, cuando en los experimentos anteriores desde las celdas se podían realizar dos o tres acciones. Ya que el agente *Good* tendrá una mayor probabilidad de mantenerse en la celda en la que se encuentre sus recompensas serán más veces compartidas y, por lo tanto, la media de recompensas no aumenta como era de esperar.

## 5.2. Entornos generados automáticamente

A continuación mostramos los resultados a los experimentos realizados generando aleatoriamente los espacios tal y como se describe en el apartado 4.4.3.

### 5.2.1. Espacios generados automáticamente (Conectados)

Veamos los resultados obtenidos utilizando una media de 100 sesiones con los distintos agentes e iteraciones utilizando distintos espacios en cada sesión teniendo como única restricción que ninguna celda quede desconectada del resto.

## 4-celdas 3-acciones

Espacio generado automáticamente (Conectado)

Número de celdas: 4

Número de acciones: 3

Good y Evil: Comportamiento aleatorio

Media de resultados tras 100 sesiones realizadas.

| Agente<br>Iteraciones | Aleatorio | Observador | Aleatorio (Sin cambio) | Observador (Sin cambio) |
|---|---|---|---|---|
| **5** | 0.04675 | 0.522 | -0.008 | 0.4825 |
| **10** | 0.02975 | 0.543625 | 0.0120546875 | 0.5098125 |
| **20** | 0.0395100098 | 0.565125 | -0.0101875 | 0.56775 |
| **50** | 0.0152491213 | 0.506125 | 0.0058412355 | 0.5735 |
| **100** | 0.0092045895 | 0.5229386719 | 0.0242653084 | 0.5480625 |
| **200** | -0.003388797 | 0.4989532227 | -0.0251157016 | 0.5288375 |
| **500** | -0.0006818127 | 0.478607851 | -0.0309036888 | 0.537445 |
| **1000** | 0.0044138150 | 0.4782791846 | 0.0101541339 | 0.56831375 |
| **2000** | 0.0012512472 | 0.4670196489 | 0.0160087209 | 0.556584375 |
| **100000** | 0.0018664954 | 0.4964514401 | -0.0035218350 | 0.56775625 |

# 4-celdas 4-acciones

Espacio generado automáticamente (Conectado)

Número de celdas: 4

Número de acciones: 4

Good y Evil: Comportamiento aleatorio

Media de resultados tras 100 sesiones realizadas.

| Agente<br>Iteraciones | Aleatorio | Observador | Aleatorio<br>(Sin cambio) | Observador<br>(Sin cambio) |
|---|---|---|---|---|
| **5** | 0.036375 | 0.59 | 0.022125 | 0.577 |
| **10** | -0.031546875 | 0.5985 | -0.013078125 | 0.6525 |
| **20** | 0.0068369140 | 0.6365 | -0.0067470703 | 0.6276875 |
| **50** | 0.0070780273 | 0.6219117188 | -0.0065885833 | 0.6309 |
| **100** | -1.7125701829 | 0.6148359130 | 0.0050069703 | 0.651575 |
| **200** | -0.0031231798 | 0.6204983398 | -0.0150923710 | 0.6380875 |
| **500** | 0.0021759797 | 0.6267056823 | -0.0115480778 | 0.655715 |
| **1000** | 0.0037967063 | 0.5886907487 | -0.0030248689 | 0.636795 |
| **2000** | -0.0025212755 | 0.6180830476 | 0.0154158361 | 0.645344375 |
| **100000** | 0.0019821357 | 0.6394119529 | 0.0072174910 | 0.643137375 |

# 6-celdas 4-acciones

Espacio generado automáticamente (Conectado)

Número de celdas: 6

Número de acciones: 4

Good y Evil: Comportamiento aleatorio

Media de resultados tras 100 sesiones realizadas.

| Agente  Iteraciones | Aleatorio | Observador | Aleatorio (Sin cambio) | Observador (Sin cambio) |
|---|---|---|---|---|
| **5** | 0.071625 | 0.553375 | -0.061 | 0.57875 |
| **10** | 0.0483203125 | 0.6700625 | 0.0099902343 | 0.656375 |
| **20** | 0.0061313476 | 0.6654373798 | 0.0136739501 | 0.68552929687 |
| **50** | 0.0057017860 | 0.6950187499 | 0.0234200533 | 0.7143734375 |
| **100** | 0.0201091887 | 0.6770288970 | -0.0002505793 | 0.722665625 |
| **200** | 0.0015357381 | 0.6742627235 | -0.0030017522 | 0.72453125 |
| **500** | 0.0070556115 | 0.6666639230 | 0.0029403885 | 0.71003 |
| **1000** | 0.0056254281 | 0.6698362898 | 0.0136878712 | 0.7185325 |
| **2000** | 0.0079251038 | 0.6946516539 | -0.0032442967 | 0.72596 |
| **100000** | 0.0040647799 | 0.6860570746 | -0.0038440883 | 0.731519625 |

# 6-celdas 6-acciones

Espacio generado automáticamente (Conectado)

Número de celdas: 6

Número de acciones: 6

Good y Evil: Comportamiento aleatorio

Media de resultados tras 100 sesiones realizadas.

| Agente<br>Iteraciones | Aleatorio | Observador | Aleatorio<br>(Sin cambio) | Observador<br>(Sin cambio) |
|---|---|---|---|---|
| **5** | 0.0075625 | 0.707125 | -0.0470625 | 0.689 |
| **10** | -0.0184033203 | 0.7268125 | 0.0123828125 | 0.746375 |
| **20** | -0.0197189598 | 0.74665625 | -0.0074642486 | 0.75175 |
| **50** | -0.0037345133 | 0.757561615 | 0.0013374073 | 0.769125 |
| **100** | 0.0076907712 | 0.7611456665 | 0.0012145443 | 0.77585 |
| **200** | 0.0033358568 | 0.7593348992 | 0.0064964131 | 0.7769125 |
| **500** | 0.0026132700 | 0.756693748 | 0.0007169638 | 0.771575 |
| **1000** | -0.0000698110 | 0.7598566768 | 0.0006846164 | 0.775176875 |
| **2000** | 0.0010336988 | 0.759557398 | -0.0044459509 | 0.7708675 |
| **100000** | 0.0019426404 | 0.7601891151 | 0.0018648231 | 0.778565075 |

# 8-celdas 4-acciones

Espacio generado automáticamente (Conectado)

Número de celdas: 8

Número de acciones: 4

Good y Evil: Comportamiento aleatorio

Media de resultados tras 100 sesiones realizadas.

| Agente<br>Iteraciones | Aleatorio | Observador | Aleatorio<br>(Sin cambio) | Observador<br>(Sin cambio) |
|---|---|---|---|---|
| **5** | 0.024375 | 0.5385625 | 0.0236875 | 0.4915 |
| **10** | -0.007171875 | 0.65865625 | -0.0072167968 | 0.629703125 |
| **20** | 0.0188020324 | 0.71584375 | 0.0037644996 | 0.67070898437 |
| **50** | -0.0033710265 | 0.6881226531 | -0.0117281041 | 0.7314 |
| **100** | -0.0022428926 | 0.7237871719 | 0.0012601332 | 0.75131875 |
| **200** | 0.0173139745 | 0.7016593765 | 0.0022134205 | 0.75735419921 |
| **500** | 0.0028826708 | 0.7079315228 | -0.0077612121 | 0.751844375 |
| **1000** | 0.0023655002 | 0.7005680074 | 0.0001749424 | 0.767868125 |
| **2000** | 0.0018537326 | 0.7140603729 | 0.0026119929 | 0.7673409375 |
| **100000** | 0.0043064303 | 0.7171128281 | 0.0082247295 | 0.7661230625 |

# 8-celdas 6-acciones

Espacio generado automáticamente (Conectado)

Número de celdas: 8

Número de acciones: 6

Good y Evil: Comportamiento aleatorio

Media de resultados tras 100 sesiones realizadas.

| Agente  Iteraciones | Aleatorio | Observador | Aleatorio (Sin cambio) | Observador (Sin cambio) |
|---|---|---|---|---|
| **5** | -0.042125 | 0.6885 | 0.006 | 0.681 |
| **10** | 0.010375 | 0.7228125 | -0.026921875 | 0.7365625 |
| **20** | -0.0080527896 | 0.7798125 | -0.0188741035 | 0.7798125 |
| **50** | -0.0015797378 | 0.7846015625 | -0.0023470562 | 0.8034875 |
| **100** | -0.0002917564 | 0.7820518463 | -0.0029158655 | 0.80525 |
| **200** | 0.0007346433 | 0.7814969837 | -0.0053570862 | 0.810796875 |
| **500** | -0.0023997683 | 0.7827005168 | -0.0012930265 | 0.809630625 |
| **1000** | -0.0000345703 | 0.7824668042 | 0.0009552746 | 0.810276875 |
| **2000** | 0.0012363421 | 0.7830145245 | -0.0005975208 | 0.8101525 |
| **100000** | 0.0002894769 | 0.7804884726 | 0.0009075608 | 0.8102544 |

# 8-celdas 8-acciones

Espacio generado automáticamente (Conectado)

Número de celdas: 8

Número de acciones: 8

Good y Evil: Comportamiento aleatorio

Media de resultados tras 100 sesiones realizadas.

| Agente Iteraciones | Aleatorio | Observador | Aleatorio (Sin cambio) | Observador (Sin cambio) |
|---|---|---|---|---|
| 5 | -0.064875 | 0.757 | -0.01025 | 0.74925 |
| 10 | -0.0170507812 | 0.795125 | 0.0113867187 | 0.79775 |
| 20 | 0.0048254966 | 0.8055625 | 0.0104834518 | 0.815375 |
| 50 | -0.0065295809 | 0.8153328063 | 0.0034930459 | 0.820025 |
| 100 | -0.0016721380 | 0.8144338834 | -0.0001513715 | 0.8249 |
| 200 | -0.0039921943 | 0.8135470728 | -0.0044161980 | 0.831725 |
| 500 | 0.0005881382 | 0.8154931645 | -0.0014925117 | 0.83276 |
| 1000 | -0.0056278378 | 0.8151737539 | 0.0004738523 | 0.8343375 |
| 2000 | -0.0030229432 | 0.8151252478 | -0.0035334687 | 0.83393125 |
| 100000 | -0.0004472968 | 0.8144933822 | -0.0022796966 | 0.833287725 |

# 10-celdas 4-acciones

Espacio generado automáticamente (Conectado)

Número de celdas: 10

Número de acciones: 4

Good y Evil: Comportamiento aleatorio

Media de resultados tras 100 sesiones realizadas.

| Agente<br>Iteraciones | Aleatorio | Observador | Aleatorio<br>(Sin cambio) | Observador<br>(Sin cambio) |
|---|---|---|---|---|
| **5** | -0.003375 | 0.503 | 0.0348125 | 0.496 |
| **10** | -0.0067871093 | 0.6367578125 | 0.0029042968 | 0.618125 |
| **20** | 0.0029400844 | 0.662703125 | 0.0060056285 | 0.6786015625 |
| **50** | -0.0126927395 | 0.720691467 | -0.0003916556 | 0.74152036132 |
| **100** | 0.0010064249 | 0.7064358273 | -0.0025521861 | 0.777025 |
| **200** | 0.0021113307 | 0.7126902210 | -0.0017698301 | 0.76821494140 |
| **500** | 0.0081824189 | 0.7088160847 | -0.0064782360 | 0.781974375 |
| **1000** | 0.0023236015 | 0.7163777811 | 0.0137967548 | 0.78701601562 |
| **2000** | 0.0054380301 | 0.7104929906 | 0.0007256349 | 0.78096863281 |
| **100000** | 0.0047555243 | 0.7138575059 | 0.0045655560 | 0.78658997968 |

# 10-celdas 7-acciones

Espacio generado automáticamente (Conectado)

Número de celdas: 10

Número de acciones: 7

Good y Evil: Comportamiento aleatorio

Media de resultados tras 100 sesiones realizadas.

| Agente<br>Iteraciones | Aleatorio | Observador | Aleatorio<br>(Sin cambio) | Observador<br>(Sin cambio) |
|---|---|---|---|---|
| **5** | 0.0025 | 0.66275 | 0.007375 | 0.62725 |
| **10** | 0.0162539062 | 0.7975 | 0.0067558593 | 0.77346875 |
| **20** | -0.0139494881 | 0.80575 | -0.0144310302 | 0.818671875 |
| **50** | 0.0058377407 | 0.811679782 | -0.0038752322 | 0.83715 |
| **100** | -0.0123828109 | 0.8106688241 | 0.0032449395 | 0.833878125 |
| **200** | -0.0005640175 | 0.8066227070 | 0.0021693778 | 0.8405 |
| **500** | -0.0002181396 | 0.8103837011 | 0.0013080553 | 0.8436875 |
| **1000** | -0.0018019882 | 0.8113200001 | 0.0003544006 | 0.845283125 |
| **2000** | -0.0005624005 | 0.8105365493 | 0.0004085095 | 0.8450575 |
| **100000** | 0.0000173130 | 0.8120781481 | 0.0000560432 | 0.84557391875 |

# 10-celdas 10-acciones

Espacio generado automáticamente (Conectado)

Número de celdas: 10

Número de acciones: 10

Good y Evil: Comportamiento aleatorio

Media de resultados tras 100 sesiones realizadas.

| Agente<br>Iteraciones | Aleatorio | Observador | Aleatorio<br>(Sin cambio) | Observador<br>(Sin cambio) |
|---|---|---|---|---|
| **5** | 0.0100625 | 0.7665 | -0.019375 | 0.7645 |
| **10** | 0.00984375 | 0.8095625 | 0.0020917968 | 0.8275 |
| **20** | -0.003994873 | 0.84775 | 0.0191303443 | 0.84225 |
| **50** | -0.0153001013 | 0.8492577506 | -0.0099471920 | 0.86035 |
| **100** | 0.0017704384 | 0.8515060465 | 0.0041125094 | 0.8641625 |
| **200** | 0.0055527598 | 0.8467617110 | 0.0028669731 | 0.86915 |
| **500** | -0.0025842754 | 0.8505609456 | 0.0037606309 | 0.868675 |
| **1000** | -0.0016398764 | 0.8476845503 | 0.0003439481 | 0.8681925 |
| **2000** | 0.0003998080 | 0.8493090214 | 0.0002639389 | 0.86781 |
| **100000** | 0.0004557696 | 0.8486287340 | 0.0004261841 | 0.8679295 |

Como hemos podido observar el agente aleatorio siempre obtiene resultados cercanos a 0, cumpliéndose las propiedades de un entorno balanceado.

A medida que va creciendo el número de celdas y de acciones, al igual que con los espacios definidos, el agente *Good* tiene menor probabilidad de encontrarse con el agente *Evil* y más posibilidades de cambiar de celda, lo cual explica los crecientes resultados a medida que va aumentando el número de celdas y de acciones. Como cabría esperar el observador sigue obteniendo mejores resultados cuando no se relocalizan ni a *Good* ni a *Evil*.

## 5.2.2. Espacios generados automáticamente (Fuertemente conectados)

Veamos los resultados obtenidos utilizando una media de 100 sesiones con los distintos agentes e iteraciones utilizando distintos espacios en cada sesión teniendo como restricciones que ninguna celda quede desconectada del resto de celdas y que siempre exista la posibilidad de llegar de una celda a cualquier otra del espacio.

# 4-celdas 3-acciones

Espacio generado automáticamente (Fuertemente conectado)

Número de celdas: 4

Número de acciones: 3

Good y Evil: Comportamiento aleatorio

Media de resultados tras 100 sesiones realizadas.

| Agente<br>Iteraciones | Aleatorio | Observador | Aleatorio (Sin cambio) | Observador (Sin cambio) |
|---|---|---|---|---|
| **5** | -0.0559375 | 0.52925 | -0.054 | 0.5615 |
| **10** | 0.0311015625 | 0.616875 | 0.0290976562 | 0.63703125 |
| **20** | -0.0162795429 | 0.669 | -0.0282729492 | 0.650625 |
| **50** | -0.0041670043 | 0.6596765625 | 0.0190241519 | 0.672775 |
| **100** | 0.0059552291 | 0.6535291015 | 0.0021379867 | 0.666625 |
| **200** | 0.0028428422 | 0.6621362792 | -0.0039155030 | 0.67644375 |
| **500** | 0.0090564506 | 0.6549043872 | -0.0027577401 | 0.67529 |
| **1000** | -0.0040643431 | 0.6541745177 | -0.0019210290 | 0.67767125 |
| **2000** | -0.0016869102 | 0.6556849958 | 0.0041417218 | 0.678718125 |
| **100000** | -0.0000455535 | 0.6540757565 | -0.0007266093 | 0.6762191 |

# 4-celdas 4-acciones

Espacio generado automáticamente (Fuertemente conectado)

Número de celdas: 4

Número de acciones: 4

Good y Evil: Comportamiento aleatorio

Media de resultados tras 100 sesiones realizadas.

| Agente<br>Iteraciones | Aleatorio | Observador | Aleatorio<br>(Sin cambio) | Observador<br>(Sin cambio) |
|---|---|---|---|---|
| **5** | 0.0060625 | 0.604 | 0.0163125 | 0.585125 |
| **10** | -0.0125957031 | 0.65425 | 0.0022734375 | 0.658375 |
| **20** | -0.0071548156 | 0.675875 | 0.0328095550 | 0.6716875 |
| **50** | 0.0172693517 | 0.6737371093 | 0.0148806640 | 0.68635 |
| **100** | 0.0063840385 | 0.6824394531 | 0.0041909473 | 0.691725 |
| **200** | 0.0030967557 | 0.6686276851 | -0.0033190012 | 0.68488125 |
| **500** | -0.0017187205 | 0.6787061772 | -0.0023710905 | 0.6845 |
| **1000** | 0.0040511488 | 0.6716999725 | 0.0028339158 | 0.69140375 |
| **2000** | 0.0026429969 | 0.6824245528 | 0.0033481580 | 0.6919275 |
| **100000** | 0.0036169436 | 0.6779132341 | 0.0034483574 | 0.68647725625 |

# 6-celdas 4-acciones

Espacio generado automáticamente (Fuertemente conectado)

Número de celdas: 6

Número de acciones: 4

Good y Evil: Comportamiento aleatorio

Media de resultados tras 100 sesiones realizadas.

| Agente<br>Iteraciones | Aleatorio | Observador | Aleatorio<br>(Sin cambio) | Observador<br>(Sin cambio) |
|---|---|---|---|---|
| **5** | -0.008375 | 0.6135 | 0.0236875 | 0.5605 |
| **10** | 0.0096523437 | 0.65303125 | -0.0134296875 | 0.6638125 |
| **20** | 0.0170052347 | 0.7070625 | 0.0073968658 | 0.7104375 |
| **50** | 0.0015991142 | 0.7195690429 | -0.0071953084 | 0.726825 |
| **100** | 0.0009080716 | 0.7096931018 | 0.0102078705 | 0.74024375 |
| **200** | 0.0037127370 | 0.7114983885 | 0.0098327245 | 0.74724335937 |
| **500** | 0.0037571275 | 0.7191618014 | 0.0066783974 | 0.74959375 |
| **1000** | 0.0067209183 | 0.7128960171 | 0.0073153420 | 0.75042359375 |
| **2000** | 0.0091325779 | 0.7167080586 | 0.0072821669 | 0.7497528125 |
| **100000** | 0.0032163755 | 0.7157017915 | 0.0034025599 | 0.7522874875 |

# 6-celdas 6-acciones

Espacio generado automáticamente (Fuertemente conectado)

Número de celdas: 6

Número de acciones: 6

Good y Evil: Comportamiento aleatorio

Media de resultados tras 100 sesiones realizadas.

| Agente / Iteraciones | Aleatorio | Observador | Aleatorio (Sin cambio) | Observador (Sin cambio) |
|---|---|---|---|---|
| **5** | 0.03225 | 0.68525 | 0.047625 | 0.715 |
| **10** | -0.011875 | 0.724375 | -0.0522226562 | 0.75675 |
| **20** | -0.0195453338 | 0.76275 | 0.0059307861 | 0.7605625 |
| **50** | 0.0023880575 | 0.7547959472 | 0.0118458852 | 0.76375 |
| **100** | -0.0015145474 | 0.7639124877 | 0.0021453768 | 0.7742625 |
| **200** | 0.0006489422 | 0.7598449461 | -0.0046083085 | 0.7756625 |
| **500** | 0.0034337616 | 0.7645893438 | 0.0034215969 | 0.77374 |
| **1000** | 0.0060917931 | 0.7606200810 | 0.0028078536 | 0.7797375 |
| **2000** | 0.0023283563 | 0.7632689170 | 0.0015290917 | 0.77973375 |
| **100000** | 0.0006137602 | 0.7573598717 | 0.0014342578 | 0.779636525 |

# 8-celdas 4-acciones

Espacio generado automáticamente (Fuertemente conectado)

Número de celdas: 8

Número de acciones: 4

Good y Evil: Comportamiento aleatorio

Media de resultados tras 100 sesiones realizadas.

| Agente<br>Iteraciones | Aleatorio | Observador | Aleatorio<br>(Sin cambio) | Observador<br>(Sin cambio) |
|---|---|---|---|---|
| **5** | 0.0224375 | 0.613375 | -0.0093125 | 0.5381875 |
| **10** | 0.0230917968 | 0.676484375 | -0.0049335937 | 0.64228125 |
| **20** | -0.0042501735 | 0.7115859375 | 0.0056370620 | 0.72321875 |
| **50** | 0.0279502706 | 0.7356921047 | 0.0033554223 | 0.758615625 |
| **100** | 0.0124253130 | 0.7113351461 | 0.0156291859 | 0.772246875 |
| **200** | 0.0118206168 | 0.7252318740 | 0.0092197083 | 0.7791625 |
| **500** | 0.0070662716 | 0.7219434424 | 0.0083700529 | 0.77841421875 |
| **1000** | 0.0101475394 | 0.7241803343 | 0.0089241963 | 0.78193796875 |
| **2000** | 0.0088895304 | 0.7209748786 | 0.0047236072 | 0.778891875 |
| **100000** | 0.0073447912 | 0.7218125279 | 0.0089199839 | 0.77512321093 |

# 8-celdas 6-acciones

Espacio generado automáticamente (Fuertemente conectado)

Número de celdas: 8

Número de acciones: 6

Good y Evil: Comportamiento aleatorio

Media de resultados tras 100 sesiones realizadas.

| Agente  Iteraciones | Aleatorio | Observador | Aleatorio (Sin cambio) | Observador (Sin cambio) |
|---|---|---|---|---|
| **5** | -0.001375 | 0.7145 | -0.034375 | 0.706625 |
| **10** | 0.0178945312 | 0.74125 | 0.0070058593 | 0.7331875 |
| **20** | -0.0105775337 | 0.7750625 | -0.0031084747 | 0.765125 |
| **50** | 0.001415135 | 0.7886808349 | -0.006634225 | 0.796225 |
| **100** | 0.0019076936 | 0.7850293668 | -0.0033597725 | 0.804325 |
| **200** | -0.0056681451 | 0.7861249763 | -0.0043902148 | 0.806825 |
| **500** | 0.0023970364 | 0.7844784053 | -0.002102987 | 0.81118 |
| **1000** | 0.0001708415 | 0.7815085989 | 0.0003116397 | 0.80873875 |
| **2000** | -0.0007306984 | 0.7794557296 | 0.0006464564 | 0.811493125 |
| **100000** | 0.0002585017 | 0.7821838415 | 0.0004136819 | 0.81182739375 |

# 8-celdas 8-acciones

Espacio generado automáticamente (Fuertemente conectado)

Número de celdas: 8

Número de acciones: 8

Good y Evil: Comportamiento aleatorio

Media de resultados tras 100 sesiones realizadas.

| Agente<br>Iteraciones | Aleatorio | Observador | Aleatorio<br>(Sin cambio) | Observador<br>(Sin cambio) |
|---|---|---|---|---|
| **5** | -0.0059375 | 0.7355 | 0.0151875 | 0.78625 |
| **10** | -0.0131113281 | 0.78775 | 0.0077460937 | 0.7980625 |
| **20** | -0.0044685211 | 0.807625 | -0.0247678604 | 0.818625 |
| **50** | 0.0025694720 | 0.8138312362 | -0.0081238818 | 0.83015 |
| **100** | -0.0094017859 | 0.8167993530 | 0.0052363706 | 0.83363125 |
| **200** | 0.0004332004 | 0.8149542020 | 0.0084704842 | 0.8295125 |
| **500** | -0.0018826736 | 0.8152654870 | -0.0030245552 | 0.8297325 |
| **1000** | -0.0021321778 | 0.8156104487 | -0.0006740240 | 0.831035 |
| **2000** | 0.0016597958 | 0.8157987536 | 0.0003517454 | 0.83301125 |
| **100000** | -0.0017017071 | 0.8133519223 | -0.0008519155 | 0.83378475 |

# 10-celdas 4-acciones

Espacio generado automáticamente (Fuertemente conectado)

Número de celdas: 10

Número de acciones: 4

Good y Evil: Comportamiento aleatorio

Media de resultados tras 100 sesiones realizadas.

| Agente<br>Iteraciones | Aleatorio | Observador | Aleatorio<br>(Sin cambio) | Observador<br>(Sin cambio) |
|---|---|---|---|---|
| **5** | -0.00775 | 0.545875 | -0.003375 | 0.529125 |
| **10** | -0.0133144531 | 0.6095546875 | 0.0135009765 | 0.62215625 |
| **20** | 0.0223903026 | 0.7179296875 | 0.0149272508 | 0.681453125 |
| **50** | 0.0086390515 | 0.7318503353 | -0.0141471834 | 0.7633 |
| **100** | 0.0130662086 | 0.7185673772 | 0.0061895334 | 0.77265546875 |
| **200** | 0.0018630807 | 0.7143527745 | 0.0076909615 | 0.77997851562 |
| **500** | -0.0006402785 | 0.7220285703 | 0.0063592927 | 0.790005625 |
| **1000** | 0.0026376552 | 0.7135880569 | -0.0015747931 | 0.79191449218 |
| **2000** | 0.0047512678 | 0.7179938130 | 0.0014539369 | 0.789928125 |
| **100000** | 0.0031840408 | 0.7188911904 | 0.0025455900 | 0.797118975 |

# 10-celdas 7-acciones

Espacio generado automáticamente (Fuertemente conectado)

Número de celdas: 10

Número de acciones: 7

Good y Evil: Comportamiento aleatorio

Media de resultados tras 100 sesiones realizadas.

| Agente  Iteraciones | Aleatorio | Observador | Aleatorio (Sin cambio) | Observador (Sin cambio) |
|---|---|---|---|---|
| **5** | 0.0315625 | 0.722 | -0.056125 | 0.70175 |
| **10** | 0.0343896484 | 0.769375 | 0.0120039062 | 0.78725 |
| **20** | -0.0026242761 | 0.79628125 | 0.0078400764 | 0.80175 |
| **50** | 0.0017691624 | 0.8128048820 | -0.0073491173 | 0.831412 |
| **100** | -0.0023990369 | 0.8117279079 | 0.0019531691 | 0.83767 |
| **200** | 0.0000650018 | 0.8123028149 | 0.0005083199 | 0.84008828125 |
| **500** | 0.0005450867 | 0.8121905251 | 0.0008084398 | 0.8453025 |
| **1000** | -0.0019673875 | 0.8136747984 | 0.0000549403 | 0.8451721875 |
| **2000** | 0.0007466091 | 0.8102686322 | 0.0004461757 | 0.845355625 |
| **100000** | -0.0000180668 | 0.8137150793 | -0.0001898676 | 0.84364703125 |

# 10-celdas 10-acciones

Espacio generado automáticamente (Fuertemente conectado)

Número de celdas: 10

Número de acciones: 10

Good y Evil: Comportamiento aleatorio

Media de resultados tras 100 sesiones realizadas.

| Agente<br>Iteraciones | Aleatorio | Observador | Aleatorio<br>(Sin cambio) | Observador<br>(Sin cambio) |
|---|---|---|---|---|
| 5 | -0.0226875 | 0.7615 | -0.0031875 | 0.775 |
| 10 | 0.0153847656 | 0.816875 | -0.0310585937 | 0.8255 |
| 20 | -0.0223733577 | 0.849765625 | 0.0195264005 | 0.8361875 |
| 50 | -0.0142524479 | 0.8452241943 | -0.0022178101 | 0.86205 |
| 100 | -0.0032213712 | 0.8499759917 | -0.0010602513 | 0.8656125 |
| 200 | 0.0033599156 | 0.8477624884 | 0.0016957615 | 0.8649625 |
| 500 | -0.0024712512 | 0.8475952018 | 0.0012036416 | 0.867875 |
| 1000 | -0.0002777197 | 0.8477101014 | 0.0013118449 | 0.8696825 |
| 2000 | 0.0005061001 | 0.8487137980 | 0.0018370838 | 0.8685675 |
| 100000 | -0.0001684560 | 0.8475726295 | -0.0002814336 | 0.8673165875 |

El agente aleatorio sigue la misma línea que el resto de espacios vistos hasta el momento, por lo cual los espacios generados fuertemente conectados siguen cumpliendo las propiedades de un entorno balanceado.

Con respecto al agente observador, sigue exactamente la misma dinámica que en los experimentos realizados anteriormente con los espacios conectados generados. Si nos fijamos en estos resultados y los obtenidos con los espacios generados conectados podemos ver como son prácticamente iguales. Esto se debe a que al permitir que se generen tanta cantidad de acciones, los espacios que se crean raramente tienen un pequeño conjunto de estados en donde el agente evaluable pueda quedarse encerrado durante toda la sesión. De hecho, haciendo pruebas con esta distribución del espacio (10 celdas y 10 acciones) teniendo como restricción que los espacios generados sean conectados, únicamente 3 de los 2000 espacios que se han generado no son también fuertemente conectados, lo que explica la gran similitud entre los resultados obtenidos.

## 5.3. Entorno sesgado

A continuación veremos los resultados a experimentos realizados en el espacio definido de 4 celdas utilizando comportamientos distintos para el agente *Good* y para *Evil*.

# Entorno sesgado 1

Entorno sesgado

Número de celdas: 4

Número de acciones: 3

Good y Evil: Comportamiento aleatorio

Visión gráfica

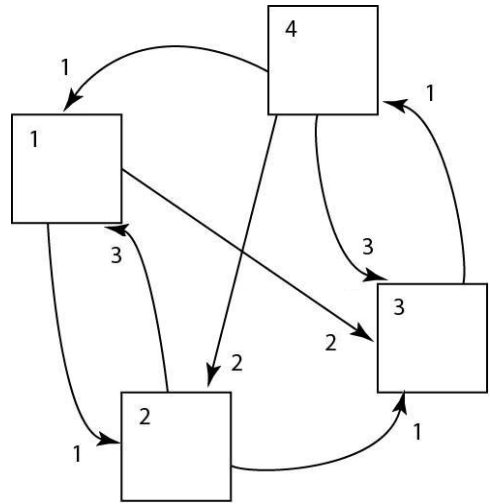

Media del resultado tras 10 sesiones realizadas donde *Good* se mantiene siempre en la misma celda.

| Agente<br>Iteraciones | Aleatorio | Aleatorio<br>(Sin cambio) |
|---|---|---|
| **5** | 0.0275 | -0.0925 |
| **10** | -0.0853125 | -0.0834570312 |
| **20** | -0.0733984375 | -0.1013476943 |
| **50** | -0.0744572753 | -0.0545395507 |
| **100** | -0.0721672363 | -0.080703125 |
| **200** | -0.0798338012 | -0.1081768798 |
| **500** | -0.0763609828 | -0.0759708496 |
| **1000** | -0.0705885308 | -0.0916902229 |
| **2000** | -0.0833871496 | -0.0610014770 |
| **1000000** | -0.0864727326 | -0.0988842842 |

En estos experimentos el agente aleatorio siempre compartirá las recompensas positivas con el agente *Good* y seguirá recogiendo las recompensas negativas como anteriormente hacía. Esto hace que la media de recompensas recogidas siempre sea negativa.

# Entorno sesgado 2

Entorno sesgado

Número de celdas: 4

Número de acciones: 3

Good y Evil: Comportamiento aleatorio

Visión gráfica

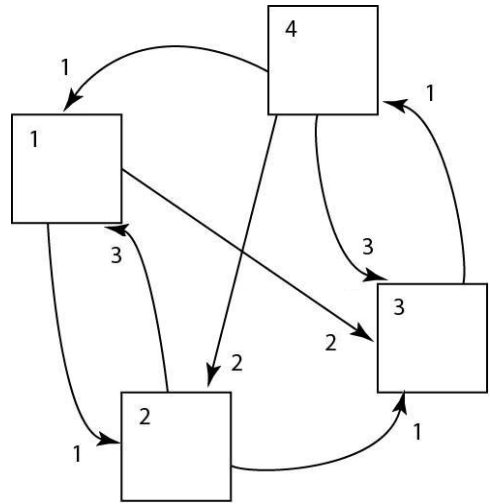

Media del resultado tras 10 sesiones realizadas donde *Good* siempre trata de cambiar de celda.

| Agente \ Iteraciones | Aleatorio | Aleatorio (Sin cambio) |
|---|---|---|
| 5 | 0.01875 | 0.21125 |
| 10 | 0.095625 | 0.21125 |
| 20 | 0.072265625 | 0.0569726562 |
| 50 | -0.0179550781 | 0.0548925781 |
| 100 | 0.0143828125 | 0.0466845703 |
| 200 | 0.0472050781 | 0.0500693359 |
| 500 | 0.0506410400 | 0.0425306884 |
| 1000 | 0.0511209838 | 0.0387437622 |
| 2000 | 0.0498806686 | 0.0543033935 |
| 1000000 | 0.0523993359 | 0.0546207446 |

En este caso el agente *Good* siempre tratará de cambiar de celda, a menos que se encuentre con *Evil* en cuyo caso uno de los dos tendrá que esperar en su celda, por lo que generalmente el agente no compartirá la recompensa con el agente *Good*.

Como podemos ver en ambos entornos existe un sesgo con respecto a los resultados obtenidos por el agente aleatorio, lo cual significa que no se cumplen las propiedades de un entorno balanceado. Esto es debido a que no se han introducido elementos complementarios en el entorno (el agente *Good* y el agente *Evil* no se comportan exactamente igual), siendo predominante en cada uno de los experimentos uno de los elementos.

## 5.4. Evaluación social

En este caso vamos a medir a varios agentes evaluándose simultáneamente en la misma sesión, por lo que podremos ver como se ven modificados los resultados de los distintos agentes al existir otros compitiendo por las recompensas durante el transcurso de las sesiones. En estos experimentos introduciremos a los agentes aleatorio y observador simultáneamente en el entorno.

En la tabla vemos los resultados por parejas, donde cada resultado de Aleatorio y de Observador será el resultado para cada uno al evaluarlos sobre las mismas sesiones.

# Espacio definido manualmente - 8 celdas

Evaluación social - Espacio definido manualmente

Número de celdas: 8

Número de acciones: 3

Good y Evil: Comportamiento aleatorio

Visión gráfica

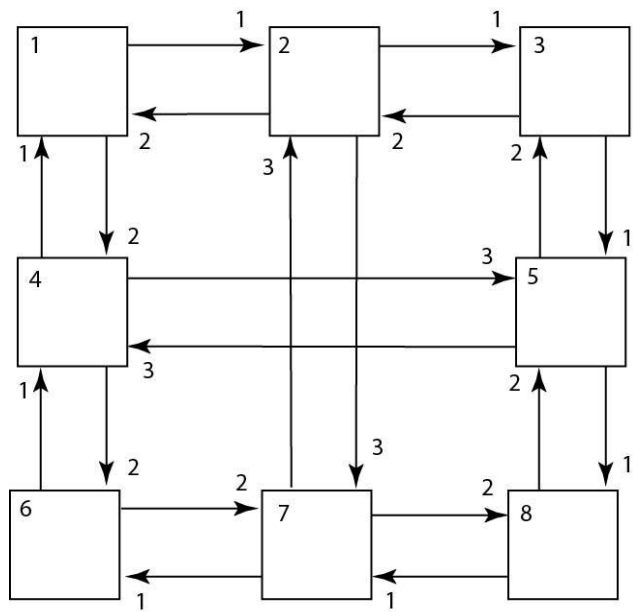

Media del resultado tras 10 sesiones realizadas.

| Agente<br>Iteraciones | Aleatorio | Observador | Aleatorio (Sin cambio) | Observador (Sin cambio) |
|---|---|---|---|---|
| **5** | -0.0925 | 0.5316666666 | 0.0316666666 | 0.365 |
| **10** | -0.0190104166 | 0.61 | -0.0347135416 | 0.61541666666 |
| **20** | -0.0845389811 | 0.7216666666 | -0.1060742187 | 0.61833333333 |
| **50** | -0.0875097198 | 0.7161197915 | -0.0821009115 | 0.71566666666 |
| **100** | -0.0736087347 | 0.7053124389 | -0.0480256779 | 0.74825 |
| **200** | -0.0784876931 | 0.6839524720 | -0.0745241745 | 0.741 |
| **500** | -0.0781876082 | 0.6871417968 | -0.0746585093 | 0.73788333333 |
| **1000** | -0.0810844671 | 0.6835628865 | -0.0829523893 | 0.73548333333 |
| **2000** | -0.0724479806 | 0.6936914750 | -0.0688205723 | 0.74119895833 |
| **1000000** | -0.0713102075 | 0.6929307734 | -0.0793472724 | 0.74182583333 |

# Espacio generado automáticamente (Conectado)

# 8-celdas 6-acciones

Evaluación social - Espacio generado automáticamente (Conectado)

Número de celdas: 8

Número de acciones: 6

Good y Evil: Comportamiento aleatorio

Media del resultado evaluando 100 sesiones realizadas con espacios generados al azar.

| Agente<br>Iteraciones | Aleatorio | Observador | Aleatorio (Sin cambio) | Observador (Sin cambio) |
|---|---|---|---|---|
| 5 | -0.0634583333 | 0.6508333333 | -0.0559583333 | 0.62533333333 |
| 10 | -0.0872135416 | 0.6813958333 | -0.0926223958 | 0.688625 |
| 20 | -0.1037266680 | 0.7412083333 | -0.0984965496 | 0.7297916666 |
| 50 | -0.0785706488 | 0.7436425781 | -0.0929490238 | 0.743475 |
| 100 | -0.0998179061 | 0.7346942586 | -0.1027326131 | 0.7559 |
| 200 | -0.0921733936 | 0.7361930467 | -0.0953013076 | 0.7541885416 |
| 500 | -0.0939403367 | 0.7300403093 | -0.0977293596 | 0.754136666 |
| 1000 | -0.0931912955 | 0.7300650927 | -0.0982186178 | 0.75964125 |
| 2000 | -0.0928543908 | 0.7290798694 | -0.0985127970 | 0.75536802083 |
| 100000 | -0.0942372888 | 0.7236268737 | -0.0982732850 | 0.74866333541 |

# Espacio generado automáticamente (Fuertemente conectado)

# 8-celdas 6-acciones

Evaluación social - Espacio generado automáticamente (Fuertemente conectado)

Número de celdas: 8

Número de acciones: 6

Good y Evil: Comportamiento aleatorio

Media del resultado evaluando 100 sesiones realizadas con espacios generados al azar.

| Agente<br>Iteraciones | Aleatorio | Observador | Aleatorio (Sin cambio) | Observador (Sin cambio) |
|---|---|---|---|---|
| 5 | -0.0839166666 | 0.657 | -0.0596875 | 0.62591666666 |
| 10 | -0.0926793619 | 0.7100416666 | -0.0718736979 | 0.6895 |
| 20 | -0.0907177225 | 0.7303958333 | -0.0816079489 | 0.7171875 |
| 50 | -0.0822846768 | 0.7236609375 | -0.0895003887 | 0.74218333333 |
| 100 | -0.0889639014 | 0.7329951766 | -0.1074970967 | 0.7507125 |
| 200 | -0.0911916412 | 0.7366449244 | -0.0916739162 | 0.75441666666 |
| 500 | -0.0947957406 | 0.7306393021 | -0.0982189874 | 0.75883 |
| 1000 | -0.0932831643 | 0.7326538204 | -0.0987799716 | 0.75638041666 |
| 2000 | -0.0927858311 | 0.7291214223 | -0.0969757873 | 0.7553403125 |
| 100000 | -0.0938490195 | 0.7351404194 | -0.0975634640 | 0.758923725 |

Como podemos observar la introducción de un agente observador obliga al agente aleatorio a recoger mayoritariamente recompensas negativas y a compartir las positivas con el observador, por lo que sus resultados son siempre negativos.

Sin embargo, el agente observador también resulta perjudicado al incluir al agente aleatorio, ya que de vez en cuando también debe compartir las recompensas positivas con el agente aleatorio, lo cual se traduce en una disminución de los resultados con respecto a los mismos experimentos evaluación social.

## 5.5. Varios movimientos de los agentes generadores

Hasta el momento hemos realizado los experimentos teniendo en cuenta que los agentes generadores *Good* y *Evil* solo podían moverse de una casilla a otra adyacente, sin embargo estos agentes deberían poder moverse varias celdas simultáneamente. En estos experimentos hemos dado a los agentes *Good* y *Evil* la oportunidad de moverse realizando 2, 3 y 4 acciones simultáneas a través del espacio.

# 2 Movimientos

# Espacio definido manualmente - 8 celdas

2 Movimientos – Espacio definido manualmente

Número de celdas: 8

Número de acciones: 3

Good y Evil: Comportamiento aleatorio

Visión gráfica

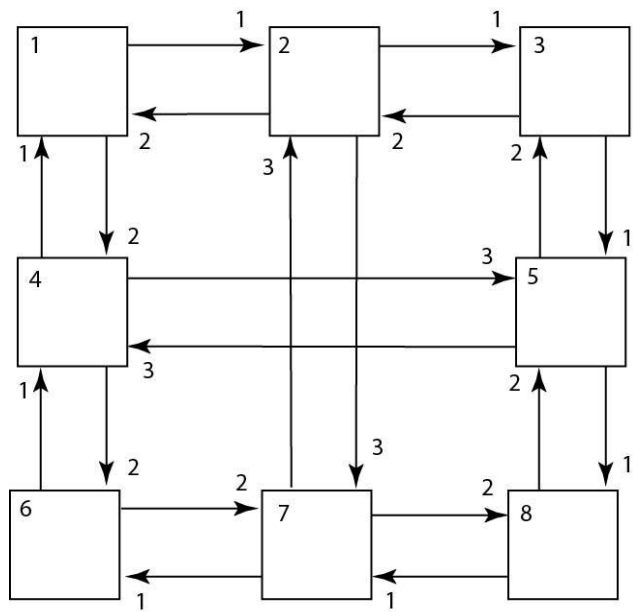

Media del resultado tras 10 sesiones realizadas.

| Iteraciones \ Agente | Aleatorio | Observador | Aleatorio (Sin cambio) | Observador (Sin cambio) |
|---|---|---|---|---|
| **5** | -0.04 | 0.405 | -0.05 | 0.42625 |
| **10** | 0.03796875 | 0.3821875 | 0.0153515625 | 0.412578125 |
| **20** | -0.0363085937 | 0.4236915588 | -0.0047903442 | 0.51425292968 |
| **50** | 0.0097206726 | 0.5009875142 | -0.0005978928 | 0.55971380615 |
| **100** | -0.0066461486 | 0.4774788761 | -0.0101414661 | 0.51140614509 |
| **200** | -0.0080702943 | 0.4481070556 | 0.0126746131 | 0.49184478485 |
| **500** | -0.0166377544 | 0.4458106359 | 0.0032025186 | 0.50211970045 |
| **1000** | -0.0026152616 | 0.4622730091 | -0.0045706541 | 0.49065660272 |
| **2000** | 0.0006304102 | 0.4678949345 | -0.0016159894 | 0.48975005375 |
| **1000000** | -0.0007433638 | 0.4675208032 | -0.0006717888 | 0.47645791772 |

# 2 Movimientos

# Espacio generado automáticamente (Conectado)

# 8-celdas 6-acciones

2 Movimientos – Espacio generado automáticamente (Conectado)

Número de celdas: 8

Número de acciones: 6

Good y Evil: Comportamiento aleatorio

Media del resultado evaluando 100 sesiones realizadas con espacios generados al azar.

| Agente<br>Iteraciones | Aleatorio | Observador | Aleatorio (Sin cambio) | Observador (Sin cambio) |
|---|---|---|---|---|
| 5 | -0.0119375 | 0.5341875 | -0.0072 | 0.55075 |
| 10 | 0.0064091796 | 0.5379550781 | -0.0058457031 | 0.53960351562 |
| 20 | 0.0082336196 | 0.5430779724 | -0.0035347251 | 0.56109193801 |
| 50 | 0.006022677 | 0.5298084959 | -0.0000125740 | 0.55853964548 |
| 100 | -0.0081134941 | 0.5518187971 | -0.0095100618 | 0.55345392186 |
| 200 | -0.0064818422 | 0.5524241160 | -0.0036136333 | 0.54766000853 |
| 500 | -0.0043393588 | 0.5505275479 | -0.0027577104 | 0.55116168168 |
| 1000 | 0.0007476604 | 0.5445017800 | 0.0016774965 | 0.54745065584 |
| 2000 | -0.0002261323 | 0.5371404278 | 0.0006449761 | 0.54885647237 |
| 100000 | 0.0000726315 | 0.5445200320 | -0.0006092810 | 0.54809953458 |

# 2 Movimientos

# Espacio generado automáticamente (Fuertemente conectado)

# 8-celdas 6-acciones

2 Movimientos – Espacio generado automáticamente (Fuertemente conectado)

Número de celdas: 8

Número de acciones: 6

Good y Evil: Comportamiento aleatorio

Media del resultado evaluando 100 sesiones realizadas con espacios generados al azar.

| Agente<br>Iteraciones | Aleatorio | Observador | Aleatorio (Sin cambio) | Observador (Sin cambio) |
|---|---|---|---|---|
| **5** | -0.006875 | 0.538125 | -0.0324375 | 0.572062 |
| **10** | 0.0100273437 | 0.5299609375 | 0.0036308593 | 0.5477578125 |
| **20** | 0.0226711483 | 0.5366146240 | -0.002918045 | 0.54955162048 |
| **50** | 0.0077164861 | 0.5502106540 | -0.0006324388 | 0.54632609325 |
| **100** | -0.0046460433 | 0.5414063711 | -0.0064156002 | 0.53228034219 |
| **200** | 0.0003139742 | 0.5421837545 | 0.0018089704 | 0.55251558018 |
| **500** | -0.0000909382 | 0.5393919641 | 0.0001765426 | 0.55325090242 |
| **1000** | 0.000387935 | 0.5465902676 | 0.0000794154 | 0.54417609374 |
| **2000** | -0.0005180299 | 0.5382306384 | -0.0020435247 | 0.55555706237 |
| **100000** | -0.0003742777 | 0.5446807327 | -0.0004220987 | 0.55247444172 |

# 3 Movimientos

# Espacio definido manualmente - 8 celdas

3 Movimientos – Espacio definido manualmente

Número de celdas: 8

Número de acciones: 3

Good y Evil: Comportamiento aleatorio

Visión gráfica

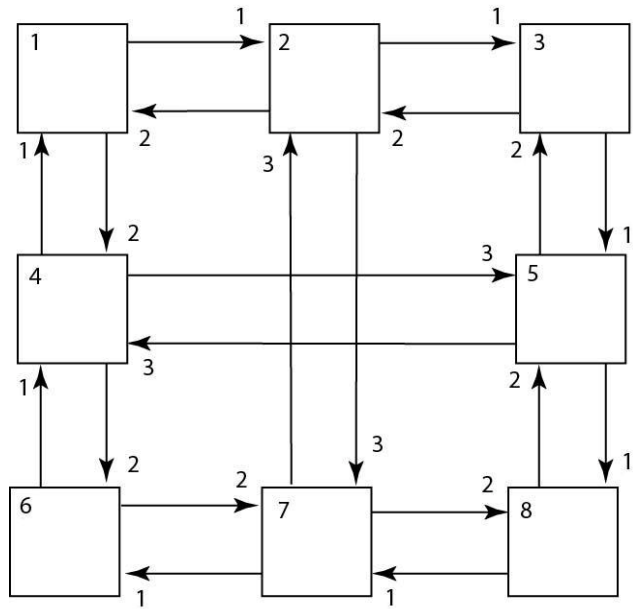

Media del resultado tras 10 sesiones realizadas.

| Agente<br>Iteraciones | Aleatorio | Observador | Aleatorio<br>(Sin cambio) | Observador<br>(Sin cambio) |
|---|---|---|---|---|
| **5** | 0.015 | 0.3075 | 0.0675 | 0.525 |
| **10** | 0.0778125 | 0.495 | 0.0000195312 | 0.4809765625 |
| **20** | -0.0155657958 | 0.3947705078 | -0.0081640625 | 0.48546264648 |
| **50** | -0.0277063903 | 0.4672634277 | -0.0041567382 | 0.466694458 |
| **100** | -0.0123266344 | 0.4584087065 | -0.008754115 | 0.47171520423 |
| **200** | 0.0026176177 | 0.4822813549 | -0.0004651916 | 0.44741162168 |
| **500** | -0.0020008018 | 0.4646291565 | 0.0111973494 | 0.45438940498 |
| **1000** | 0.0027786851 | 0.4469535351 | -0.0055641521 | 0.45812849095 |
| **2000** | -0.0051659124 | 0.4550553326 | 0.0015448156 | 0.45724515918 |
| **1000000** | 0.0013313368 | 0.4527268877 | 0.0021513511 | 0.45961500528 |

# 3 Movimientos

# Espacio generado automáticamente (Conectado)

# 8-celdas 6-acciones

3 Movimientos – Espacio generado automáticamente (Conectado)

Número de celdas: 8

Número de acciones: 6

Good y Evil: Comportamiento aleatorio

Media del resultado evaluando 100 sesiones realizadas con espacios generados al azar.

| Agente<br>Iteraciones | Aleatorio | Observador | Aleatorio (Sin cambio) | Observador (Sin cambio) |
|---|---|---|---|---|
| **5** | -0.0219375 | 0.49325 | 0.0156875 | 0.4818125 |
| **10** | 0.008234375 | 0.5291503906 | 0.0188515625 | 0.52532421875 |
| **20** | -0.0003603782 | 0.5175086669 | 0.0001126976 | 0.51278277587 |
| **50** | 0.0021808862 | 0.5166270456 | -0.0115737323 | 0.51555128552 |
| **100** | 0.001067267 | 0.5179938519 | -0.0027140024 | 0.52435035397 |
| **200** | 0.0020297547 | 0.5180077101 | -0.0039365435 | 0.51644639428 |
| **500** | 0.0021116354 | 0.5238462684 | 0.0019011115 | 0.51839674259 |
| **1000** | 0.0041763178 | 0.5146958811 | -0.0013714009 | 0.5152991525 |
| **2000** | 0.0002451868 | 0.5107994995 | 0.0006154981 | 0.52274517534 |
| **100000** | 0.0002321834 | 0.5141149108 | 0.0007630815 | 0.5186971441 |

# 3 Movimientos

## Espacio generado automáticamente (Fuertemente conectado)

## 8-celdas 6-acciones

3 Movimientos – Espacio generado automáticamente (Fuertemente conectado)

Número de celdas: 8

Número de acciones: 6

Good y Evil: Comportamiento aleatorio

Media del resultado evaluando 100 sesiones realizadas con espacios generados al azar.

| Agente Iteraciones | Aleatorio | Observador | Aleatorio (Sin cambio) | Observador (Sin cambio) |
|---|---|---|---|---|
| 5 | -0.002375 | 0.5063125 | -0.025625 | 0.4895 |
| 10 | 0.0151914062 | 0.5172207031 | -0.0110566406 | 0.54356054687 |
| 20 | -0.0289168739 | 0.5175667724 | 0.003539978 | 0.53750369262 |
| 50 | 0.004336184 | 0.5057879984 | -0.0066616579 | 0.50712485746 |
| 100 | 0.0048225043 | 0.5112355104 | 0.0027196029 | 0.51155243528 |
| 200 | 0.0025973139 | 0.5132000347 | -0.0041791193 | 0.51492348993 |
| 500 | 0.0011507622 | 0.5113789919 | 0.0018582105 | 0.51214954175 |
| 1000 | -0.0010128764 | 0.5093485237 | 0.0011275919 | 0.52257547003 |
| 2000 | 0.0025834348 | 0.5034636958 | 0.0011110905 | 0.51276348561 |
| 100000 | 0.0005249084 | 0.5076676257 | -0.0001681908 | 0.51862807874 |

# 4 Movimientos

# Espacio definido manualmente - 8 celdas

4 Movimientos – Espacio definido manualmente

Número de celdas: 8

Número de acciones: 3

Good y Evil: Comportamiento aleatorio

Visión gráfica

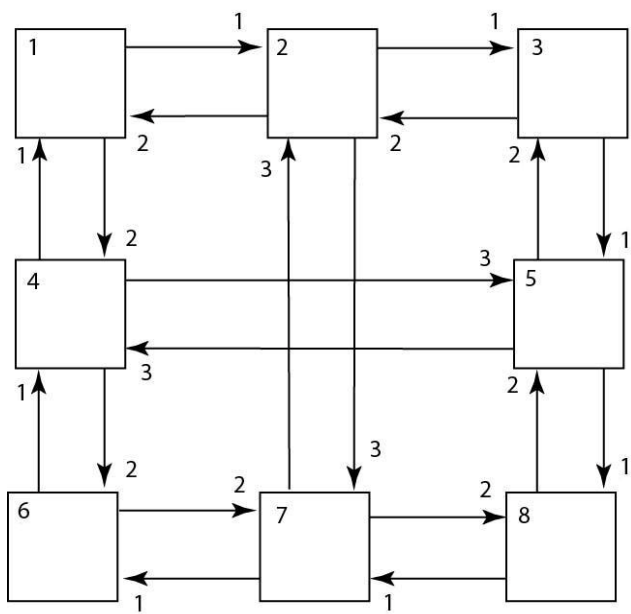

Media del resultado tras 10 sesiones realizadas.

| Agente<br>Iteraciones | Aleatorio | Observador | Aleatorio<br>(Sin cambio) | Observador<br>(Sin cambio) |
|---|---|---|---|---|
| **5** | 0.03625 | 0.295 | -0.02 | 0.4325 |
| **10** | 0.04359375 | 0.40609375 | -0.0296875 | 0.39453125 |
| **20** | -0.0076364135 | 0.3544236755 | 0.0228564453 | 0.43929321289 |
| **50** | 0.0297216796 | 0.4356188964 | 0.0276926269 | 0.45490655517 |
| **100** | 0.0031295781 | 0.4159604034 | 0.0263404192 | 0.41855424499 |
| **200** | -0.0018015873 | 0.4172509078 | 0.0131416435 | 0.41539114761 |
| **500** | -0.0047641256 | 0.4048449179 | 0.000372763 | 0.41228777224 |
| **1000** | 0.0091366043 | 0.4071086136 | 0.0001288413 | 0.40876417505 |
| **2000** | 0.0003983884 | 0.4061799454 | -0.0062200826 | 0.40967137190 |
| **1000000** | 0.0000415605 | 0.4042661129 | 0.0000581977 | 0.40725332197 |

# 4 Movimientos

# Espacio generado automáticamente (Conectado)

# 8-celdas 6-acciones

4 Movimientos – Espacio generado automáticamente (Conectado)

Número de celdas: 8

Número de acciones: 6

Good y Evil: Comportamiento aleatorio

Media del resultado evaluando 100 sesiones realizadas con espacios generados al azar.

| Agente  Iteraciones | Aleatorio | Observador | Aleatorio (Sin cambio) | Observador (Sin cambio) |
|---|---|---|---|---|
| **5** | -0.019125 | 0.437375 | 0.0034375 | 0.4781875 |
| **10** | -0.0234580078 | 0.5111425781 | -0.0145878906 | 0.49786328125 |
| **20** | 0.0228353652 | 0.5105960083 | -0.0145878906 | 0.49482608032 |
| **50** | -0.0199328547 | 0.4969350477 | -0.0145878906 | 0.51526614642 |
| **100** | -0.0081563807 | 0.5143437625 | -0.0029857476 | 0.49809998298 |
| **200** | 0.0014267878 | 0.5008350189 | 0.0020672845 | 0.50130601562 |
| **500** | -0.0018580099 | 0.5021560366 | -0.000277568 | 0.50506489835 |
| **1000** | -0.0005966052 | 0.5052771370 | 0.0009364216 | 0.51094270662 |
| **2000** | -0.0000055866 | 0.5043260169 | -0.0014542887 | 0.51026966205 |
| **100000** | -0.0004186722 | 0.5009204262 | -0.0005286776 | 0.50773962979 |

# 4 Movimientos

# Espacio generado automáticamente (Fuertemente conectado)

# 8-celdas 6-acciones

4 Movimientos – Espacio generado automáticamente (Fuertemente conectado)

Número de celdas: 8

Número de acciones: 6

Good y Evil: Comportamiento aleatorio

Media del resultado evaluando 100 sesiones realizadas con espacios generados al azar.

| Agente<br>Iteraciones | Aleatorio | Observador | Aleatorio<br>(Sin cambio) | Observador<br>(Sin cambio) |
|---|---|---|---|---|
| **5** | 0.003125 | 0.49075 | 0.0433125 | 0.4849375 |
| **10** | 0.0253671875 | 0.515 | -0.0003730468 | 0.4478125 |
| **20** | 0.0045796928 | 0.502413928 | 0.0094997062 | 0.51073068237 |
| **50** | 0.0153383504 | 0.5048348373 | -0.0027522481 | 0.49465015575 |
| **100** | -0.0094204826 | 0.496507526 | -0.0097997535 | 0.50584293028 |
| **200** | -0.0042613635 | 0.4981908917 | 0.0006317677 | 0.50560896818 |
| **500** | -0.0029455798 | 0.5028718157 | -0.0016293615 | 0.50808030833 |
| **1000** | -0.0006098342 | 0.5102287993 | -0.002906116 | 0.49683524799 |
| **2000** | 0.0008494687 | 0.5019137961 | -0.0001656485 | 0.51116060912 |
| **100000** | -0.0004016026 | 0.5041956411 | -0.0002865026 | 0.50274058414 |

Como podemos observar el agente aleatorio no ve modificados sus resultados, ya que el cambio que se ha realizado ha sido en ambos agentes generadores (*Good* y *Evil*) y, por lo tanto, se siguen manteniendo las propiedades de un entorno balanceado.

Al agente observador sí que le afecta este cambio, ya que ahora no le resulta tan fácil seguir a *Good* a través del espacio. Mientras que antes tenía que seguir continuamente a *Good* durante toda la sesión ahora tiene que realizar movimientos aleatorios hasta encontrarle y seguramente le volverá a perder de vista al poco tiempo, por lo que no le resulta tan fácil conseguir tan buenos resultados en este tipo de entornos. También podemos ver como conforme aumenta el número de movimientos que pueden realizar *Good* y *Evil* al observador le resulta más y más difícil conseguir buenos resultados. Esto se debe a que existen más posibilidades de que *Good* se aleje lo suficiente del observador como para que éste no pueda seguirle en vez de mantenerse a una distancia lo suficientemente cercana como para que continúe siguiéndole.

# 6. Conclusiones y trabajo futuro

<u>Conclusiones</u>

Podemos analizar los resultados de este proyecto de fin de carrera conforme a dos perspectivas diferentes: el cumplimiento de los objetivos marcados y el conocimiento que se ha adquirido a partir del desarrollo del proyecto y de los experimentos.

1. Tras la realización del proyecto se han cumplido todos los objetivos propuestos en el apartado 1.3.
   - Tras la construcción de la arquitectura del sistema, éste está preparado para realizar los tests propuestos a distintos tipos de agentes: Los experimentos realizados demuestran que el sistema y su arquitectura son lo suficientemente potentes como para ser utilizado para evaluar agentes. El hecho de que el resultado medio obtenido por agentes aleatorios, en los entornos tanto definidos como autogenerados, sea cercano a 0 demuestra que la arquitectura construida está preparada para realizar tests en entornos balanceados y, por lo tanto, está listo para la evaluación de la inteligencia en agentes.
   - El sistema permite la construcción de los entornos de forma manual: El sistema es capaz de construir y preparar un entorno desde cero a partir de la descripción del entorno que el usuario desee construir.
   - El sistema permite la construcción de los entornos de forma automática y aleatoria: El sistema es capaz de construir y preparar un entorno desde cero autogenerando la descripción que tendrá el entorno y posteriormente construyéndolo a partir de éste. La autogeneración puede incluir la comprobación de ciertas propiedades que el entorno debe cumplir para ser un entorno válido.
   - El sistema cuenta con una interfaz de usuario que permite la evaluación para agentes humanos: Se ha construido una interfaz que permite que los agentes humanos sean evaluados dentro del sistema, así como mostrar el resultado obtenido por este al finalizar la prueba. Además, la interfaz también permite que el usuario observe el desarrollo de pruebas en todo tipo de agentes.
   - Las pruebas y experimentos realizados han sido de ayuda para evaluar el sistema construido: Tras realizar las pruebas y el estudio de los experimentos hemos podido ver que el sistema cumple con las propiedades inicialmente deseadas y hemos podido observar la inteligencia de varios tipos de agentes.

2. Con la generación de entornos sencillos y generadores de recompensas aleatorios hemos podido ver la diferencia entre agentes aleatorios y agentes con cierta estrategia. También hemos visto su interrelación cuando varios agentes compiten por las mismas recompensas.
   - Tras lo visto durante los experimentos realizados hemos podido observar como cambios en el comportamiento de los agentes que tratamos de evaluar suponen cambios significativos en su inteligencia. Tal y como hemos podido ver un comportamiento aleatorio de los agentes no supone inteligencia alguna, ya que no trata de resolver el problema, únicamente se mueve aleatoriamente sin importarle los resultados a obtener. Por otro lado el simple hecho de ver las casillas adyacentes y, por lo tanto, poder decidir moverse a la mejor celda posible demuestra un aumento significativo de inteligencia como hemos podido observar en los experimentos. De lo que podemos deducir que si añadimos más comportamiento inteligente a los agentes resultará en una mejora en su evaluación.
- La interrelación de varios agentes en el mismo entorno entorpece la evaluación del agente a evaluar en lo que a recompensas se refiere, ya que hace disminuir su recompensa media conforme se encuentren simultáneamente más agentes en el entorno. Según lo visto, la compartición de las recompensas entre todos los agentes no resulta adecuado, por lo que se deberá estudiar otra forma de introducir nuevos agentes sin que, por ello, entorpezca la evaluación del agente que se esté evaluando.

Trabajo futuro

A la vista de todo lo anterior, pensamos que la arquitectura supone una base de evaluación sobre la cual iremos refinando y completando la variedad y complejidad de entornos y de agentes, así como el diseño de experimentos.

En concreto, el trabajo futuro se centrará en:

- Generar los espacios siguiendo una distribución universal: De momento los espacios se generan siguiendo una distribución uniforme. Los espacios deberán generarse siguiendo una distribución universal, para acercarse más a la noción intuitiva de complejidad.
- Generar objetos: La generación de los entornos deberá permitir la generación automática de los objetos.
- Generar otros agentes y su comportamiento: El entorno deberá poder generar automáticamente nuevos agentes y su comportamiento utilizando para ello el lenguaje de especificación descrito en la sección 3.5.5.
- Generar entornos automáticamente siguiendo alguna distribución: Se deberán poder generar y construir automáticamente todos los elementos de los entornos (espacios, objetos y agentes).
- Construir la observación del entorno siguiendo el lenguaje de especificación descrito en la sección 3.5.3: De momento a los agentes se les entrega la observación como una copia del espacio utilizado durante la prueba, se deberá construir una tira de caracteres que represente a las observaciones para entregárselas a los agentes.
- Modificar la interfaz para que se ajuste a las interfaces mostradas en el apartado 3.5.2: Se deberá completar el desarrollo de la interfaz y también se insertarán nuevas funcionalidades.
- Construir los tests a partir de sesiones: Se deberán crear los tests a partir de las sesiones ya construidas.

- Autoajustar la complejidad del entorno para cada sesión en función de los resultados obtenidos por el agente que se está evaluando: Para la evaluación correcta de los agentes se deberá poder autoajustar la complejidad de los entornos generados para cada sesión, siguiendo el algoritmo en [Hernandez-Orallo & Dowe 2010].
- Evaluar personas y animales.
- Evaluar sistemas de IA, como las variantes de AIXI [Veness et al 2009] o agentes con técnicas de Q-learning u otros.

# Apéndices

Debido a que el código fuente de este proyecto debe ser portable y ejecutable en cualquier máquina, preferiblemente como Applet Web, la aplicación ha sido desarrollada en JAVA utilizando el entorno Eclipse. Los comentarios se han realizado en inglés, por si en el futuro parte del proyecto es utilizado o extendido por otros equipos investigadores.

## A1. Extractos representativos del Código Fuente

En este apéndice incluimos dos partes relevantes del código fuente; la que se encarga del bucle principal de la interacción entre entorno y agente; y la generación aleatoria de entornos.

### A1.1. Bucle principal

```java
    // Loop to interact between the Environment and the Agent
    for (int actualInteraction = 0; actualInteraction <
numberOfInteractions; actualInteraction++)
    {
        if (numberOfInteractionsToRelocate != 0 && actualInteraction != 0
&& actualInteraction % numberOfInteractionsToRelocate == 0)
            this.RelocateGoodAndEvilAgents();

        // Set the rewards to the cells where are the Good and Evil Agents
        Cell goodCell = this.goodAgent.GetLocation();
        Cell evilCell = this.evilAgent.GetLocation();
        goodCell.SetReward(this.maxReward);
        evilCell.SetReward(this.minReward);

        for (int actualAgent = 0; actualAgent < this.agents.size();
actualAgent++)
        {
            Agent agent = this.agents.get(actualAgent);

            // Sends the last Reward to the agent
            double reward = interactions[actualAgent].GetReward();
            this.SendReward(agent, reward);
        }

        Observation observation = this.TakeObservation();
        for (int actualAgent = 0; actualAgent < this.agents.size();
actualAgent++)
        {
            Agent agent = this.agents.get(actualAgent);

            // Shows the actual Observation to the agent
            interactions[actualAgent].SetObservation(observation);
            this.SendObservation(agent, observation);
        }
        for (int actualAgent = 0; actualAgent < this.agents.size();
actualAgent++)
        {
            Agent agent = this.agents.get(actualAgent);
```

```java
            // Asks to the Agent for the next action
            int[] action = this.GetAction(agent);
            long elapsedTime = agent.GetElapsedTime();
            interactions[actualAgent].SetAction(action);
            interactions[actualAgent].SetElapsedTime(elapsedTime);
        }

        // Makes the actions made by the Agents
        for (int actualAgent = 0; actualAgent < this.agents.size(); actualAgent++)
        {
            Agent agent = this.agents.get(actualAgent);
            int[] action = interactions[actualAgent].GetAction();
            this.MakeAction(agent, action);
        }
        // If Good and Evil Agents are in the same cell one of them doesn't move
        if (this.goodAgent.GetLocation() == this.evilAgent.GetLocation())
            // If the Good Agent didn't move the Evil Agent doesn't move
            if (this.goodAgent.GetLocation() == goodCell)
                this.LocateAgent(this.evilAgent, evilCell.GetNumber());
            // If the Evil Agent didn't move the Good Agent doesn't move
            else if (this.evilAgent.GetLocation() == evilCell)
                this.LocateAgent(this.goodAgent, goodCell.GetNumber());
            // If both Agents had move one of them doesn't move
            else if (new Random().nextInt(2) == 0)
                this.LocateAgent(this.goodAgent, goodCell.GetNumber());
            else
                this.LocateAgent(this.evilAgent, evilCell.GetNumber());

        // Set rewards to the Agents
        for (int actualAgent = 0; actualAgent < this.agents.size(); actualAgent++)
        {
            Agent agent = this.agents.get(actualAgent);
            // Evaluates the Agent action in the Environment
            double obtainedReward = this.EvaluateAction(agent, interactions[actualAgent]);
            interactions[actualAgent].SetReward(obtainedReward);
        }

        // Update the rewards of all cells by dividing by 2
        for (int index = 1; index <= this.space.GetNumberOfCells(); index++)
        {
            Cell cell = this.space.GetCell(index);
            cell.SetReward(cell.GetReward() / 2);
        }

        // Update the reward to 0 of all cells where there is an Agent
        for (Agent agent:this.agents)
        {
            Cell cell = agent.GetLocation();
            cell.SetReward(0.0);
        }
    }
```

## A1.2. Generación aleatoria de espacios

```
Random random = new Random();
Distribution distribution = new UniversalDistribution();
// Select the number of cells of the space
int numCells = distribution.GetValue(Environment_L.MIN_CELLS);
// Select the maximum actions available for the selected number of cells
int maxActions = Math.min(numCells, Environment_L.MAX_ACTIONS);
// Select the number of actions of the space
int numActions = distribution.GetValue(Environment_L.MIN_ACTIONS, maxActions);

// Constructs the space with the number of cells and actions selected
String description = "";
// Construct each cell
for (int cell = 0; cell < numCells; cell++)
{
    if (cell != 0)
        description += "|";
    // For each cell, generate each action
    for (int action = 1; action < numActions; action++)
    {
        int movements = random.nextInt(numCells);
        char movement = random.nextInt(2) == 0 ? '+' : '-';
        description += action;
        for (int index = 0; index < movements; index++)
            description += movement;
    }
}

return description;
```

# A2. Referencias